\documentclass{article}

\usepackage[final]{neurips_2022}

\usepackage[utf8]{inputenc} 
\usepackage[T1]{fontenc}    
\usepackage{hyperref}
\hypersetup{colorlinks=false, allcolors=black, pdfborder={0 0 0}}
\usepackage{url}[hyphens]            
\usepackage{booktabs}       
\usepackage{amsfonts}       
\usepackage{nicefrac}       
\usepackage{xcolor}         
\usepackage{enumitem}

\usepackage[pdftex]{graphicx}
\usepackage[font=small, labelformat=simple]{subcaption}

\usepackage{tikz}
\usepackage{changepage}
\usepackage{float}

\newcommand{\blue}[1]{\textcolor{black}{{#1}}}

\usepackage{amsmath,amsfonts,bm}

\def\eqref#1{equation~\ref{#1}}

\def\1{\bm{1}}

\def\rvb{{\mathbf{b}}}

\def\rvu{{\mathbf{i}}}

\def\rvs{{\mathbf{s}}}

\def\rvu{{\mathbf{u}}}

\def\rvx{{\mathbf{x}}}

\DeclareMathAlphabet{\mathsfit}{\encodingdefault}{\sfdefault}{m}{sl}
\SetMathAlphabet{\mathsfit}{bold}{\encodingdefault}{\sfdefault}{bx}{n}

\usepackage{array,booktabs}
\newcolumntype{M}[1]{>{\centering\arraybackslash}m{#1}} 

\usepackage[colorinlistoftodos]{todonotes}

\title{
Understanding the Evolution of Linear Regions in Deep Reinforcement Learning
}

\author{%
  Setareh Cohan\\ 
  Department of Computer Science\\
  University of British Columbia\\
  \texttt{setarehc@cs.ubc.ca}\\
  \And
  Nam Hee Kim\\
  Department of Computer Science\\
  Aalto University\\
  \texttt{namhee.kim@aalto.fi}\\
  \And
  David Rolnick\\
  School of Computer Science\\
  McGill University\\
  \texttt{drolnick@cs.mcgill.ca}\\
  \And
  Michiel van de Panne\\
  Department of Computer Science\\
  University of British Columbia\\
  \texttt{van@cs.ubc.ca}\\
}

\begin{document}

\maketitle

\begin{abstract}
 Policies produced by deep reinforcement learning are typically characterised by their learning curves, but they remain poorly understood in many other respects. ReLU-based policies result in a partitioning of the input space into piecewise linear regions. We seek to understand how observed region counts and their densities evolve during deep reinforcement learning using empirical results that span a range of continuous control tasks and policy network dimensions. Intuitively, we may expect that during training, the region density increases in the areas that are frequently visited by the policy, thereby affording fine-grained control. We use recent theoretical and empirical results for the linear regions induced by neural networks in supervised learning settings for grounding and comparison of our results. Empirically, we find that the region density increases only moderately throughout training, as measured along fixed trajectories coming from the final policy. However, the trajectories themselves also increase in length during training, and thus the region densities decrease as seen from the perspective of the current trajectory. Our findings suggest that the complexity of deep reinforcement learning policies does not principally emerge from a significant growth in the complexity of functions observed on-and-around trajectories of the policy. 

\end{abstract}

\section{Introduction}

Deep reinforcement learning (RL) utilizes neural networks to represent the policy and to train this network to optimize an objective, typically the expected value of time-discounted future rewards. Deep RL algorithms have been successfully applied to diverse applications including robotics, challenging games, and an increasing number of real-world decision-and-control problems \citep{franccois2018introduction}. 
For a given choice of task, RL algorithm, and policy network configuration, the performance is commonly characterised via the learning curves, which provide insight into the learning efficiency and the final performance. 
However, little has been done to understand the detailed structure of the state-to-action mappings induced by the control policies and how these evolve over time.

\begin{figure}[tb]
     \centering
\includegraphics[width=0.5\textwidth]{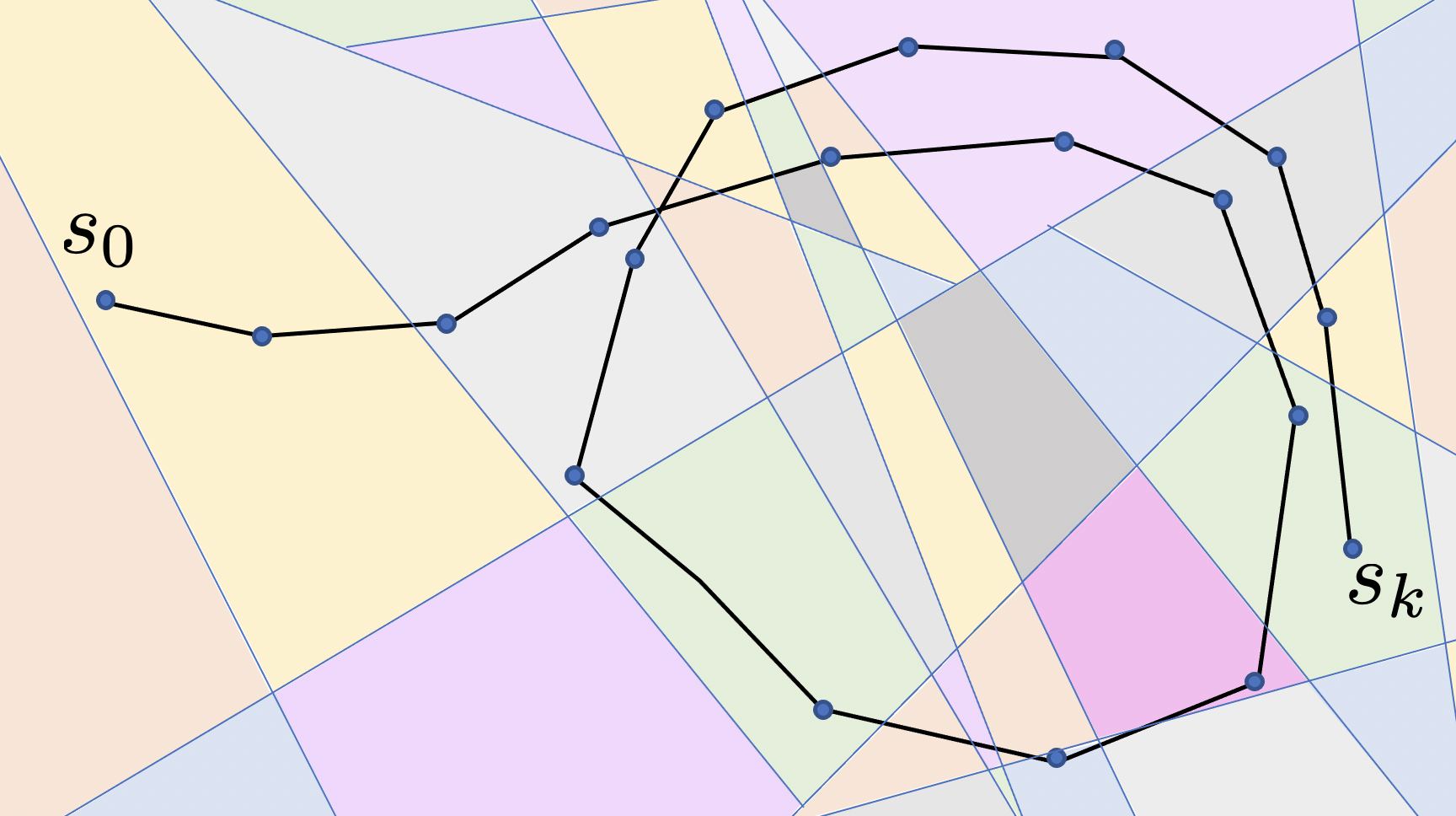}
\caption{Schematic illustration of a trajectory traversing the piecewise linear regions in the policy state space. $S_0$ and $S_k$ indicate the initial and final states of the trajectory.}
\label{fig:abstract_regions}
\end{figure}

In this work, we aim to further understand deep feed-forward neural network policies that use rectified linear activation functions (rectifier linear units or ReLUs). ReLUs \citep{nair2010rectified} are among the most popular choices of activation functions due to their practical successes \citep{montufar2014number}. For RL, these activations induce a piecewise linear mapping from states to actions, where the input space, i.e, the state space, is divided into distinct \emph{linear regions}, where within each region, the actions are a linear function of the states. Note that the regions are formally an affine function of the input, due to the constant-valued bias terms. For simplicity and convenience, these are more commonly described as linear regions. Figure~\ref{fig:abstract_regions} provides a schematic illustration of these regions, along with a policy trajectory.

The number of distinct regions into which the input space is divided is a natural measure of network expressivity. Learned functions with many linear regions have the capacity to build complex and flexible decision boundaries. Thus, the problem of counting the number of linear regions has been extensively studied in recent literature \citep{montufar2014number, raghu2017expressive, hanin2019complexity, serra2018bounding}. While the maximum number of regions is exponential with respect to network depth \citep{montufar2014number}, recent work has demonstrated that the number of regions is instead typically proportional to the number of neurons \citep{hanin2019complexity}.

For RL, we are interested in the local granularity (density) of linear regions along trajectories arising from the policy. Fine-grained regions afford fine-grained control, and thus we may hypothesize that region density \emph{increases} in regions frequently visited by the policy, in order to afford better control. Recent work in supervised-learning of image classification is inconsistent with regard to findings about the region density seen in the vicinity of data points, with some reporting a decrease~\citep{novak2018sensitivity} to provide better generalization and robustness to perturbation, and others not~\citep{hanin2019complexity}.  For the RL setting, we note that counting regions visited along an episode trajectory arguably provides a meaningful and task-grounded measurement in contrast to line-segments and ellipses that pass through randomly sampled points sampled from training data, which have been used in the prior works mentioned above. 
\blue{We further note that piecewise-affine control strategies are commonly designed into control systems, e.g., via gain scheduling. Understanding how these regions are designed and distributed by deep RL thus helps establish bridges with these existing methods.}

To the best of our knowledge, our work is the first to investigate the structure and evolution of linear regions of ReLU-based deep RL policies in detail. We seek to answer several basic empirical questions : 
\begin{description}[labelindent=0.55cm]
\item[Q1] Do findings for network expressivity, originally developed in supervised learning settings, apply to RL policies? How are the region densities affected by the policy network configuration? Do deeper policy networks result in finer-grained regions and hence an increased expressivity?
\item[Q2] How do the linear regions of a policy evolve during training? Do we see a significantly greater density of regions emerge along the areas of the state space frequently visited by the episodic trajectories, thereby allowing for finer-grained control? Do random-action trajectories see different densities?
\end{description}

The key results can be summarized as follows, for policies trained using proximal policy optimization (PPO)~\citep{schulman2017proximal}, and evaluated on four different continuous control tasks.
Q1: There is a general alignment with recent theoretical and empirical results for supervised learning settings. Region density is principally proportional to the number of neurons, with an additional small observed increase in density for deeper networks. 
Q2: Only a moderate increase of density is observed during training, as measured along fixed final-policy trajectories. Therefore, the complexity of a final learned policy does \emph{not} come principally from increased density on-and-around the optimal trajectories, which is a potentially surprising result.
In contrast, as measured along the evolving current-policy trajectories, a decrease in region density is observed during training. Across all settings, we also observe that the region-transition count, as observed during fixed time-duration episodes, grows during training before converging to a plateau.  However, the trajectory length, as measured in the input space, also grows towards a plateau, although not at the same rate, and this leads to variations in the mean region densities as observed along current trajectories during training.

\section{Related Work}

\label{related_work}


Understanding the expressivity of a neural network is fundamental to better understanding its operation. Several works study exprressivity of deep neural networks with piecewise linear activations by counting their linear regions \citep{arora2016understanding, bianchini2014complexity}. On the theoretical side, \citet{pascanu2013number} show that in the asymptotic limit of many hidden layers, deep ReLU networks are capable of separating their input space into exponentially more linear regions compared with shallow networks, despite using the same number of computational units. Following this, \citet{montufar2014number} also explore the complexity of functions computable by deep feed-forward neural networks with ReLU and maxout activations, and provide tighter upper and lower bounds for the maximal number of linear regions. They show that the number of linear regions is polynomial in the network width and exponential with respect to the network depth. Furthermore, \citet{raghu2017expressive} improve the upper bound for ReLU networks by introducing a new set of expressivity measures and showing that expressivity grows exponentially with network depth for ReLU and tanh activated neural networks. \citet{serra2018bounding} generalizes these results by providing even tighter upper and lower bounds for the number of regions for ReLU networks and show that the maximal number of regions grows exponentially with depth when input dimension is sufficiently large.  


More recent works that touch on expressivity of depth show that the effect of depth on the expressivity of neural networks is likely far below that of the theoretical maximum proposed by prior literature. \citet{hanin2019deep} study the importance of depth on expressivity of neural networks in practice. They show that the average number of linear regions for ReLU networks at initialization is bounded by the number of neurons raised to the input dimension, and is independent of network depth. They also empirically show that this bound remains tight during training. Similarly, \citet{hanin2019complexity} find that the average distance to the boundary of the linear regions depends only on the number of neurons and not on the network depth – both at initialization and during training for supervised-learning tasks on ReLU networks. This strongly suggests that deeper networks do not necessarily learn more complex functions in comparison to shallow networks. Prior to this, a number of works have shown that the strength of deep learning may arise in part from a good match between deep architectures and current training procedures \citep{mhaskar2016deep, mhaskar2016learning, zhang2021understanding}. Notably, \citet{ba2014deep} show that, once deep networks are trained to perform a task successfully, their behavior can often be replicated by shallow networks, suggesting that the advantages of depth may be linked to easier learning.


Another line of work studies function complexity in terms of robustness to perturbations to the input. \citet{sokolic2017robust} theoretically studies the input-output Jacobian, which is a measure of robustness and also relates to generalization. Similarly, \citet{zahavy2016ensemble} propose a sensitivity measure in terms of adversarial robustness, and provides theoretical and experimental insights on how it relates to generalization. \citet{novak2018sensitivity} also study robustness and sensitivity using the input-output Jacobian and number of transitions along trajectories in the input space as measures of robustness. They show that neural networks trained for image classification tasks are more robust to input perturbations in the vicinity of the training data manifold, due to training points lying in regions of lower density. Several other recent works have also focused on proposing tight generalization bounds for neural networks \citep{bartlett2017spectrally, dziugaite2017computing, neyshabur2017exploring}.


There are a number of works that touch on understanding deep neural networks by finding general principles and patterns during training. \citet{arpit2017closer} empirically show that deep networks prioritize learning simple patterns of the data during training. \citet{xu2019training} find a similar phenomenon in the case of $2$-layer networks with Sigmoid activations. \citet{rahaman2019spectral} study deep ReLU activated networks through the lens of Fourier analysis and show that while deep neural networks can approximate arbitrary functions, they favour low frequency ones and thus, they exhibit a bias towards smooth functions. \citet{samek2017explainable} present two approaches in explaining predictions of deep learning models in a classification task, with the first method computing the sensitivity of the prediction with respect to input perturbations and the second method that meaningfully decomposes the decision in terms of input variables.

While deep RL methods are widely used and extensively studied, few works focus on understanding the policy structure in detail. \citet{zahavy2016graying} propose a visualization method to interpret the agent's actions by describing the Markov Decision Process as a directed graph on a t-SNE map. They then suggest ways to interpret, debug and optimize deep neural network policies using the proposed visualization maps. \citet{rupprecht2019finding} train a generative model over the state space of Atari games to visualize states which minimize or maximize given action probabilities. \citet{luo2018visual} adapt three visualization techniques to the domain of image-based RL using convolutional neural networks in order to understand the decision making process of the RL agent.
\section{Piecewise Linear Regions}  

Throughout this work, we consider RL policies based on \emph{ReLU networks}. A ReLU network is a ReLU-activated feed-forward neural network, or a multi-layer perceptron (MLP) which can be formulated as a scalar function $f: \mathbb{R}^d \rightarrow \mathbb{R}^o$ defined by a neural network with $L$ hidden layers of width $d_1,...,d_L$ and a $o$-dimensional output. Assuming the output to be 1-dimensional, and following the notation of \cite{rahaman2019spectral}, we have:

\begin{align}
    f(\rvx) = (T^{(L+1)} \circ \sigma \circ T^{(L)} \circ ... \circ \sigma \circ T^{(1)}) (\rvx),
    \label{eq:relu_network}
\end{align}

where $T^{(k)}: \mathbb{R}^{d_{k-1}} \rightarrow \mathbb{R}^{d_{k}}$ computes the weighted sum $T^{(k)}(\rvx) = W^{(k)}\rvx + \rvb^{(k)}$ for some weight matrix $W^{(k)}$ and bias vector $\rvb^{(k)}$. Here, $\sigma(\rvu) = \max (0, \rvu)$ denotes the ReLU activation function acting element-wise on vector $\rvu=(u_1, ..., u_n)$.

Given the ReLU network $f$ from Equation~\ref{eq:relu_network}, again following \citet{rahaman2019spectral}, piecewise linearity can be explicit written by  

\begin{align}
    f(\rvx) = \sum_{r \in R} 1_{P_r}(\rvx) (W_r \rvx + \rvb_r)
    \label{eq:piecewise_linearity}
\end{align}

where $r$ is the index for the linear region $P_r$ and $1_{P_r}$ is the indicator function on $P_r$. The $1 \times d$ matrix $W_r$ is given by:

\begin{align}
    W_r = W^{(L+1)} W_r^{(L)} ... W_r^{(1)}
    \label{eq:W}
\end{align}

where $W_r^{(k)}$ is obtained from the original weight $W^{(k)}$ by setting its $j^{\text{th}}$ column to zero whenever neuron $j$ of layer $k{-}1$ is inactive for $k > 1$.

\section{Counting Linear Regions in RL Policies}
\label{sec:counting}

\begin{figure}[tb]
    \centering
    \begin{subfigure}[t]{0.69\textwidth}
        \vskip 0pt
        \centering 
        \includegraphics[width=\textwidth]{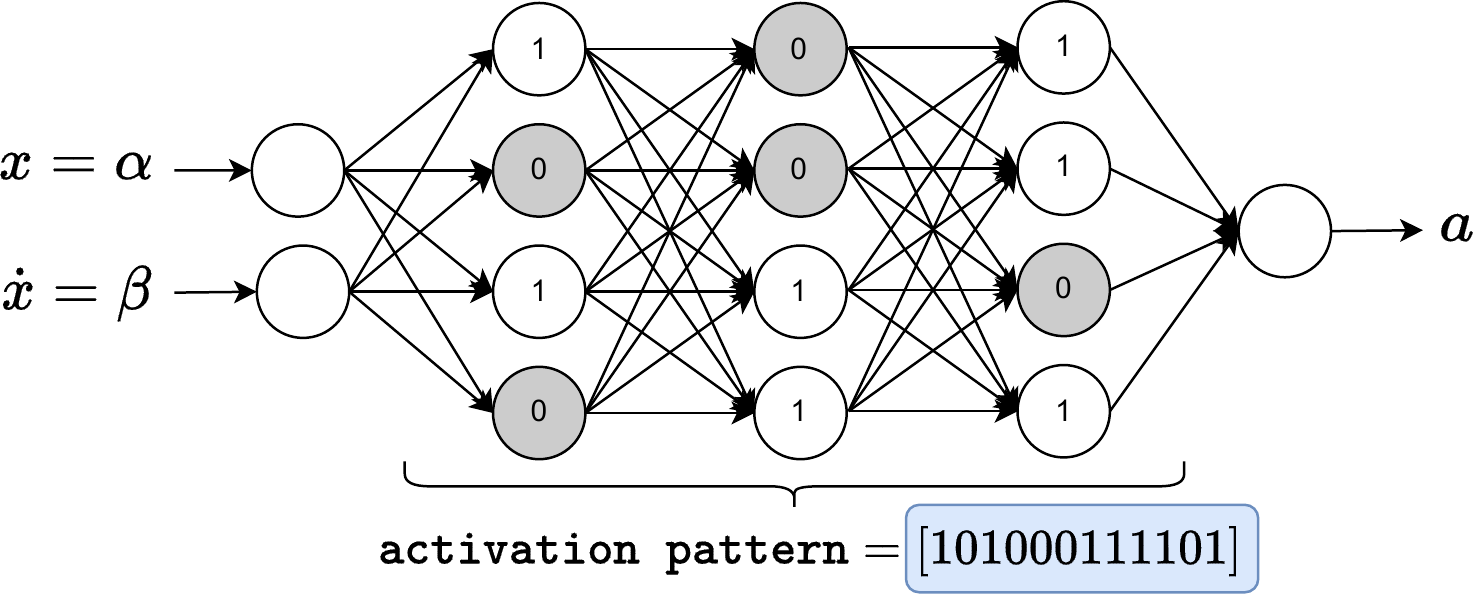}
    \end{subfigure}
    \hfill
    \begin{subfigure}[t]{0.30\textwidth}
        \vskip 2pt
        \centering 
        \includegraphics[width=\textwidth]{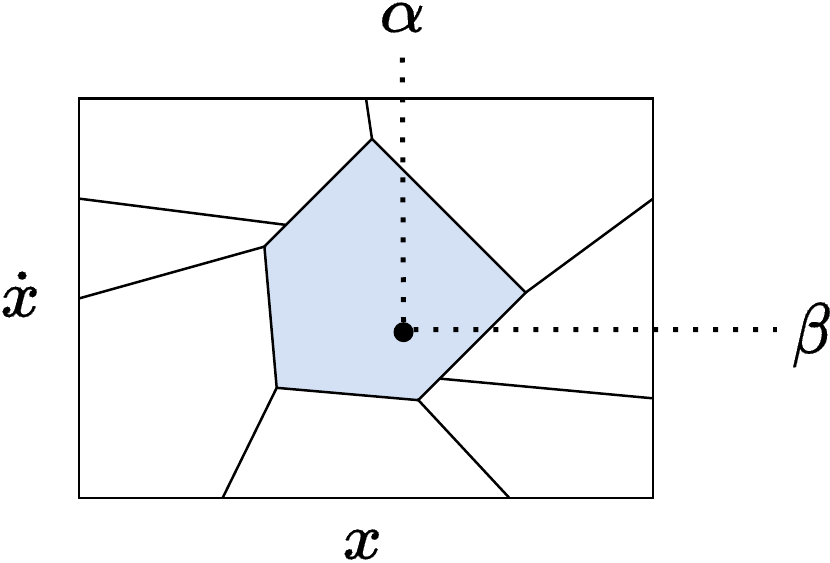}
    \end{subfigure}
    \caption{An overview of a ReLU-activated policy network and the binary labeling scheme of the linear regions. State $\rvs = [x,\dot{x}]$ is within the linear region uniquely identified by the activation pattern computed by concatenating the binary states of the activations of the policy network given input $\rvs$.}
    \label{fig:framework}
\end{figure}

To answer the key questions posed in the introduction, we need a method for counting the linear regions encountered during the episodic trajectories taken during RL. For each input $\rvs$, we encode each neuron of the policy network with a binary code of $0$ if its pre-activation is negative, and with a binary code $1$ otherwise. The linear region of the input $\rvs$ can thus be uniquely identified by the concatenation of binary codes of all the neurons in the network called an \emph{activation pattern}. 
Figure~\ref{fig:framework} illustrates the activation pattern construction for a 2D input space.

Figure~\ref{fig:abstract_regions} provides a schematic illustration of the linear regions for a 2D state space, together with an example trajectory that consists of sequential states encountered by the policy, connected with straight-line segments. 
As seen in the figure, regions can be revisited, which leads to a distinction between the number of transitions between regions and the number of unique regions visited along a trajectory. We compute both of these metrics along episodic trajectories, as enabled by keeping a list of the regions already visited at any point in time when processing a given episode trajectory. Our region-counting method also includes regions that are crossed by the straight-line segments between successive trajectory states, although the policy itself does not explicitly encounter these regions due to the discrete-time nature of typical RL environments.

To represent the $k$-th line segment of the episodic trajectories in the input space, we develop a parameterized line segment given by $s(u)=(1-u) s_{k-1} + u s_{k}$, 
where $u\in[0,1]$ and $s_{k-1}, s_{k}\in\mathbb{R}^d$ denote the endpoints of the line segment. We then calculate the exact number of the linear regions over this line segment by considering the hidden layers of the policy network one at a time, starting from input towards the output, and observe how each region can be split by the neurons. Starting with the first layer, we consider neurons one by one, and identify the point in the domain of $u$, if any, that induces a change in the binary labeling for that neuron. Each such point subdivides the domain of $u$ into two new regions where the pre-activation will be zero for one of them in the next layer. By maintaining a list of these regions and the linear functions defined over them, and whether their pre-activation vanishes in the next layer, we proceed to the neurons of the next layer. This process repeats for each of the regions resulting from the previous layer. We record the activation patterns of all the final regions, to track state visits during all segments of episode trajectories. In the end, for each trajectory $\tau$, we compute the total region transitions, $R_T(\tau)$, as well as the number of unique visited regions, $R_U(\tau)$, where $R_T(\tau) + 1 \geq R_U(\tau) $. We further compute the trajectory length in the input space, $L(\tau)$. This allows us to compute normalized region densities according to $\rho(\tau) = R_T(\tau)/(N L(\tau))$ where $N$ is the total number of neurons in the policy network, in accordance with \citet{hanin2019complexity}.

\begin{figure}[t]
\setlength\tabcolsep{1pt} 
\centering
\begin{tabular}{@{} r M{0.22\linewidth} M{0.22\linewidth} M{0.22\linewidth} M{0.22\linewidth} @{}}
& $\texttt{Epoch}=0$ & $\texttt{Epoch}=10$ & $\texttt{Epoch}=20$ & $\texttt{Epoch}=100$\\
\begin{subfigure}{0.07\linewidth} \caption*{$\tau^*$} \end{subfigure} 
  & \includegraphics[width=\hsize, trim={0 0 0 0.7cm },clip] {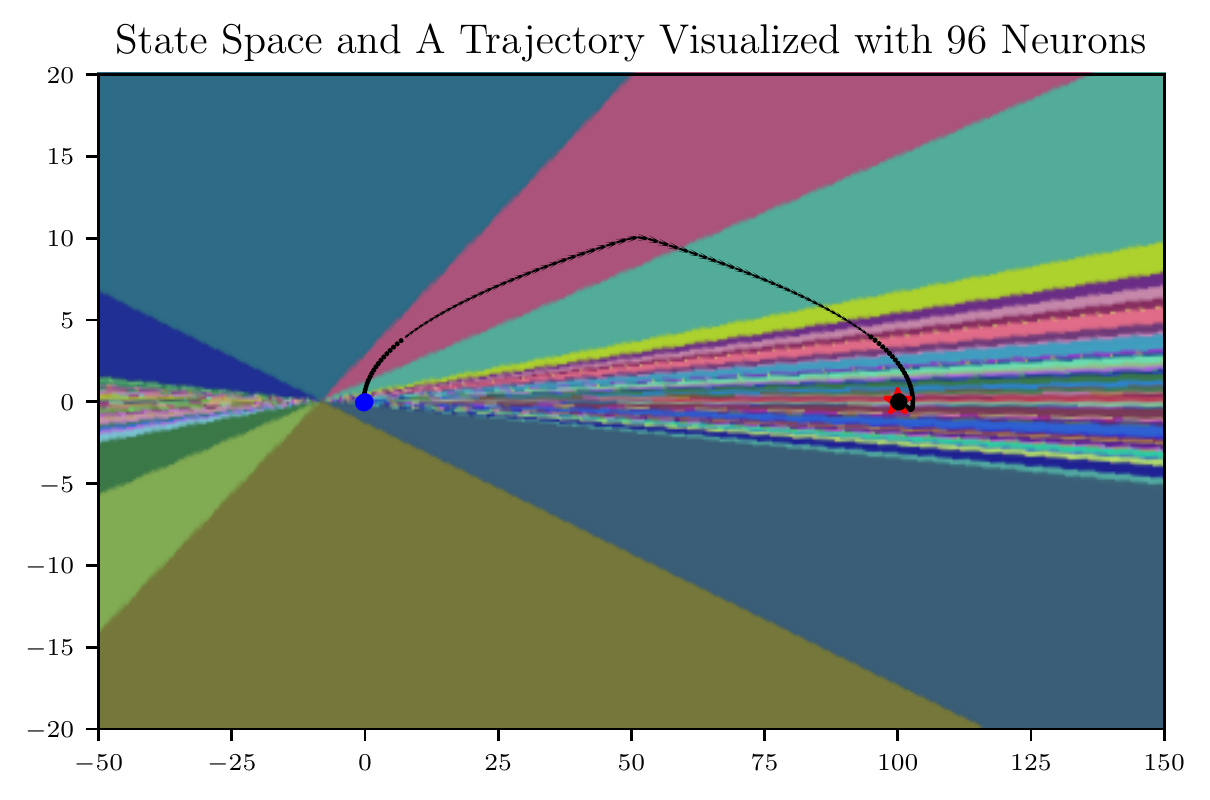} 
  & \includegraphics[width=\hsize, trim={0 0 0 0.7cm },clip] {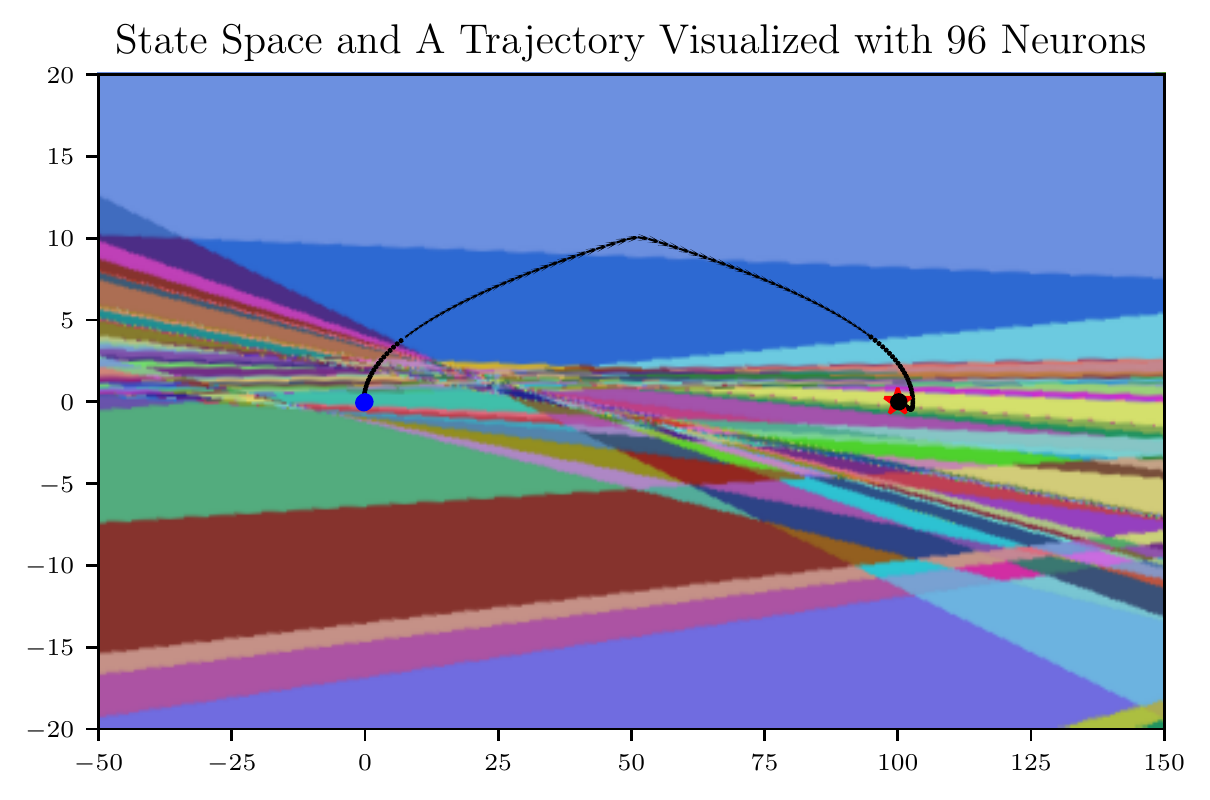} 
  & \includegraphics[width=\hsize, trim={0 0 0 0.7cm },clip] {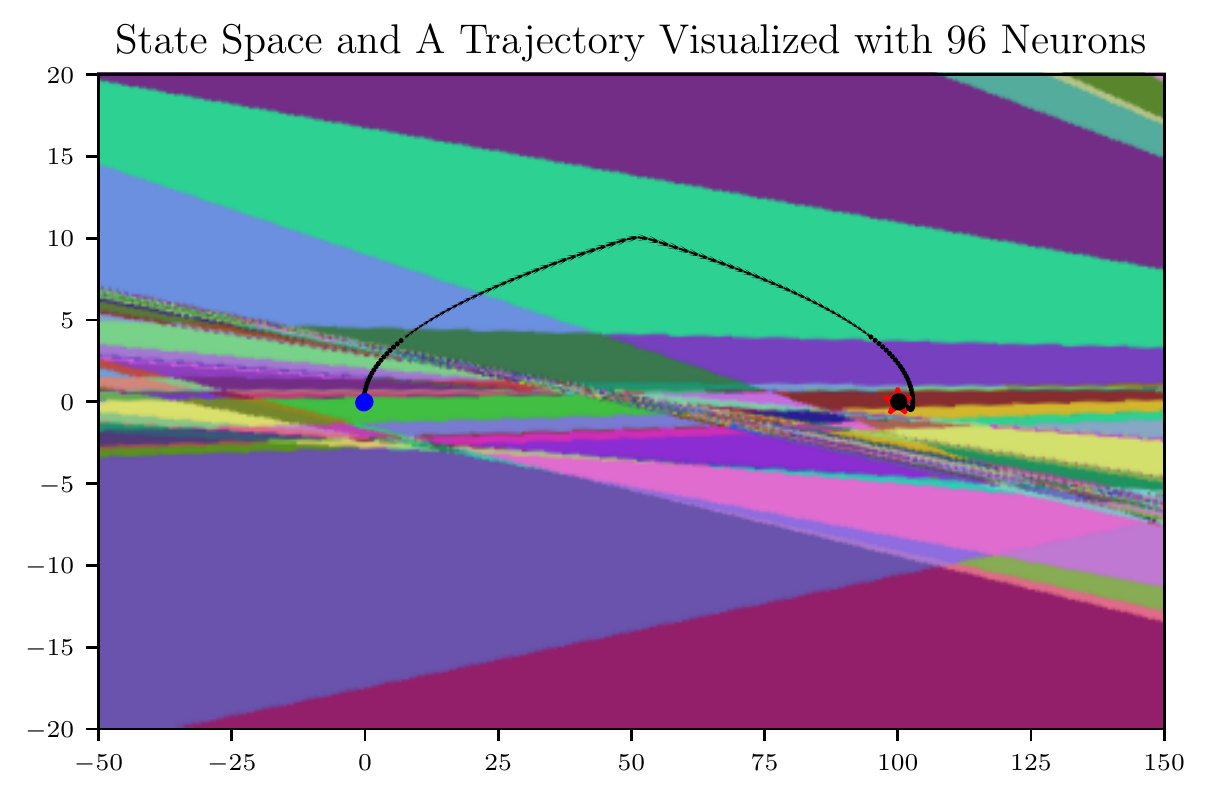} 
  & \includegraphics[width=\hsize, trim={0 0 0 0.7cm },clip] {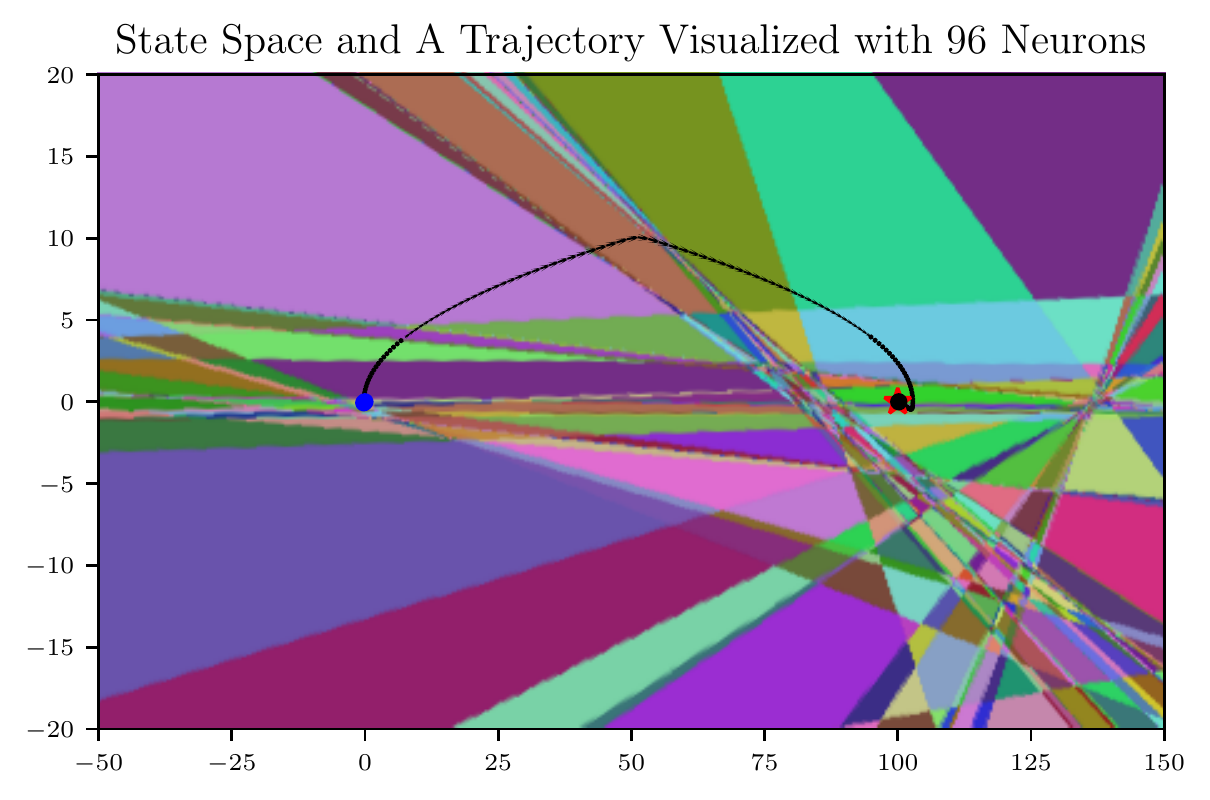}\\ \addlinespace
\begin{subfigure}{0.07\linewidth} \caption*{$\tau$} \end{subfigure} 
  & \includegraphics[width=\hsize, trim={0 0 0 0.7cm },clip] {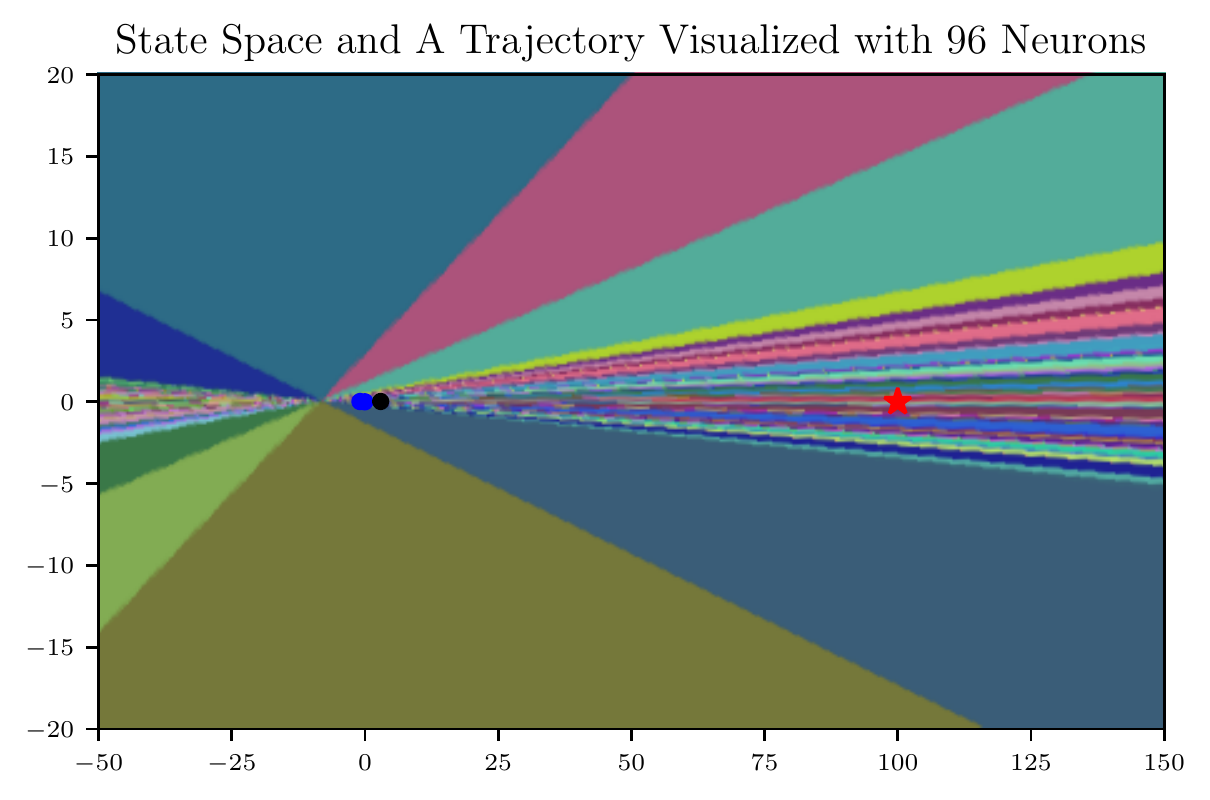}    
  & \includegraphics[width=\hsize, trim={0 0 0 0.7cm },clip] {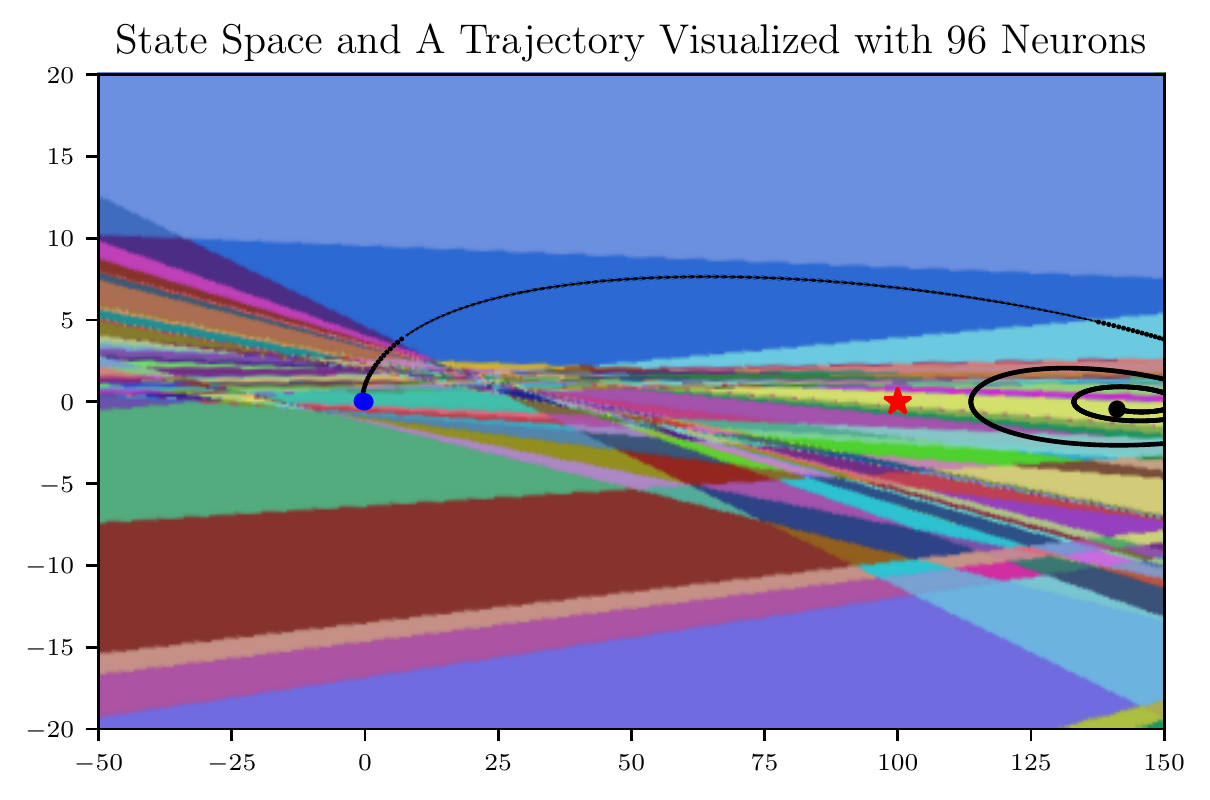} 
  & \includegraphics[width=\hsize, trim={0 0 0 0.7cm },clip] {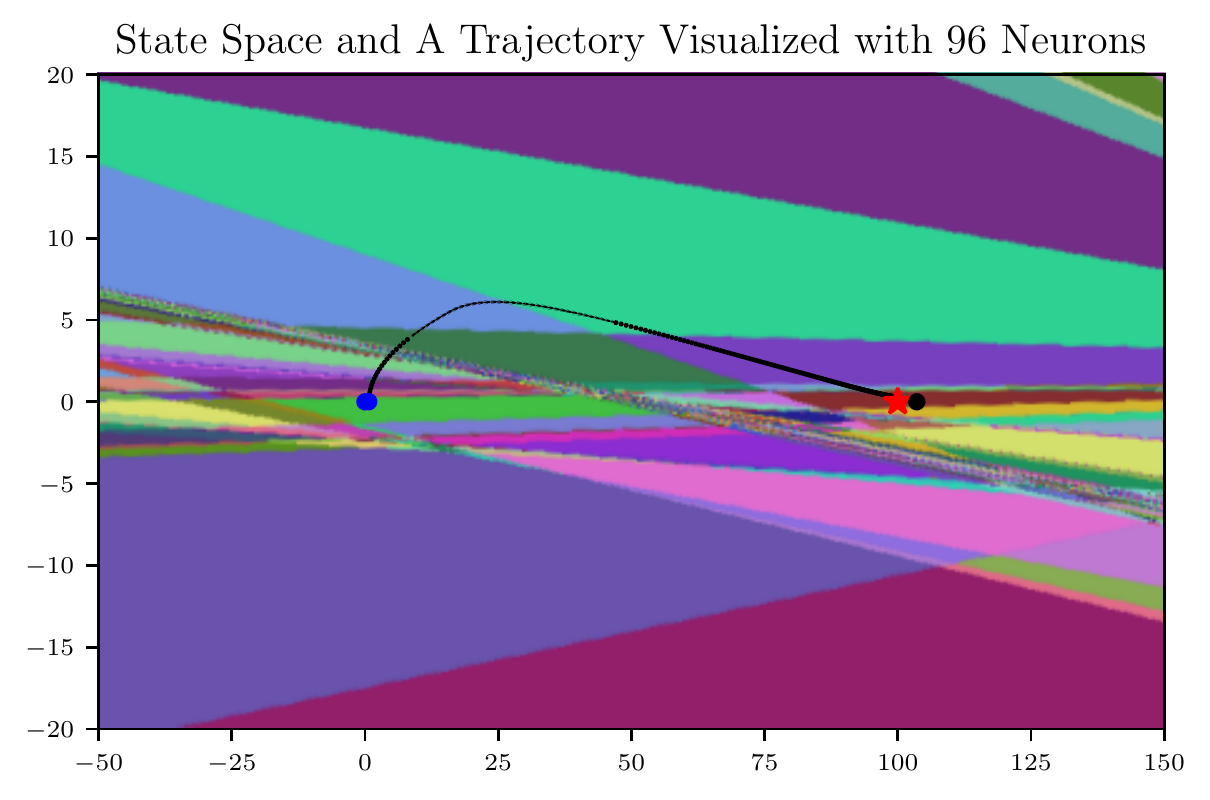} 
  & \includegraphics[width=\hsize, trim={0 0 0 0.7cm },clip] {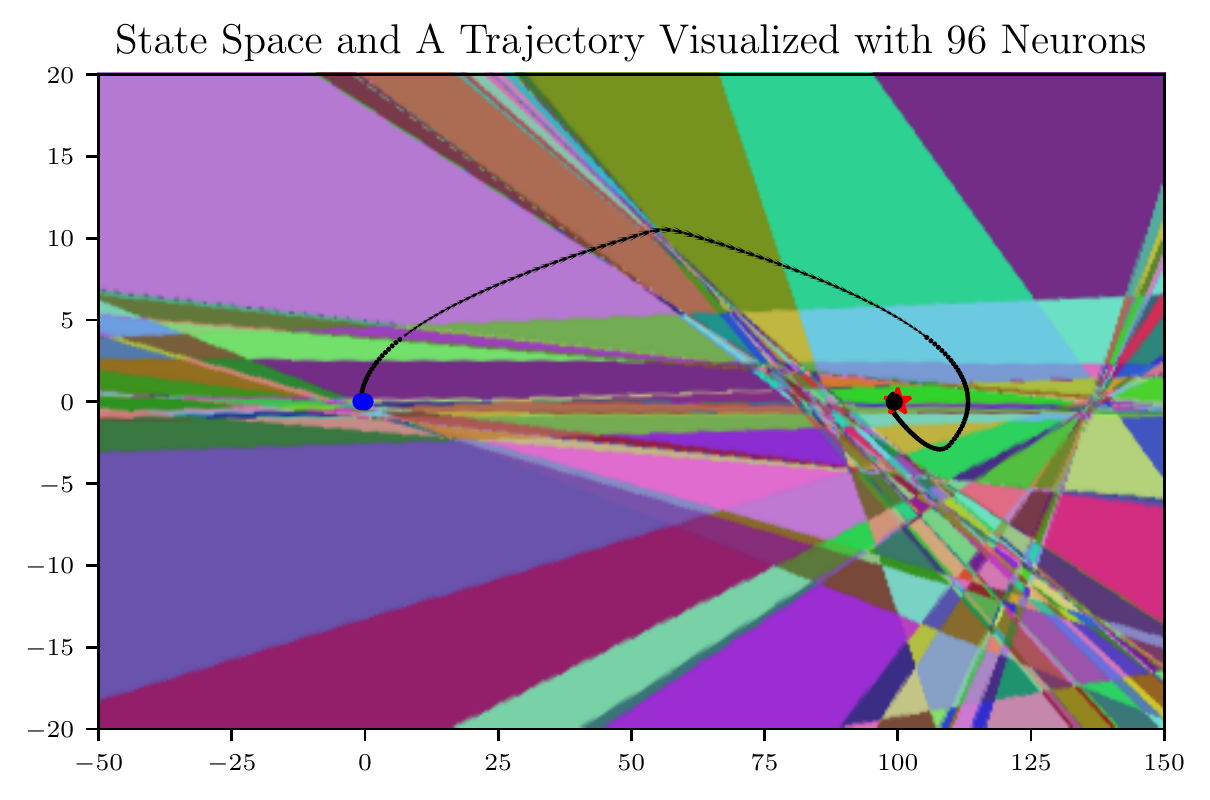}\\
\end{tabular}
\caption{State space visualization with current ($\tau$) and fixed trajectories ($\tau^*$) of a ReLU-activated policy network with depth 2 and width 8 trained on a 2D toy environment during training. The first row, plots a fixed trajectory sampled from the final policy over the state space. The second row plots a trajectory sampled from the current snapshot of the policy. Columns indicate the number of training epochs. $\texttt{Epoch}=0$ corresponds to the policy at initialization and $\texttt{Epoch}=100$ corresponds to the fully-trained policy. In this toy environment, an agent starts at the origin indicated with a blue dot and the goal is for the agent to reach a target
state located at $x=100$. States are the position $x$ (horizontal axis) and velocity $\dot{x}$ (vertical axis) of the agent, and action is the acceleration.}
\label{fig:2d_visualization}
\end{figure}

We use the above metrics along two types of trajectories. First, we consider {\em fixed} trajectories, $\tau^{*}$ as sampled from the final fully-trained policy. Second, we consider {\em current} trajectories, $\tau$, as sampled from the current policy during training.  Both of these trajectories offer informative views of the evolution of the policy. The former offers a direct picture of the linear-region density along a meaningful, fixed region of the state space, i.e., that of the final optimized policy. The latter offers a view of what the policy trajectories actually encounter during training. In order to better understand the evolution of $\tau$, we also track its length, $L(\tau)$, given that the density is determined by the region-transition count as well as the length of the trajectory. Figure~\ref{fig:2d_visualization} shows the evolution of linear regions and the types of trajectories $\tau^{*}$ and ${\tau}$ for a simple 2D toy environment, and a ReLU-activated policy network of depth 2 and width 8. 

\section{Experimental Results}
\label{sec:results}

We conduct our experiments on four continuous control tasks including HalfCheetah-v2, Walker-v2, Ant-v2, and Swimmer-v2 environments from the OpenAI gym benchmark suits~\citep{brockman2016openai}. We use the Stable-Baselines3 implementations of the PPO algorithm \citep{schulman2017proximal} in all of our experiments throughout this work \footnote{Our code is available at \url{https://github.com/setarehc/deep_rl_regions}.}. We run each experiment with 5 different random seeds and the results show the mean and the standard deviation across the random seeds. The policy network architecture is an MLP with ReLU activations in all hidden layers in our experiments.

We train 18 policy network configurations with $N \in \{32,48,64,96,128,192\}$ neurons,
widths $w \in \{8,16,32,64\}$, and depths $d \in \{1,2,3,4\}$. We use a fixed value function network structure of $(64,64)$ in all of our experiments. Details of the network configurations used for the policy network are available in Appendix~\ref{appendix:Network Dimensions}. These network configurations are chosen such that the network is fully-capable of learning the task and achieves near state-of-the-art cumulative reward on the particular task it is trained on. This allows us to study the evolution of linear regions disentangled from properties such as trainability, which is outside the scope of this work. The table showing the range of the mean cumulative reward per task, across all network configurations can be found in Appendix~\ref{appendix:Experimental Setup}. 
 
We adopt the network initialization and hyperparameters of PPO from StableBaselines3~\citep{raffin2021stable} and train our policy networks on 2M samples (i.e. 2M timesteps in the environment).
This initialization scheme also matches the best-practice recommendations for on-policy RL, e.g.,~\citep{andrychowicz2020matters}, although it differs from that used in~\cite{hanin2019complexity} for theoretical and empirical linear region analysis. A detailed explanation of the initialization and choice of hyperparameters is available in Appendix~\ref{appendix:Initialization and Hyperparameters}.

In what follows, we focus on the results for the HalfCheetah-v2 environment as a representative example.
Similar results for the other environments are provided in Appendix~\ref{appendix:Additional Results}.

    \begin{figure}[tb!]
        \centering
        \begin{subfigure}[b]{0.45\textwidth}  
            \centering 
            \includegraphics[width=\textwidth]{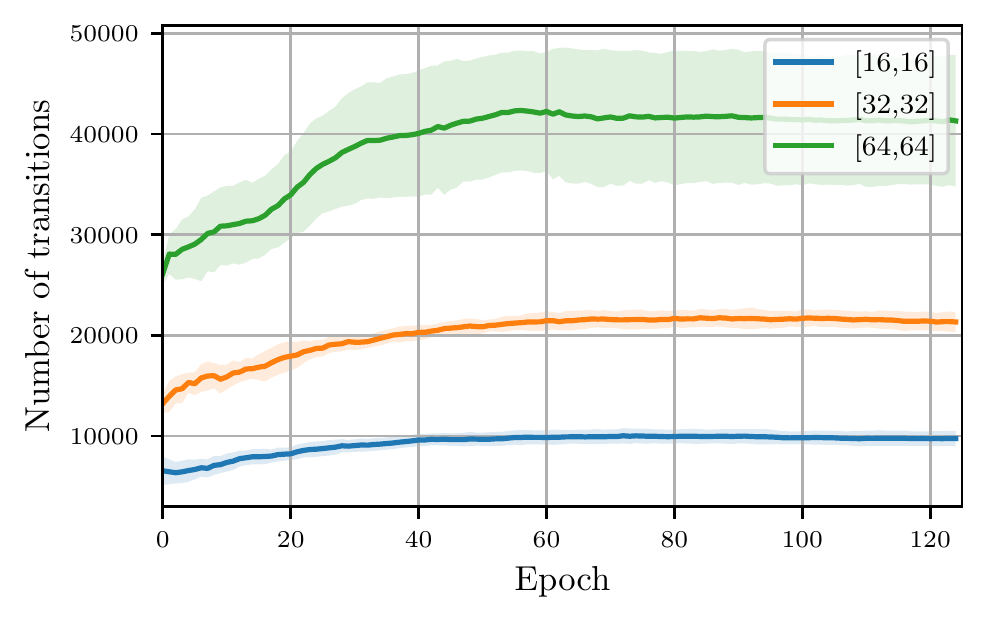}
            \caption{{\small Transition count, fixed trajectory $\tau^*$}}    
            \label{fig:trans-fixed}
        \end{subfigure}
        \hfill
        \begin{subfigure}[b]{0.45\textwidth}   
            \centering 
            \includegraphics[width=\textwidth]{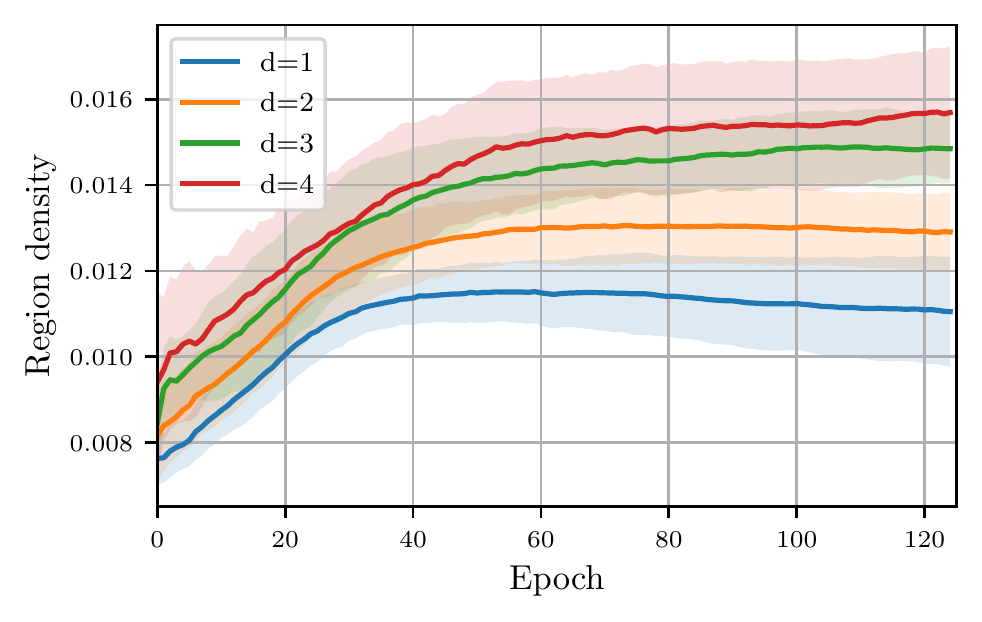}
            \caption{{\small Region density, fixed trajectory $\tau^*$}}    
            \label{fig:density-fixed}
        \end{subfigure}
        \vskip\baselineskip
        \begin{subfigure}[b]{0.45\textwidth}
            \centering
            \includegraphics[width=\textwidth]{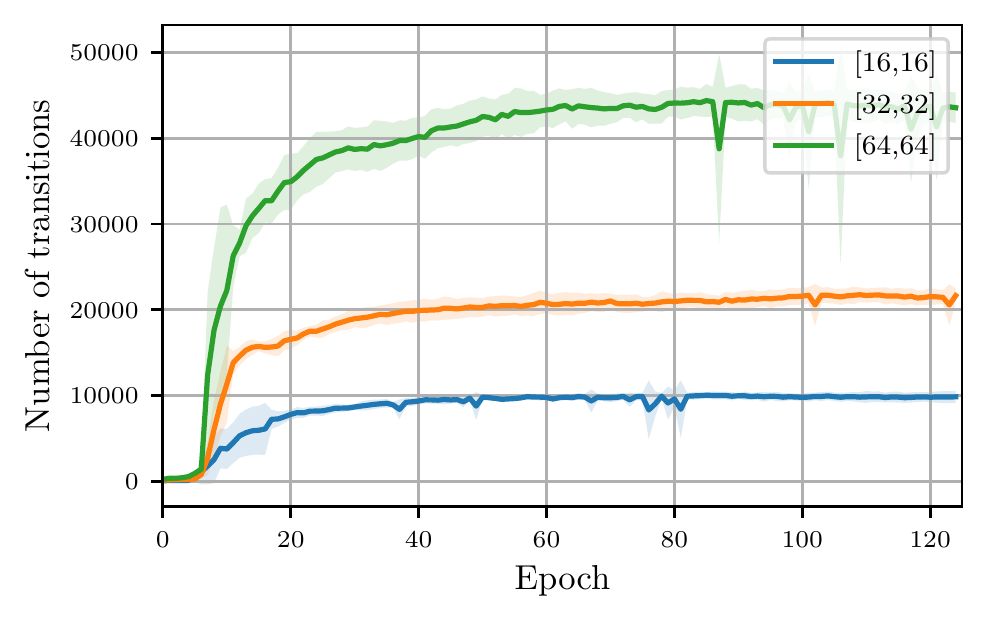}
            \caption{{\small Transition count, current trajectory $\tau$}}    
            \label{fig:trans-current}
        \end{subfigure}
        \hfill        
        \begin{subfigure}[b]{0.45\textwidth}   
            \centering 
            \includegraphics[width=\textwidth]{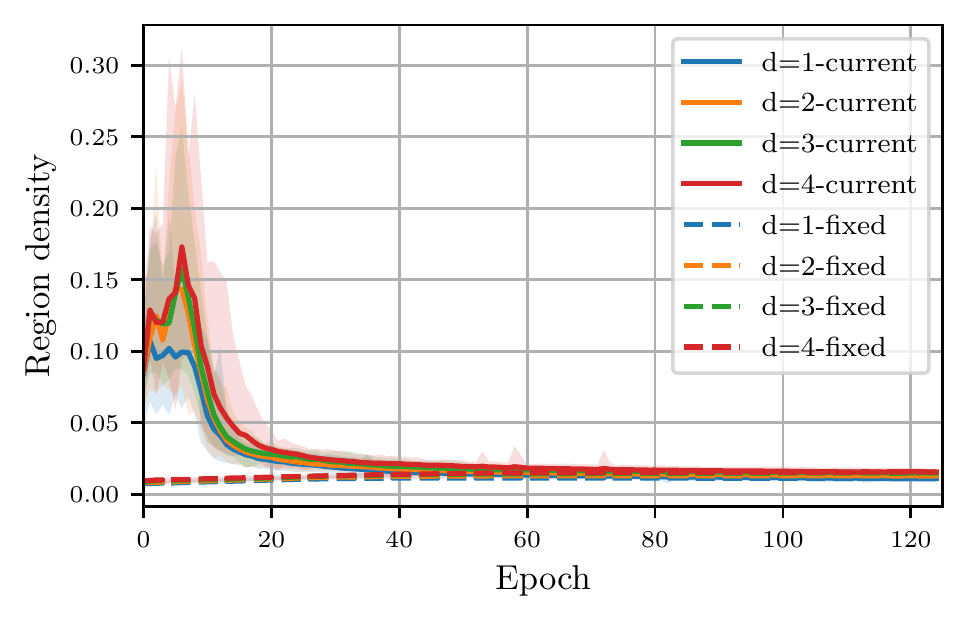}
            \caption{{\small Region density, $\tau^*$ and $\tau$}}    
            \label{fig:density-both}
        \end{subfigure}
        \vskip\baselineskip
        \begin{subfigure}[b]{0.45\textwidth}   
            \centering 
            \includegraphics[width=\textwidth]{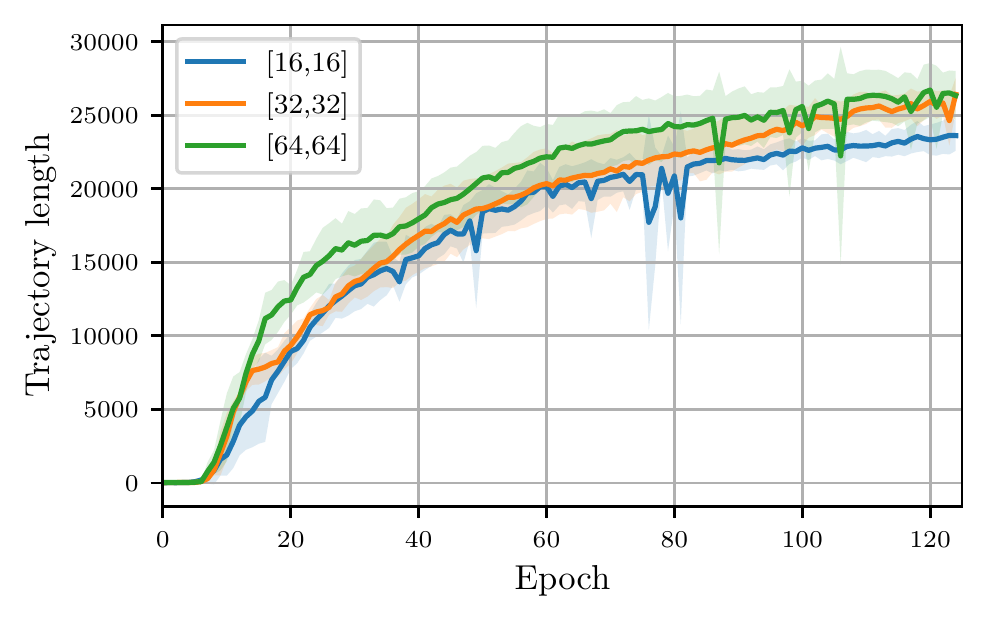}
            \caption{{\small Trajectory length, current trajectory $\tau$}}    
            \label{fig:length-curr}
        \end{subfigure}
        \hfill        
        \begin{subfigure}[b]{0.45\textwidth}   
            \centering 
            \includegraphics[width=\textwidth]{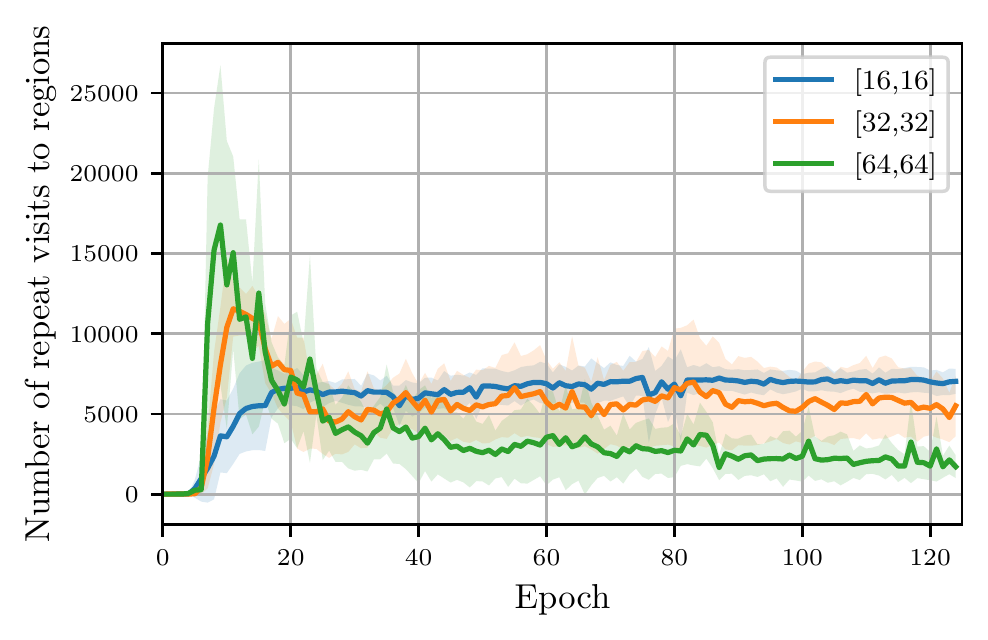}
            \caption{{\small Repeat visits to regions, current trajectory $\tau$}}    
            \label{fig:repeat-visits-current}
        \end{subfigure}
        \vskip\baselineskip
        \begin{subfigure}[b]{0.45\textwidth}  
            \centering 
            \includegraphics[width=\textwidth]{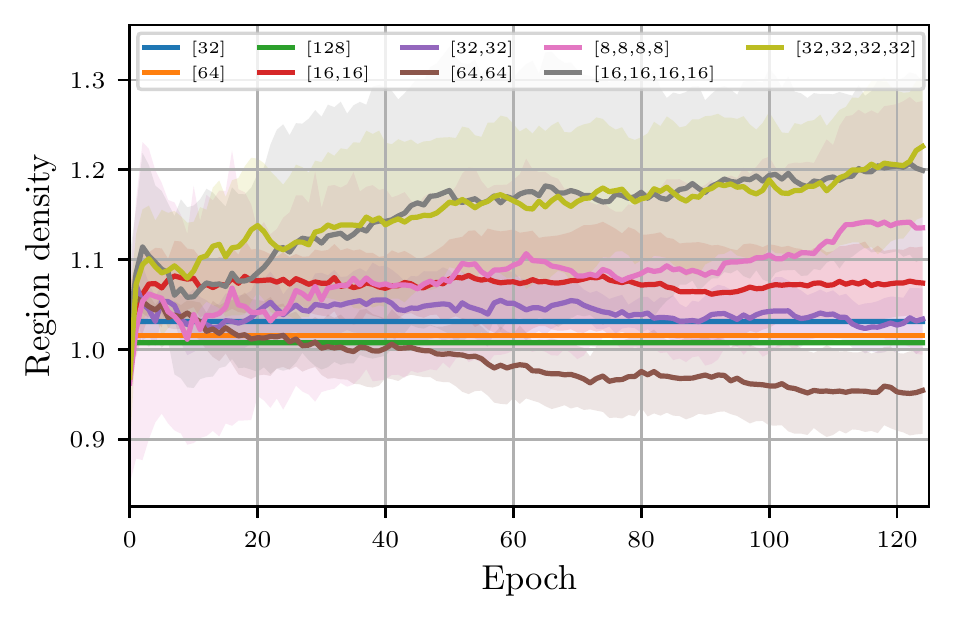}
            \caption{{\small Density for random lines through the origin}}    
            \label{fig:density-random-line-origin}
        \end{subfigure}
        \hfill
        \begin{subfigure}[b]{0.45\textwidth}   
            \centering 
            \includegraphics[width=\textwidth]{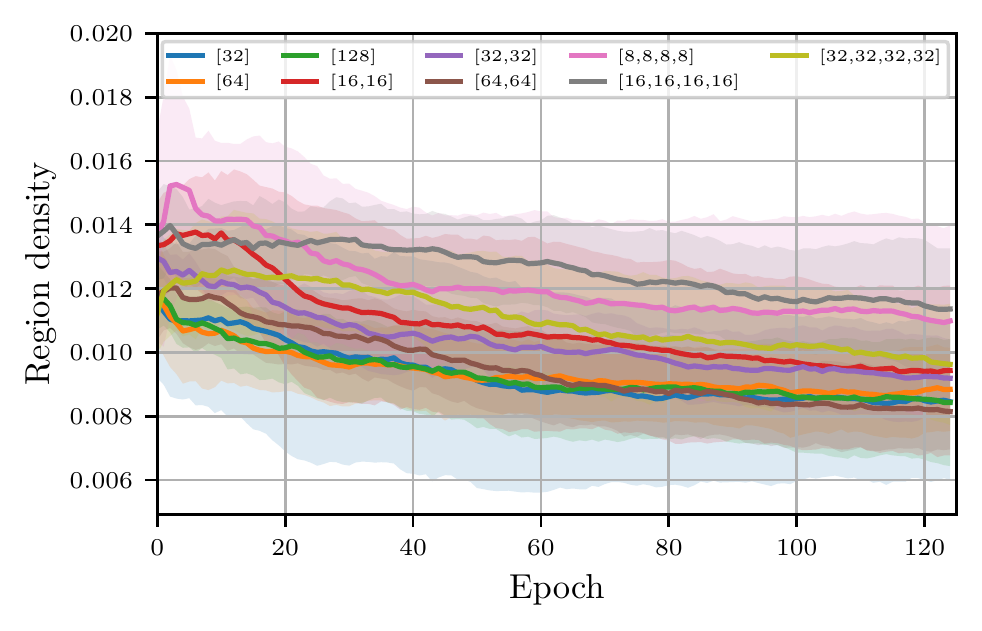}
            \caption{{\small Density for random-action trajectories}}    
            \label{fig:density-random-traj}
        \end{subfigure}
        \caption{{\small Evolution of the number of transitions, linear region densities, and length of trajectories for HalfCheetah, during training. Results for the rest of the environments are provided in Appendix~\ref{appendix:Additional Results}. The ranges indicate the standard error across 5 random seeds. $[n_1, ..., n_d]$ in the legend corresponds to a network architecture with depth $d$ and $n_i$ neurons in each layer. Summary results are grouped by network depth, $d$.}}
        \label{fig:halfCheetah}
    \end{figure}

\subsection{Evolution of the Density of Linear Regions during Training}


We begin by examining the number of transitions encountered as we sweep along the fixed final-policy trajectory, $R_T(\tau^{*})$, as shown in Figure~\ref{fig:trans-fixed}. For clarity, we present results for only three policy network configurations. This plot shows a moderate and gradual increase of 50\% in the number of region transitions observed in fixed-duration RL episodes during training, with larger policy networks having more transitions. Computing the region density requires dividing by number of neurons, $N$, and the trajectory length, $L(\tau^{*})$, which is constant for the fixed final trajectory, $\tau^{*}$. The result is given in Figure~\ref{fig:density-fixed} which shows the mean and standard deviation of the normalized region densities across policy networks with the same depth. This shows a result that is consistent across all policy network dimensions to within $\pm 30$\%, and reflects an identical moderate-and-gradual increase during training, regardless of the policy network configuration. Thus, in response to {\bf Q2}, RL policies trained with PPO see a moderate increase in region density along the parts of the state space that are frequently visited by the policy. We later discuss the observed impact of policy-network depth (\S\ref{sec:impact-expressivity}).


We next examine the number of transitions encountered as we sweep along the evolving set of current-policy trajectories, $R_T(\tau)$, shown in Figure~\ref{fig:trans-current}. These begin near zero and then rapidly increase as the policy succeeds at exploration. Again, for clarity, we show a representative sample of three policy network dimensions here. The corresponding region densities are shown in Figure~\ref{fig:density-both}, together with the fixed-region densities which are repeated from Figure~\ref{fig:density-fixed} (see the bottom dashed curve). We speculate that the observed densities are higher early on due to the unstructured nature of early exploration and the form of network initialization, and then decrease over time as the trajectory lengths begin to increase, as shown in Figure~\ref{fig:length-curr}.  


We compute the number of repeat visits to regions as $R_T - R_U$, i.e., the region transition count minus the unique region count. In Figure~\ref{fig:repeat-visits-current}, we track this for the current trajectory during training. We speculate that repeat visits are high early in training because of limited exploration. Repeat visits remain later in training because of the cyclic nature of the learned locomotion for HalfCheetah.

\subsection{Impact of Policy Network Structure on Expressivity
\label{sec:impact-expressivity}
 }

Our normalized density computation assumes that the number of transitions scales linearly with the number of neurons, $N$.  However, we empirically observe that there remains a moderate dependence on the network depth, as seen in Figure~\ref{fig:density-fixed}, with deeper networks resulting in moderately denser regions in the learned policies. 
We further examine the policy network expressivity using random infinite lines. Following \citet{hanin2019complexity}, for each policy network, we sample 100 random lines by randomly sampling 100 points from the final trajectory $\tau^*$, and connecting each randomly sampled point to the origin of the policy input space. Figure~\ref{fig:density-random-line-origin} shows the mean and standard deviation of the normalized number of transitions, $R_T/N$, as computed along these infinite lines in random directions passing through the origin, over the 5 random seeds of each network configuration. These produce approximately uniform results throughout training, showing no overall changes in mean density in various directions. These results are also consistent with the findings of \citet{hanin2019complexity} as they show normalized region densities are close to 1 at initialization and remain roughly constant during training. We repeated the same experiment with 100 random lines through the mean point instead of the origin, by connecting each randomly sampled point to the mean of the 100 sampled points. The difference between the density over lines through the origin and through the mean point is negligible.

To further examine the behavior of the policy network on areas that it does not focus on exploring during training, we compute the region density along random trajectories created by taking random actions in the environment. For each policy network, we sample a set of 10 such random trajectories $\tau^R$ and compute the normalized region density over these. Figure~\ref{fig:density-random-traj} shows the mean and standard deviation of this metric across this set of 10 random trajectories, over the 5 random seeds of each network configuration. This shows a very moderate decrease during training, with fairly consistent densities for all policy network configurations. We speculate that the slight decrease in density over time arises from the trained policy no longer visiting the same regions as the random-action policy. Comparing the results of random trajectories with fixed and current (Figure~\ref{fig:density-fixed} and Figure~\ref{fig:density-both}),  we can see that the normalized region density over random-action trajectories and a fixed final trajectory are within the same range. Also, although this metric is higher over current trajectories early on during training, it eventually decreases and converges to the same range of region densities as the fixed trajectory. 
In Appendix~\ref{appendix:Decision Regions viewed via Embedding Planes} we further directly visualize the regions of a policy network for HalfCheetah, as viewed via an embedding plane that either passes through three points sampled from the final fixed trajectory, or three points sampled from a random-action trajectory. We do not observe significant differences between these two visualizations.

\subsection{Additional Experiments}
To further study the generalization of our observations, we perform four additional experiments.

\subsubsection{RL Algorithm}
To test how our findings generalize to RL algorithms other than PPO, we repeat our experiments by training deep RL policies with stochastic actor-critic (SAC) \citep{haarnoja2018soft} algorithm. The experimental setup for SAC and its results are documented in detail in Appendix~\ref{appendix:Comparison of PPO and SAC}. Region densities observed early on during training are quite different for SAC than they are for PPO, exhibiting high observed densities which then rapidly drop, before rising again like the PPO case. We hypothesize this is due to a combination of (i) the entropy bonus for SAC, which is then annealed away, and (ii) the different network initialization used for the baseline SAC and PPO implementations. Despite this initial difference, the evolution pattern of densities are consistent with the PPO results later on during the course of training. Another observation is that similar to PPO results, densities over fixed, current and random-action trajectories are within the same range. This supports our surprising finding that the complexity of RL policies is not principally captured by increased density on-and-around the optimal trajectories. 

\subsubsection{Environment}
We examine whether a non-cyclic task, such as LunarLander, exhibits different linear-region evolution behavior compared to our four default tasks, which are naturally biased towards cyclic locomotion. From the results which are available in Appendix~\ref{appendix:Decision Regions for a Non-Cyclic Task: LunarLander}, we can see that the transition counts and densities along the fixed trajectory are similar in structure to those of the cyclic locomotion tasks, showing that a gradual increase in density appears to be a general property for the PPO setting, even for non-cyclic tasks.

\subsubsection{Value Network Analysis}
To study the difference between the value space and policy space, we repeat our experiments on the value networks trained on HalfCheetah with PPO. Full set of results for this experiment are available in Appendix~\ref{appendix:Value Network Analysis}. Our results show that linear regions in the value functions largely evolve in a very similar way to those in the policy.

\subsubsection{Behavior Cloning}
Deep RL provides us with a grounded setting to explore the impact of non-IID data on decision regions. To test if non-IID setting makes a difference, we repeat our experiments by training networks with behavior cloning (BC) using expert data from previously trained policies with PPO on HalfCheetah. The experimental setup for BC and the full results are documented in detail in Appendix~\ref{appendix:Behavior Cloning}. These results show that the evolution of region densities is different for BC than they are for PPO as the general increase in densities is not observed. In addition, number of transitions, the density values and length of current trajectories are significantly smaller for BC-trained policies. We have two hypotheses for these differences: (i) Policy's learning history plays a significant role in the evolution of regions, as early trajectories inform the template cell divisions that later evolve with further training. Since BC policies observe the entire state space from the moment training starts, they may be able to better divide their state space and avoid adding too much granularity to certain areas. (ii) Network initialization largely affects the resulting linear regions and their evolution during training.

\section{Conclusions}
\label{sec:conclusions}

The structure of ReLU-based reinforcement learning policies remains poorly understood. In this work, we explored a key property of such policies, the density of linear regions observed along policy trajectories, as a proxy for the complexity of functions they can learn. We performed a set of experiments to see how this evolves during training and whether and how changes in depth, and width of the policy network affect it. The hypothesis that the final policy trajectory sees an increase in region density is supported to a limited degree, as we observe a moderate increase in density during training, i.e., 50\%. This result is consistent across tasks and across policy networks of different configurations. When observed from the vantage point of the evolving current trajectory, the region density is high to begin with, due to the limited length of early trajectories, and then decreases asymptotically towards the observed final density. For the tasks we consider, trajectory lengths are observed to grow and asymptotically approach a plateau during training. Overall, the surprising implication, for our experimental setting, is that the complexity of RL policies appears to be captured by fine-tuning during learning, rather than a significant growth in the complexity of the functions observed along trajectories of the final policy.  

The key properties of linear regions as demonstrated for supervised learning of image classifiers, e.g., \citep{hanin2019complexity}, are shown to generally also hold true for reinforcement learning policies. In particular, their expressivity is well modeled as being proportional to the number of neurons, and largely independent of their width and depth, for ranges of these parameters where good policies can still be learned. The number of regions observed for randomly oriented line segments through the origin of the input space is also in line with earlier theoretical and empirical results, and does not change significantly during training. On a different note, properties of linear regions as measured along policy trajectories observed for policies trained with RL, do not hold for policies trained with BC. This emphasizes the unique behavior of RL compared to its supervised learning counterpart.     

Our work comes with a number of limitations. Our experiments are mainly focused on locomotion-based tasks which produce cyclic trajectories. This could mean that the results may not generalize to other types of tasks. For example, some tasks may not result in transition counts that grow, as generally observed in our experiments. In addition, our work studies deep RL policies trained with PPO and SAC. Thus, a broader analysis on more RL algorithms would shed light on the generality of the observations in this work. Since network initialization largely affects the linear regions, it would be ideal to use the same initialization when comparing different RL algorithms. However, RL algorithms are typically sensitive to network initialization, and changes in common practices in their initialization results in instabilities during training. Currently, we compute only the mean region density for a trajectory; it may be useful to further understand the distribution of observed densities across trajectories. We note that it is common practice in RL to apply input (observation) normalization based on a rolling average of observations seen so far~\citep{andrychowicz2020matters}, and thus our results also employ this strategy. Applying RL without this normalization may impact both the performance of the RL algorithm and the resulting evolution of transition counts and region-density. 

Our work suggests opportunities for improving our understanding of deep reinforcement learning through studying how states are mapped to actions and how these mappings evolve during training. For instance, a number of questions for future work includes understanding the behavior of observed linear regions in the contexts of other RL algorithms, and behavior cloning. Moreover, we envisage various future directions related to this work, including: (i) understanding the impact of non-IID data on decision regions more generally; (ii) consideration of RL algorithms that have less “learning path dependence”, e.g., via frequent (re)distillation onto randomly reinitialized policies; (iii) using visited regions to efficiently distill more interpretable non-parametric policies;  and (iv) in real-world control settings, using knowledge of decision regions, their sizes, and their orderings to better understand exploration and safety issues. Lastly, we believe that exploring possible connections with the lottery-ticket hypothesis~\citep{frankle2018lottery} will be worthwhile. 

As foundational research related to reinforcement learning, this work could be used to develop improved reinforcement learning policy structures and algorithms, with many possible applications, both good and bad in terms of societal impact.

\section*{Acknowledgments} 
This work was supported by the NSERC grant RGPIN-2020-05929. David Rolnick acknowledges support from the Canada CIFAR AI Chairs program. This research was enabled in part by technical support and computational resources provided by Compute Canada (\url{www.computecanada.ca}).

\newpage
\bibliography{references}

\newpage
\section*{Checklist}


\begin{enumerate}

\item For all authors...
\begin{enumerate}
  \item Do the main claims made in the abstract and introduction accurately reflect the paper's contributions and scope?
    \answerYes{}
  \item Did you describe the limitations of your work?
    \answerYes{See Section~\ref{sec:conclusions}.}
  \item Did you discuss any potential negative societal impacts of your work?
    \answerYes{See Section~\ref{sec:conclusions}.}
  \item Have you read the ethics review guidelines and ensured that your paper conforms to them?
    \answerYes{}
\end{enumerate}

\item If you are including theoretical results...
\begin{enumerate}
  \item Did you state the full set of assumptions of all theoretical results?
    \answerNA{}
        \item Did you include complete proofs of all theoretical results?
    \answerNA{}
\end{enumerate}

\item If you ran experiments...
\begin{enumerate}
  \item Did you include the code, data, and instructions needed to reproduce the main experimental results (either in the supplemental material or as a URL)?
    \answerYes{Code is available at: \url{https://github.com/setarehc/deep_rl_regions}.}
  \item Did you specify all the training details (e.g., data splits, hyperparameters, how they were chosen)?
    \answerYes{See Section~\ref{sec:results}.}
        \item Did you report error bars (e.g., with respect to the random seed after running experiments multiple times)?
    \answerYes{We show mean/std bars. See Section~\ref{sec:results}.}
        \item Did you include the total amount of compute and the type of resources used (e.g., type of GPUs, internal cluster, or cloud provider)?
    \answerNo{Our experiments require modest compute resources.}
\end{enumerate}

\item If you are using existing assets (e.g., code, data, models) or curating/releasing new assets...
\begin{enumerate}
  \item If your work uses existing assets, did you cite the creators?
    \answerNA{}
  \item Did you mention the license of the assets?
    \answerNA{}
  \item Did you include any new assets either in the supplemental material or as a URL?
    \answerNA{}
  \item Did you discuss whether and how consent was obtained from people whose data you're using/curating?
    \answerNA{}
  \item Did you discuss whether the data you are using/curating contains personally identifiable information or offensive content?
    \answerNA{}
\end{enumerate}

\item If you used crowdsourcing or conducted research with human subjects...
\begin{enumerate}
  \item Did you include the full text of instructions given to participants and screenshots, if applicable?
    \answerNA{}
  \item Did you describe any potential participant risks, with links to Institutional Review Board (IRB) approvals, if applicable?
    \answerNA{}
  \item Did you include the estimated hourly wage paid to participants and the total amount spent on participant compensation?
    \answerNA{}
\end{enumerate}

\end{enumerate}

\appendix
\newpage
{\Large Appendix}

\section{Non-Linearity Definitions}\label{appendix:Non-Linearity Definitions}
The following activation functions are discussed or used in this work:
    \begin{enumerate}
        \item ReLU \citep{nair2010rectified}: $\text{max}(x, 0)$
        \item Tanh (Hyperbolic tangent): $(e^x - e^{-x}) / (e^x + e^{-x})$
    \end{enumerate}

\section{Experimental Setup}\label{appendix:Experimental Setup}
All experiments are implemented in Pytorch \citep{paszke2019pytorch}, using the RL codebase from Stable Baselines3 \citep{raffin2021stable}. We conduct our analysis on four continuous control tasks from the OpenAI gym benchmark suits \citep{brockman2016openai} including HalfCheetah-v2, Walker-v2, Ant-v2, and Swimmer-v2 environments, using PPO algorithm \citep{schulman2017proximal}. All of our experiments are performed over 5 trials with distinct random seeds and the metric plots are shown as the mean and standard deviation across these seeds.

All policy networks configurations are chosen such that the network is fully-capable of learning the task and achieve near state-of-the-art cumulative rewards on the particular task it is trained on. Except where specified, network initialization and hyperparameters are set to the defaults of the PPO implementation of Stable Baselines3. We train our policy networks on 2M samples (i.e. 2M timesteps in the environment). Table~\ref{tbl:performance} shows the range of the return per environment, across all network configurations and random seeds. Return is measured as the mean trajectory cumulative reward across the batch collected at each epoch.
    
    \begin{table}[H]
        \caption{Performance Metrics}
        \label{tbl:performance}
        \centering
            \begin{tabular}{lcccc}
            \toprule
            1M Step Return & Average & Max & Min & Std \\
            \midrule
            HalfCheetah & $3920.5$ & $5035.7$ & $2576.1$ & $423.2$\\
            Walker & $1605.9$ & $4527.7$ & $135.5$ & $1305.6$\\
            Ant & $1898.1$ & $2609.8$ & $1255.9$ & $337.2$ \\
            Swimmer & $82.3$ & $128.8$ & $30.8$ & $28.8$  \\
            \midrule
            2M Step Return & Average & Max & Min & Std \\
            \midrule
            HalfCheetah & $4876.6$ & $6679.4$ & $3406.3$ & $494.3$\\
            Walker & $3618.4$ & $6066.1$ & $683.3$ & $1610$\\
            Ant & $2626.6$ & $4543.5$ & $1535.3$ & $672$\\
            Swimmer & $112.8$ & $132.7$ & $41.5$ & $19.7$  \\
            \bottomrule
            \end{tabular}
    \end{table}

\section{Network Dimensions}\label{appendix:Network Dimensions}
In all experiments, the policy and value function network architectures used are MLPs with ReLU activations in all hidden layers. Except where specified, the value function has $N=128$ neurons with layer widths $64, 64$ while the policy network has 18 structures listed in Table~\ref{tbl:dimensions}.

    \begin{table}[H]
        \caption{Network architectures}
        \label{tbl:dimensions}
        \centering
            \begin{tabular}{l ccc}
            \toprule
                Neurons  & \multicolumn{3}{c}{Layer Widths} \\
                \midrule
                N=32  & 32  & 16,16  & 8,8,8,8 \\
                N=64  & 64  & 32,32  & 16,16,16,16 \\
                N=128 & 128 & 64,64  & 32,32,32,32 \\
                N=48  & 48  & 24,24  & 16,16,16 \\
                N=96  & 96  & 48,48  & 32,32,32 \\
                N=192 & 192 & 96,96  & 64,64,64 \\
            \bottomrule
            \end{tabular}
    \end{table}

\section{Plots and Error Bars}\label{appendix:Plots and Error Bars}
Each policy network architecture is trained 5 times with 5 random seeds. All metric plots in this work are shown as the mean and standard deviation across these seeds. For the state space visualization of Figure~\ref{fig:2d_visualization}, the sampling grid in input space is obtained by sampling a $300 \times 300$ grid with equally distanced points on a window of size $40 \times 200$ where $x \in [-50, 150]$ and $\dot{x} \in [-20, 20]$. Next, we feed each point of the sampling grid into the policy network to compute its activation pattern. Then, we group all points with the same activation pattern into a single linear region uniquely colored according to its activation pattern. Note that the way of encoding a linear region of a point as described in Section~\ref{sec:counting} guarantees that each region will be uniquely identified by an activation pattern. We create a color map for regions based on their activation patterns. For the state space visualization of Figure~\ref{fig:hd_visualization}, we plot regions within a 2D plane that intersects the 17-dimensional input space of HalfCheetah-v2. To obtain the 2D plane, we sample three random points from a trajectory (final trajectory $\tau^*$ or random-action trajectory $\tau^R$) and find the plane that passes through these points. We consider all the points within a square centered at the circumcenter of the three sampled points. Square is scaled such that it is slightly bigger than the circle. In order to avoid missing small regions, which are dominant in higher dimensional spaces, we do not use the subsampling method used for the visualization of the 2D environment. Instead, we compute the exact locations of the boundaries of the regions using our counting method describe in Section~\ref{sec:counting}. Each region is then colored using a randomly sampled color.

\section{Initialization and Hyperparameters}\label{appendix:Initialization and Hyperparameters}
The initialization of the policy network has a high impact on the training performance with PPO \citep{andrychowicz2020matters}. To be able to focus on the evolution of linear regions disentangled from properties such as trainability, which is outside the scope of this work, we use the default initialization of PPO from Stable Baselines3. This initialization scheme is consistent with the best-practice recommendations for on-policy RL \citep{andrychowicz2020matters}, but it differs from that used in \citet{hanin2019complexity}. In this initialization, for both the policy and value function networks, weights of the hidden layers use orthogonal initialization with scaling $\sqrt{2}$, and the biases are set to $0$. The weights of the policy network's output layer are initialized with the scale of $0.01$, and the weights of the value function network's output layer are initialized with the scale of $1$. In \citet{hanin2019complexity}, networks are initialized with He normalization and biases are drawn i.i.d from a normal distribution centered at $0$ with variance $10^{-6}$.

We mainly adopt the default hyperparameters of PPO from the implementation of Stable Baselines3 in all of our experiments on PPO. We present the details of the choices of hyperparameters in Table~\ref{tbl:hyperparams}.

    \begin{table}[H]
        \caption{Hyperparameters}
        \label{tbl:hyperparams}
        \centering
            \begin{tabular}{l ccc}
            \toprule
                Hyperparameter  & Value \\
                \midrule
                Training steps  & $2 \times 10^6$ \\
                Learning rate & $3 \times 10^{-4}$ \\
                Number of epochs  & $125$ \\
                Minibatch size  & $64$ \\
                Discount ($\gamma$) & $0.99$ \\
                GAE parameter ($\lambda$) & $0.95$\\
                Init. log stdev. & -1.0\\
                Clipping parameter ($\epsilon$) & $0.2$\\
                VF coeff. ($c_1$) & $0.5$\\
                Entropy coeff. ($c_2$) & $0.0$\\
                Hardware & CPU \\
            \bottomrule
            \end{tabular}
    \end{table}

\section{Comparison of PPO and SAC}\label{appendix:Comparison of PPO and SAC}
To shed light on the effect of the choice of RL algorithm, we repeat our experiments with the Soft Actor Critic (SAC) algorithm \citep{haarnoja2018soft}. Similar to our main experimental setup for PPO, we use the Stable-Baselines3 implementation of the SAC algorithm and conduct our experiments on the four continuous control tasks of HalfCheetah-v2, Walker-v2, Ant-v2, and Swimmer-v2 from the OpenAI gym benchmark suits. Again, we run each experiment with 5 different random seeds and the results show the mean and the standard deviation across the random seeds. The policy network architecture used is an MLP with ReLU activations in all hidden layers. To ensure full training capacity, we choose more parameterized networks in this setup. The value function has two hidden layers with $256$ neurons in each layer, while the policy network has 12 structures as listed in Table~\ref{tbl:sac-policy-dimensions}. We adopt the network initialization and hyperparameters of SAC from Stable-Baselines3 and train our policy network on 1M samples. Table~\ref{tbl:sac-performance} shows the range of the random return per environment, across all network configurations and random seeds. Return is measured as the mean trajectory return of ten trajectories collected by running the deterministic policy without action noise once every 8000 steps. We present the details of the choices of hyperparameters in Table~\ref{tbl:sac-hyperparams}.

\begin{figure}[H]
        \centering
        \begin{subfigure}[b]{0.45\textwidth}  
            \centering 
            \includegraphics[width=\textwidth]{figs/ppo/transitions-fixed-cheetah.pdf}
            \caption{{\small HalfCheetah - PPO}}    
            \label{fig:sac-compare-trans-fixed-ppo}
        \end{subfigure}
        \hfill
        \begin{subfigure}[b]{0.45\textwidth}   
            \centering 
            \includegraphics[width=\textwidth]{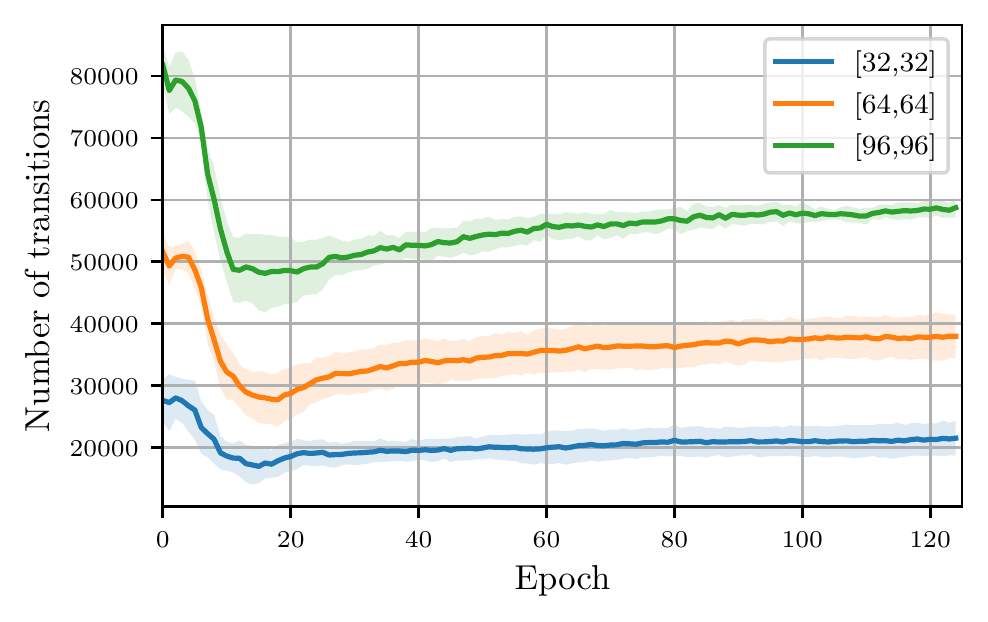}
            \caption{{\small HalfCheetah - SAC}}    
            \label{fig:sac-compare-trans-fixed-sac}
        \end{subfigure}
        \caption{Evolution of the number of transitions over a fixed trajectory sampled from the final fully trained policy ($\tau^*$) during training HalfCheetah with SAC and PPO algorithms. Plots show the mean and standard error across 5 random seeds. In the legend, $[n_1, ..., n_d]$ corresponds to a network architecture with depth $d$ and $n_i$ neurons in each layer. }
        \label{fig:sac-compare-trans-fixed}
    \end{figure}

\begin{figure}[H]
        \centering
        \begin{subfigure}[b]{0.45\textwidth}  
            \centering 
            \includegraphics[width=\textwidth]{figs/ppo/density-fixed-depth-cheetah.pdf}
            \caption{{\small HalfCheetah - PPO}}    
            \label{fig:sac-compare-density-fixed-ppo}
        \end{subfigure}
        \hfill
        \begin{subfigure}[b]{0.45\textwidth}   
            \centering 
            \includegraphics[width=\textwidth]{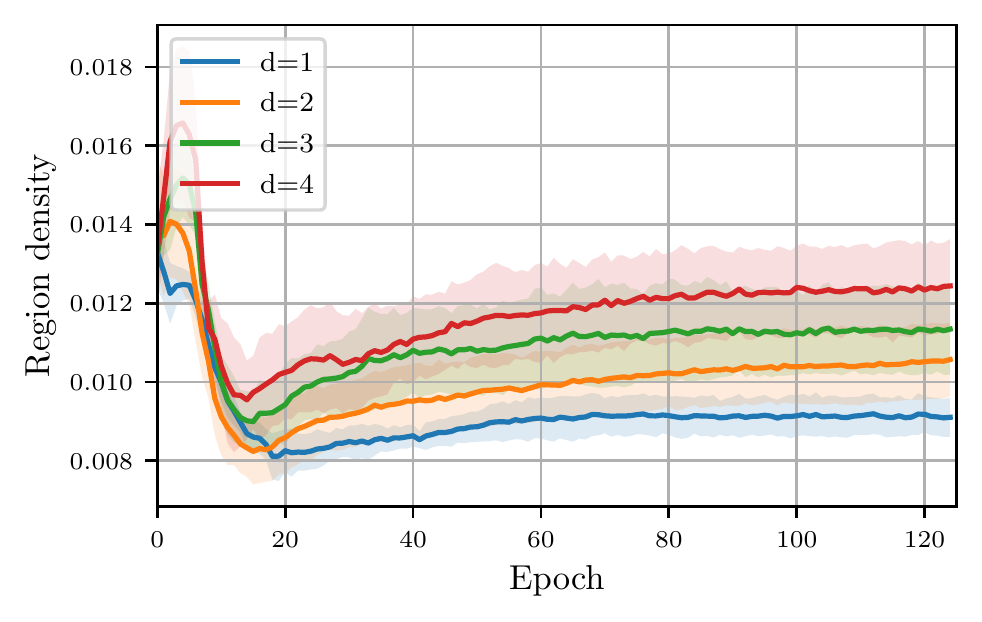}
            \caption{{\small HalfCheetah - SAC}}    
            \label{fig:sac-compare-density-fixed-sac}
        \end{subfigure}
    \caption{Evolution of the normalized region density over a fixed trajectory sampled from the final fully trained policy ($\tau^*$) during training HalfCheetah with SAC and PPO algorithms.}
        \label{fig:sac-compare-density-fixed}
    \end{figure}

\begin{figure}[H]
        \centering
        \begin{subfigure}[b]{0.45\textwidth}  
            \centering 
            \includegraphics[width=\textwidth]{figs/ppo/transitions-current-cheetah.pdf}
            \caption{{\small HalfCheetah - PPO}}    
            \label{fig:sac-compare-trans-current-ppo}
        \end{subfigure}
        \hfill
        \begin{subfigure}[b]{0.45\textwidth}   
            \centering 
            \includegraphics[width=\textwidth]{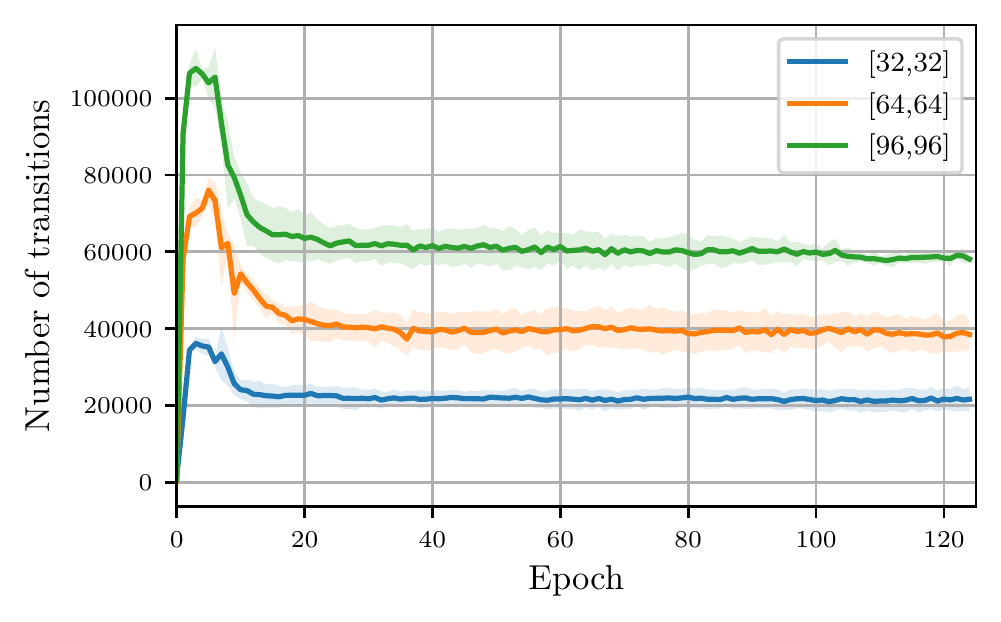}
            \caption{{\small HalfCheetah - SAC}}    
            \label{fig:sac-compare-trans-current-sac}
        \end{subfigure}
        \caption{Evolution of the number of transitions over trajectories sampled from the current snapshot of the policy ($\tau$) during training HalfCheetah with SAC and PPO algorithms.}
        \label{fig:sac-compare-trans-current}
    \end{figure}

\begin{figure}[H]
        \centering
        \begin{subfigure}[b]{0.45\textwidth}  
            \centering 
            \includegraphics[width=\textwidth]{figs/ppo/density-curr-fixed-cheetah.pdf}
            \caption{{\small HalfCheetah - PPO}}    
            \label{fig:sac-compare-density-both-ppo}
        \end{subfigure}
        \hfill
        \begin{subfigure}[b]{0.45\textwidth}   
            \centering 
            \includegraphics[width=\textwidth]{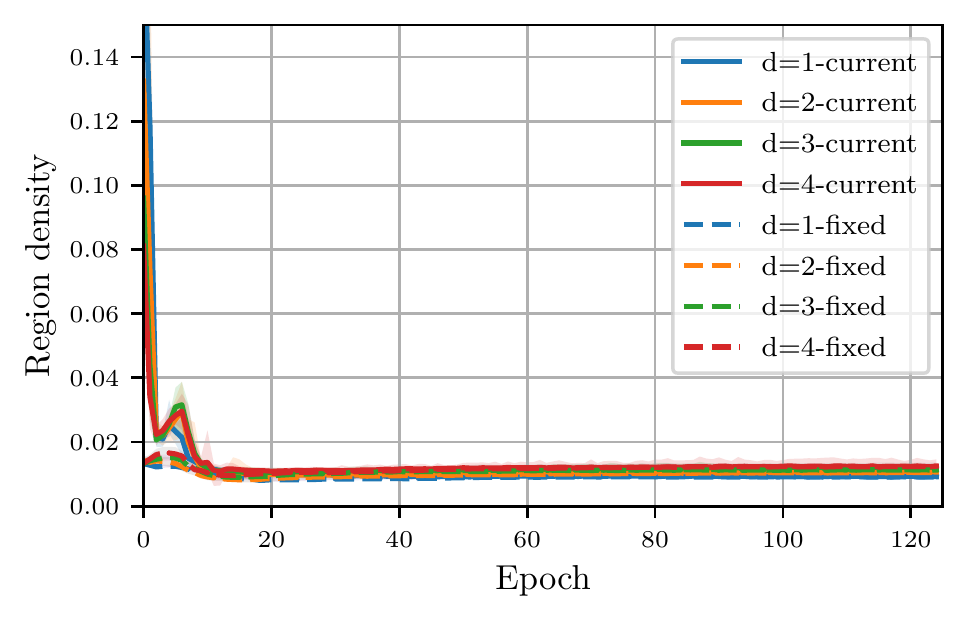}
            \caption{{\small HalfCheetah - SAC}}    
            \label{fig:sac-compare-density-both-sac}
        \end{subfigure}
        \caption{Evolution of the normalized region density over both a fixed trajectory sampled from the final fully trained policy ($\tau^*$) and current trajectories sampled from the current snapshot of the policy ($\tau$) during training HalfCheetah with SAC and PPO algorithms. }
        \label{fig:sac-compare-density-both}
    \end{figure}

\begin{figure}[H]
        \centering
        \begin{subfigure}[b]{0.45\textwidth}  
            \centering 
            \includegraphics[width=\textwidth]{figs/ppo/length-curr-cheetah.pdf}
            \caption{{\small HalfCheetah - PPO}}    
            \label{fig:sac-compare-length-curr-both-ppo}
        \end{subfigure}
        \hfill
        \begin{subfigure}[b]{0.45\textwidth}   
            \centering 
            \includegraphics[width=\textwidth]{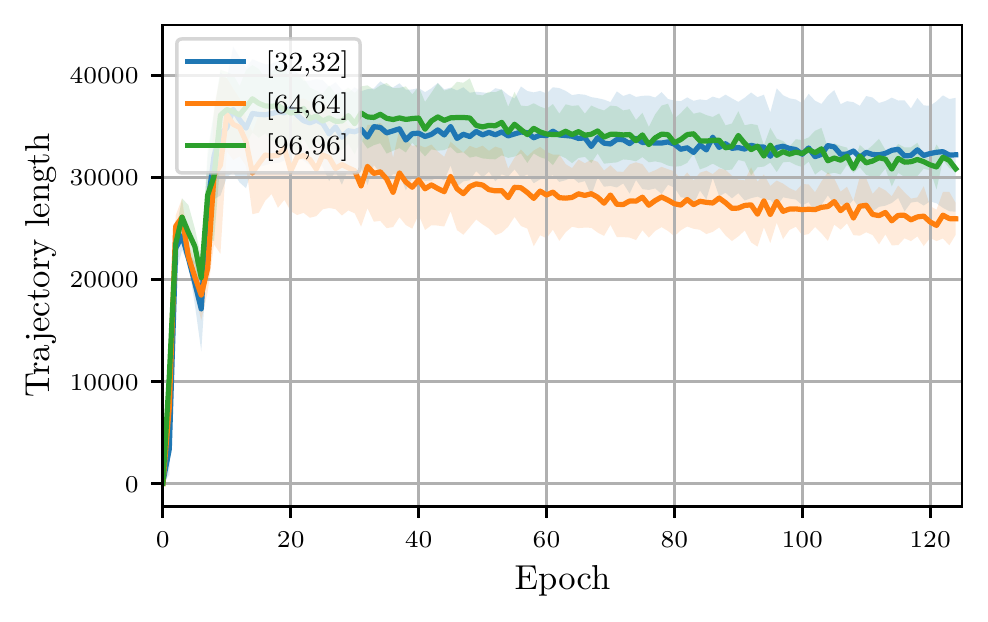}
            \caption{{\small HalfCheetah - SAC}}    
            \label{fig:sac-compare-length-curr-both-sac}
        \end{subfigure}
        \caption{Evolution of the length of trajectories sampled from the current snapshot of the policy ($\tau$) during training HalfCheetah with SAC and PPO algorithms.}
        \label{fig:sac-compare-length-curr-both}
    \end{figure}

\begin{figure}[H]
        \centering
        \begin{subfigure}[b]{0.45\textwidth}  
            \centering 
            \includegraphics[width=\textwidth]{figs/ppo/repeat-visits-current-cheetah.pdf}
            \caption{{\small HalfCheetah - PPO}}    
            \label{fig:sac-compare-repeat-visits-current-ppo}
        \end{subfigure}
        \hfill
        \begin{subfigure}[b]{0.45\textwidth}   
            \centering 
            \includegraphics[width=\textwidth]{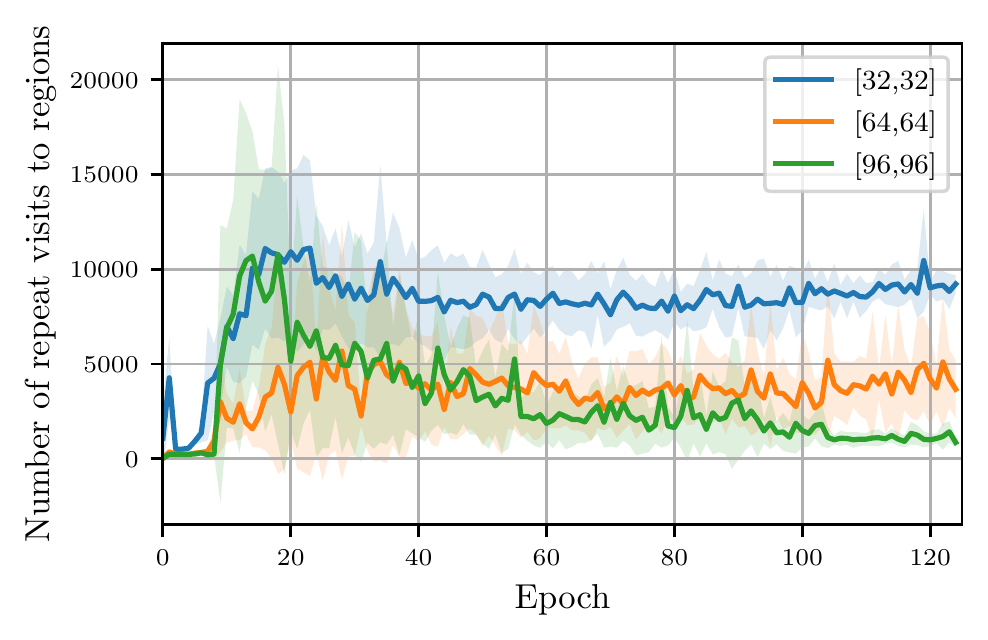}
            \caption{{\small HalfCheetah - SAC}}    
            \label{fig:sac-compare-repeat-visits-current-sac}
        \end{subfigure}
        \caption{Evolution of the number of repeat visits to regions over trajectories sampled from the current snapshot of the policy ($\tau$) during training HalfCheetah with SAC and PPO algorithms.} 
        \label{fig:sac-compare-repeat-visits-current}
    \end{figure}

\begin{figure}[H]
        \centering
        \begin{subfigure}[b]{0.45\textwidth}  
            \centering 
            \includegraphics[width=\textwidth]{figs/ppo/density-random-lines-origin-cheetah.pdf}
            \caption{{\small HalfCheetah - PPO}}    
            \label{fig:sac-compare-density-random-lines-origin-ppo}
        \end{subfigure}
        \hfill
        \begin{subfigure}[b]{0.45\textwidth}   
            \centering 
            \includegraphics[width=\textwidth]{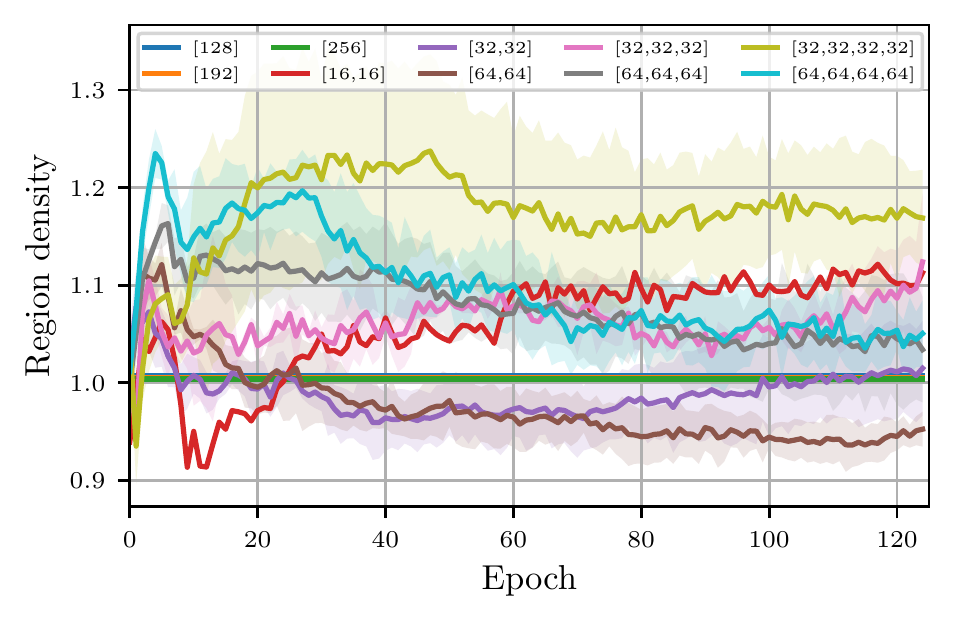}
            \caption{{\small HalfCheetah - SAC}}    
            \label{fig:sac-compare-density-random-lines-origin-sac}
        \end{subfigure}
        \caption{Evolution of the mean normalized region density over 100 random lines passing through the origin during training HalfCheetah with SAC and PPO algorithms. For each policy network configuration, we sample 100 random lines and compute the density of transitions as we sweep along these lines. We then report the mean density observed over these 100 lines.}
        \label{fig:sac-compare-density-random-lines-origin}
    \end{figure}

\begin{figure}[H]
        \centering
        \begin{subfigure}[b]{0.45\textwidth}  
            \centering 
            \includegraphics[width=\textwidth]{figs/ppo/density-random-traj-cheetah.pdf}
            \caption{{\small HalfCheetah - PPO}}    
            \label{fig:sac-compare-density-random-traj-ppo}
        \end{subfigure}
        \hfill
        \begin{subfigure}[b]{0.45\textwidth}   
            \centering 
            \includegraphics[width=\textwidth]{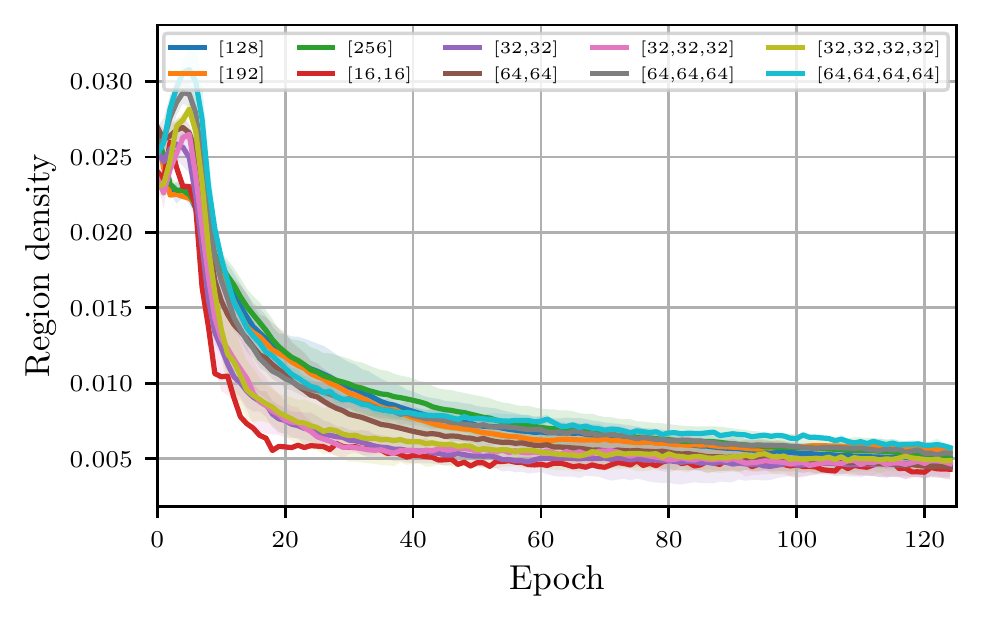}
            \caption{{\small HalfCheetah - SAC}}    
            \label{fig:sac-compare-density-random-traj-sac}
        \end{subfigure}
        \caption{Evolution of the mean normalized region density over 10 random-action trajectories ($\tau^R$) during training HalfCheetah with SAC and PPO algorithms. For each policy network configuration, we sample 10 random-action trajectories and compute the density of transitions as we sweep along these trajectories, and report the mean value of these trajectories.}
        \label{fig:sac-compare-density-random-traj}
    \end{figure}

Figures~\ref{fig:sac-compare-trans-fixed}-\ref{fig:sac-compare-density-random-traj} of this section, show the side-by-side comparison between the results of SAC trained on HalfCheetah and results of PPO trained on HalfCheetah previously shown in Figure~\ref{fig:halfCheetah}. 
We can see that the observed region densities during training are quite different for SAC than they are for PPO, early on in training. We hypothesize that the many transitions (and therefore corresponding high densities) observed early on in training for SAC is due to a combination of (i) the entropy bonus for SAC, which is then annealed away, and (ii) the different network initialization used for the baseline SAC and PPO implementations.

\begin{figure}[H]
        \centering
        \begin{subfigure}[b]{0.45\textwidth}  
            \centering 
            \includegraphics[width=\textwidth]{figs/sac/transitions-fixed-cheetah.pdf}
            \caption{{\small HalfCheetah - SAC}}    
            \label{fig:sac-trans-fixed-hc}
        \end{subfigure}
        \hfill
        \begin{subfigure}[b]{0.45\textwidth}   
            \centering 
            \includegraphics[width=\textwidth]{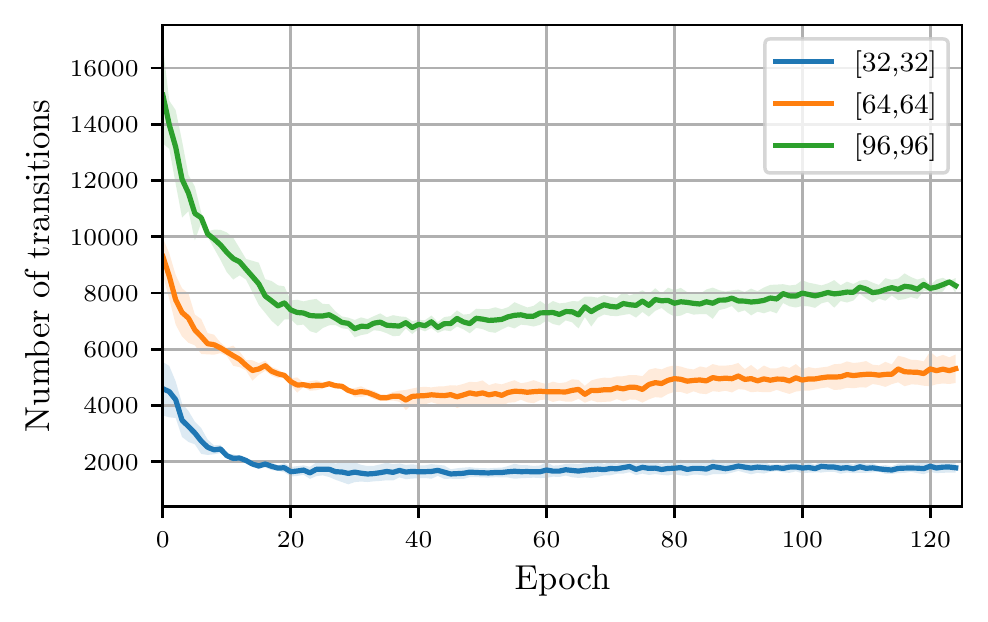}
            \caption{{\small Walker - SAC}}    
            \label{fig:sac-trans-fixed-w}
        \end{subfigure}
        \vskip\baselineskip
        \begin{subfigure}[b]{0.45\textwidth}
            \centering
            \includegraphics[width=\textwidth]{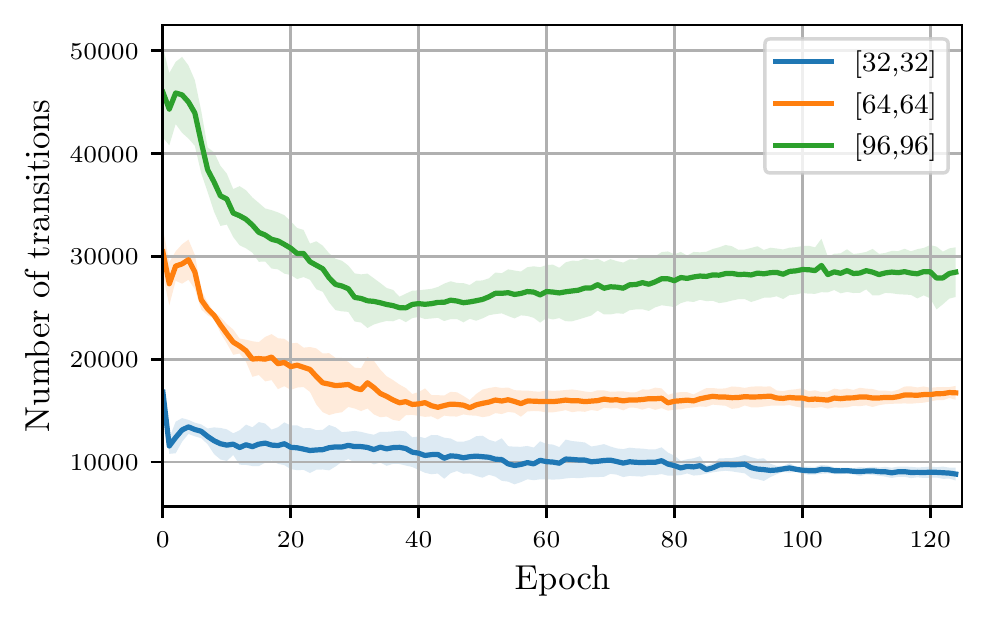}
            \caption{{\small Ant - SAC}}    
            \label{fig:sac-trans-fixed-a}
        \end{subfigure}
        \hfill        
        \begin{subfigure}[b]{0.45\textwidth}   
            \centering 
            \includegraphics[width=\textwidth]{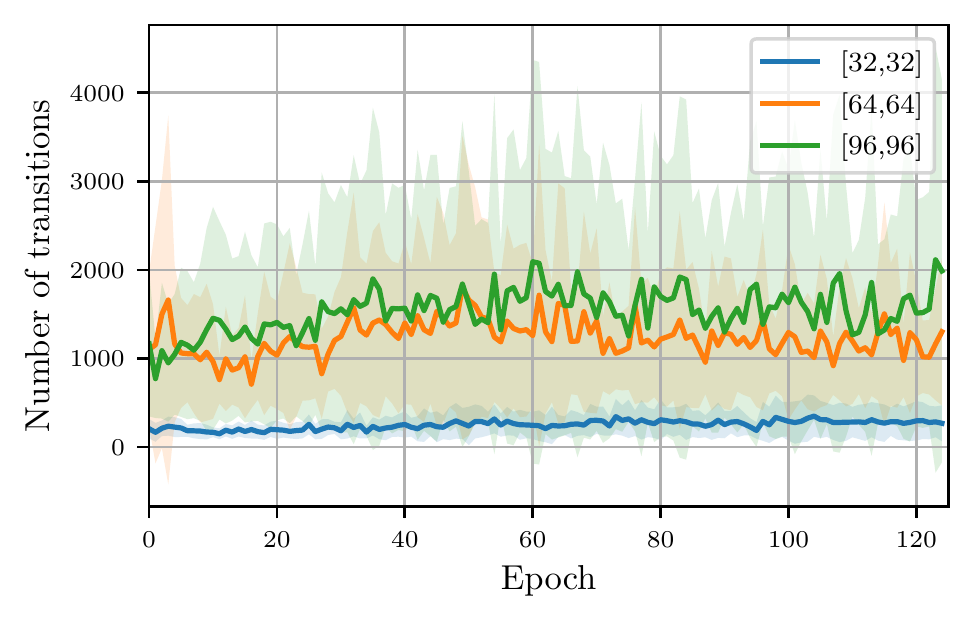}
            \caption{{\small Swimmer - SAC}}    
            \label{fig:sac-trans-fixed-s}
        \end{subfigure}
        \caption{Evolution of the number of transitions over a fixed trajectory sampled from the final fully trained policy ($\tau^*$) during training different tasks with SAC.}
        \label{fig:sac-trans-fixed}
    \end{figure}

\begin{figure}[H]
        \centering
        \begin{subfigure}[b]{0.45\textwidth}  
            \centering 
            \includegraphics[width=\textwidth]{figs/sac/density-fixed-depth-cheetah.pdf}
            \caption{{\small HalfCheetah - SAC}}    
            \label{fig:sac-density-fixed-hc}
        \end{subfigure}
        \hfill
        \begin{subfigure}[b]{0.45\textwidth}   
            \centering 
            \includegraphics[width=\textwidth]{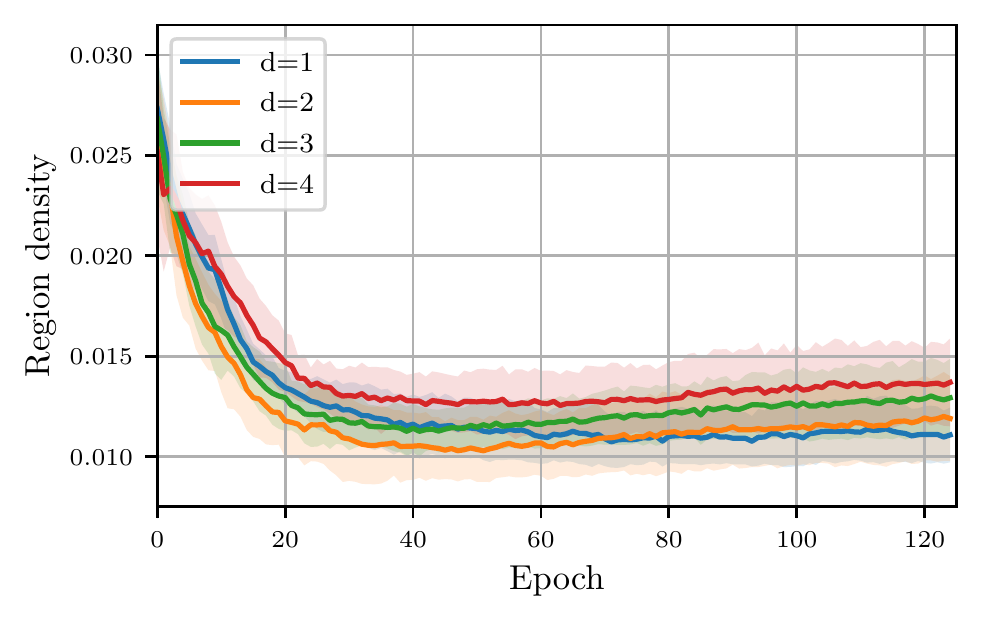}
            \caption{{\small Walker - SAC}}    
            \label{fig:sac-density-fixed-w}
        \end{subfigure}
        \vskip\baselineskip
        \begin{subfigure}[b]{0.45\textwidth}
            \centering
            \includegraphics[width=\textwidth]{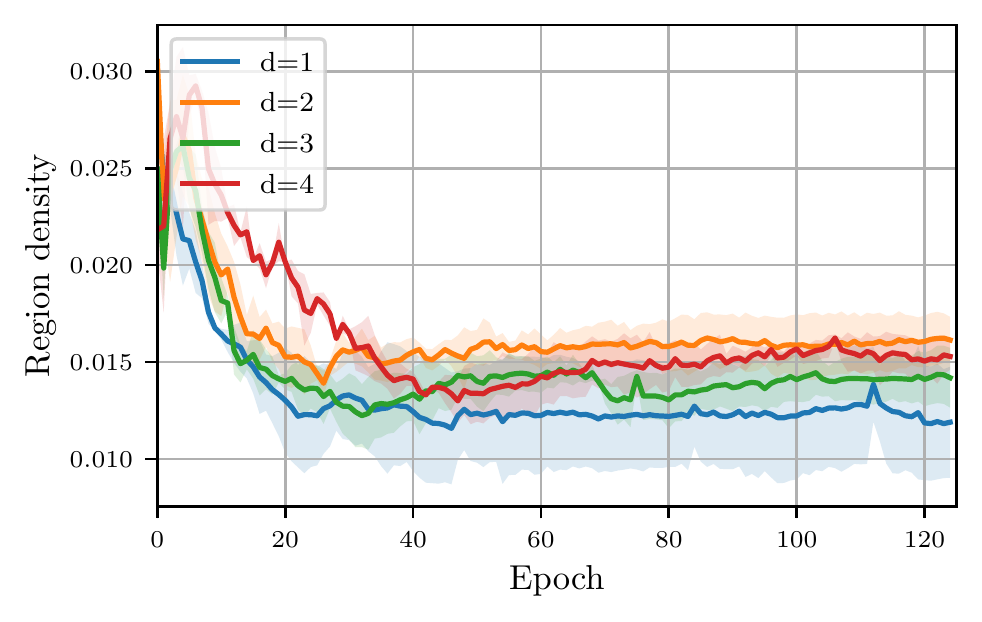}
            \caption{{\small Ant - SAC}}    
            \label{fig:sac-density-fixed-a}
        \end{subfigure}
        \hfill        
        \begin{subfigure}[b]{0.45\textwidth}   
            \centering 
            \includegraphics[width=\textwidth]{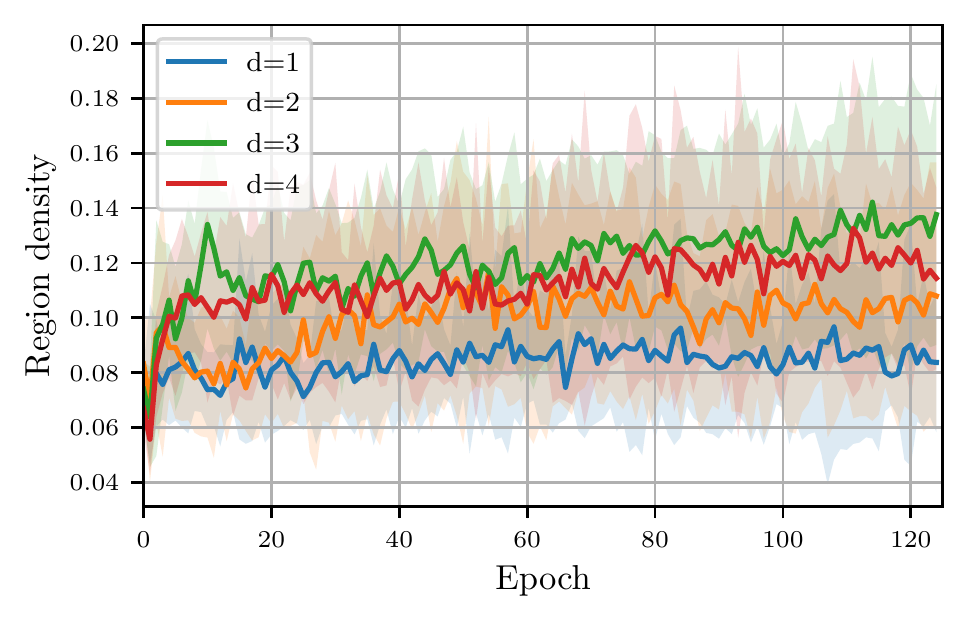}
            \caption{{\small Swimmer - SAC}}    
            \label{fig:sac-density-fixed-s}
        \end{subfigure}
        \caption{Evolution of the normalized region density over a fixed trajectory sampled from the final fully trained policy ($\tau^*$) during training different tasks with SAC. 
        }
        \label{fig:sac-density-fixed}
    \end{figure}

\begin{figure}[H]
        \centering
        \begin{subfigure}[b]{0.45\textwidth}  
            \centering 
            \includegraphics[width=\textwidth]{figs/sac/transitions-current-cheetah.pdf}
            \caption{{\small HalfCheetah - SAC}}    
            \label{fig:sac-trans-current-hc}
        \end{subfigure}
        \hfill
        \begin{subfigure}[b]{0.45\textwidth}   
            \centering 
            \includegraphics[width=\textwidth]{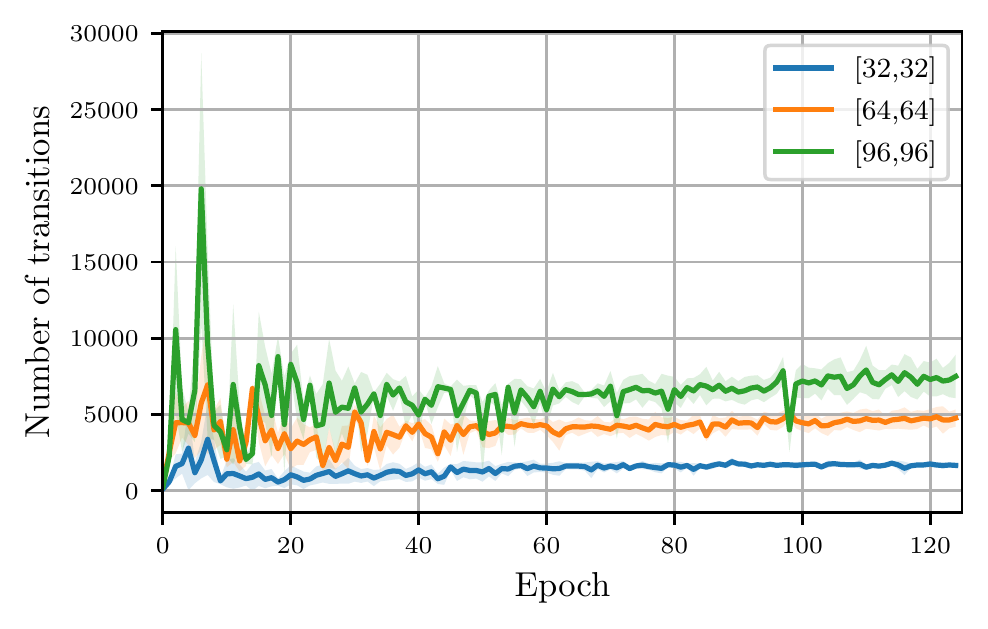}
            \caption{{\small Walker - SAC}}    
            \label{fig:sac-trans-current-w}
        \end{subfigure}
        \vskip\baselineskip
        \begin{subfigure}[b]{0.45\textwidth}
            \centering
            \includegraphics[width=\textwidth]{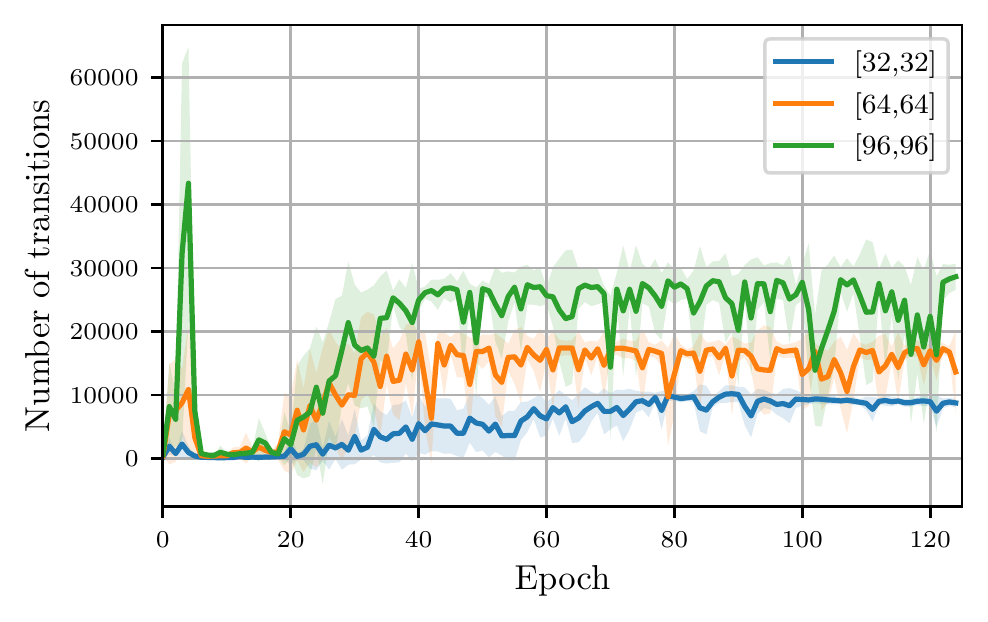}
            \caption{{\small Ant - SAC}}    
            \label{fig:sac-trans-current-a}
        \end{subfigure}
        \hfill        
        \begin{subfigure}[b]{0.45\textwidth}   
            \centering 
            \includegraphics[width=\textwidth]{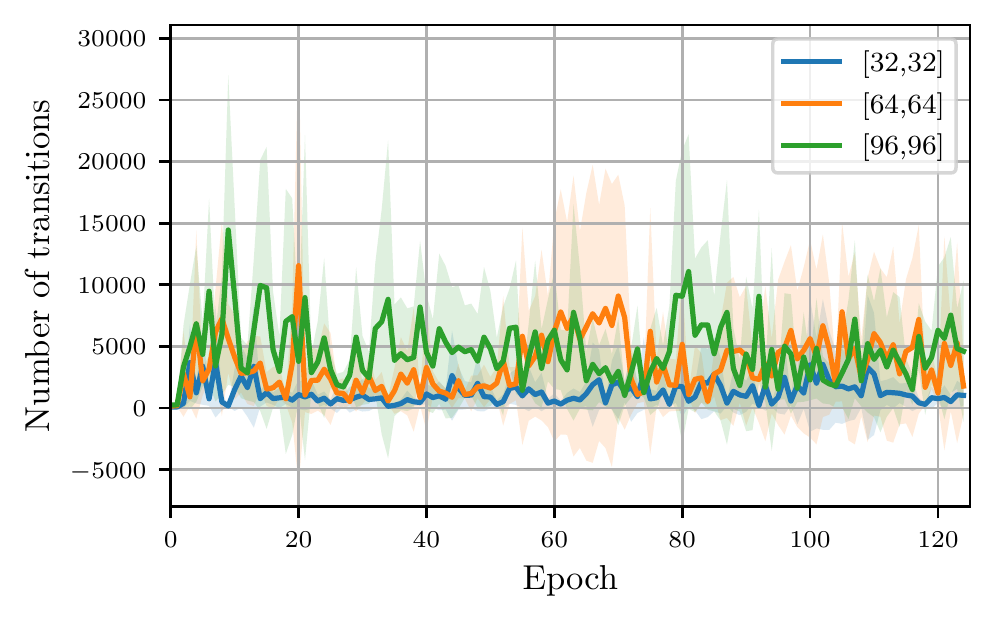}
            \caption{{\small Swimmer - SAC}}    
            \label{fig:sac-trans-current-s}
        \end{subfigure}
        \caption{Evolution of the number of transitions over trajectories sampled from the current snapshot of the policy ($\tau$) during training different tasks with SAC.
        }
        \label{fig:sac-trans-current}
    \end{figure}

\begin{figure}[H]
        \centering
        \begin{subfigure}[b]{0.45\textwidth}  
            \centering 
            \includegraphics[width=\textwidth]{figs/sac/density-curr-fixed-cheetah.pdf}
            \caption{{\small HalfCheetah - SAC}}    
            \label{fig:sac-density-both-hc}
        \end{subfigure}
        \hfill
        \begin{subfigure}[b]{0.45\textwidth}   
            \centering 
            \includegraphics[width=\textwidth]{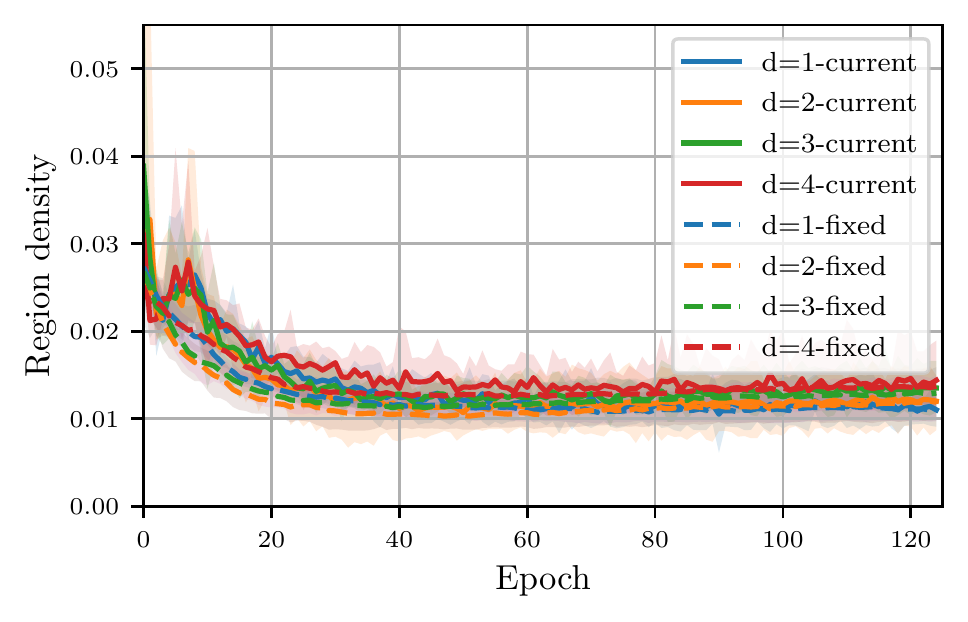}
            \caption{{\small Walker - SAC}}    
            \label{fig:sac-density-both-w}
        \end{subfigure}
        \vskip\baselineskip
        \begin{subfigure}[b]{0.45\textwidth}
            \centering
            \includegraphics[width=\textwidth]{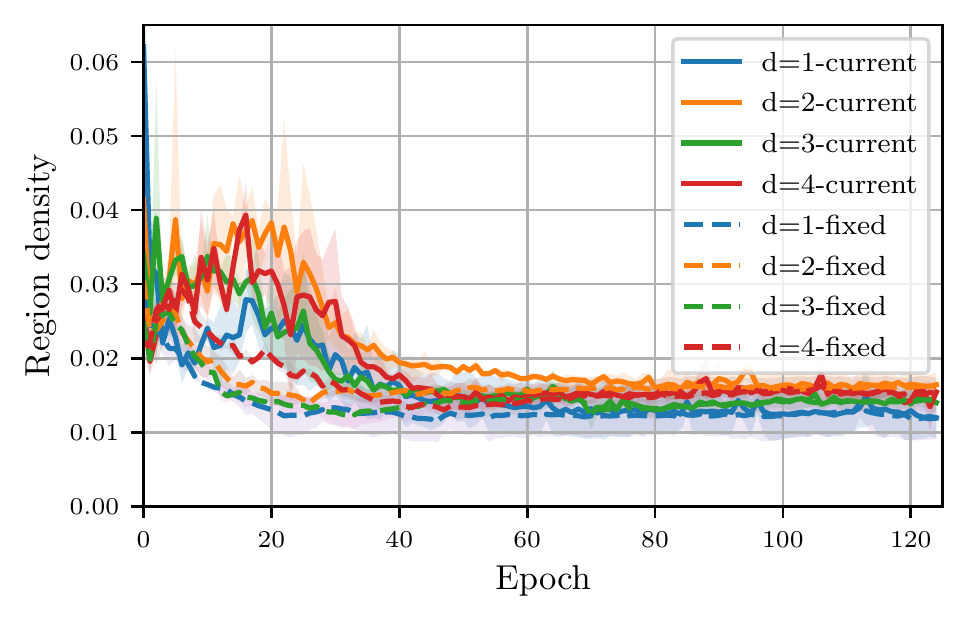}
            \caption{{\small Ant - SAC}}    
            \label{fig:sac-density-both-a}
        \end{subfigure}
        \hfill        
        \begin{subfigure}[b]{0.45\textwidth}   
            \centering 
            \includegraphics[width=\textwidth]{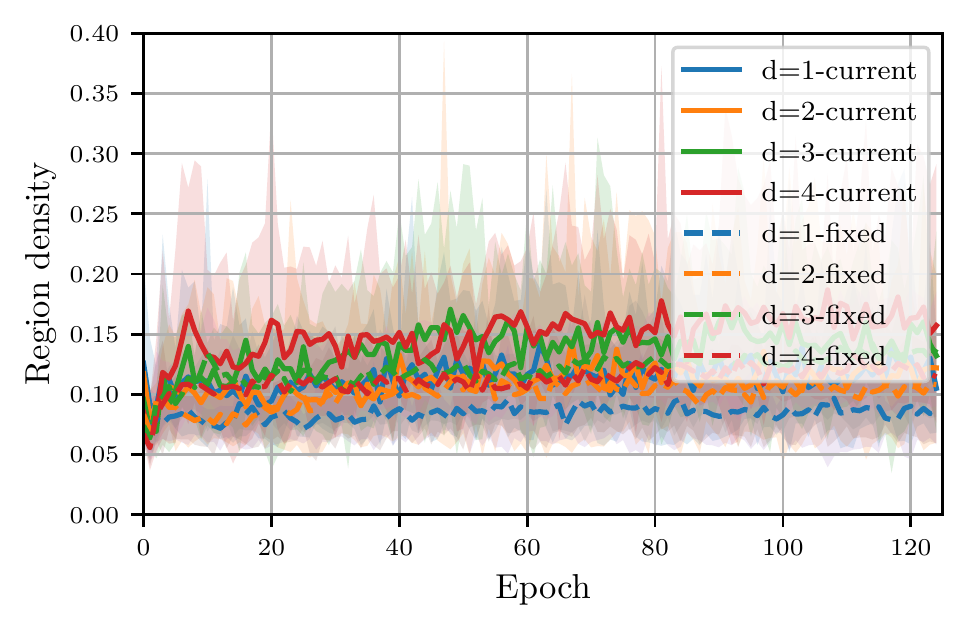}
            \caption{{\small Swimmer - SAC}}    
            \label{fig:sac-density-both-s}
        \end{subfigure}
        \caption{Evolution of the normalized region density over both a fixed trajectory sampled from the final fully trained policy ($\tau^*$) and current trajectories sampled from the current snapshot of the policy ($\tau$) during training different tasks with SAC. 
        }
        \label{fig:sac-density-both}
    \end{figure}

\begin{figure}[H]
        \centering
        \begin{subfigure}[b]{0.45\textwidth}  
            \centering 
            \includegraphics[width=\textwidth]{figs/sac/length-curr-cheetah.pdf}
            \caption{{\small HalfCheetah - SAC}}    
            \label{fig:sac-length-curr-hc}
        \end{subfigure}
        \hfill
        \begin{subfigure}[b]{0.45\textwidth}   
            \centering 
            \includegraphics[width=\textwidth]{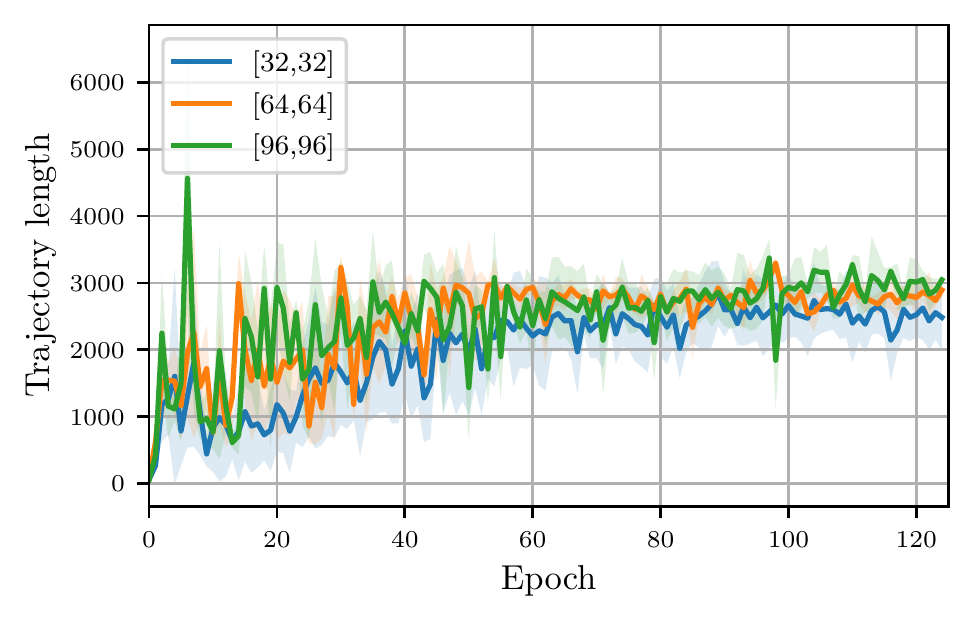}
            \caption{{\small Walker - SAC}}    
            \label{fig:sac-length-curr-w}
        \end{subfigure}
        \vskip\baselineskip
        \begin{subfigure}[b]{0.45\textwidth}
            \centering
            \includegraphics[width=\textwidth]{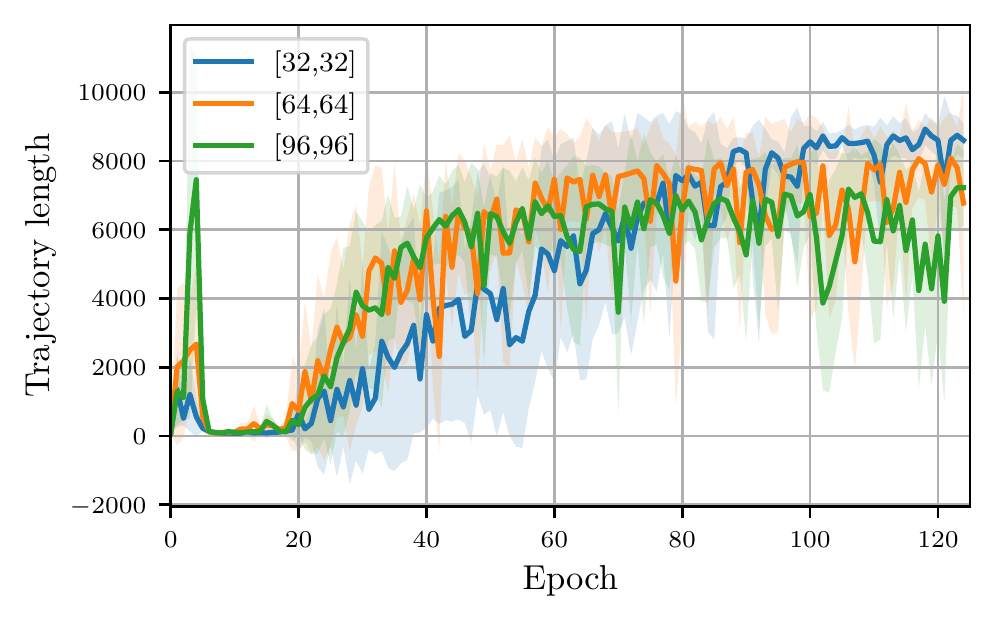}
            \caption{{\small Ant - SAC}}    
            \label{fig:sac-length-curr-a}
        \end{subfigure}
        \hfill        
        \begin{subfigure}[b]{0.45\textwidth}   
            \centering 
            \includegraphics[width=\textwidth]{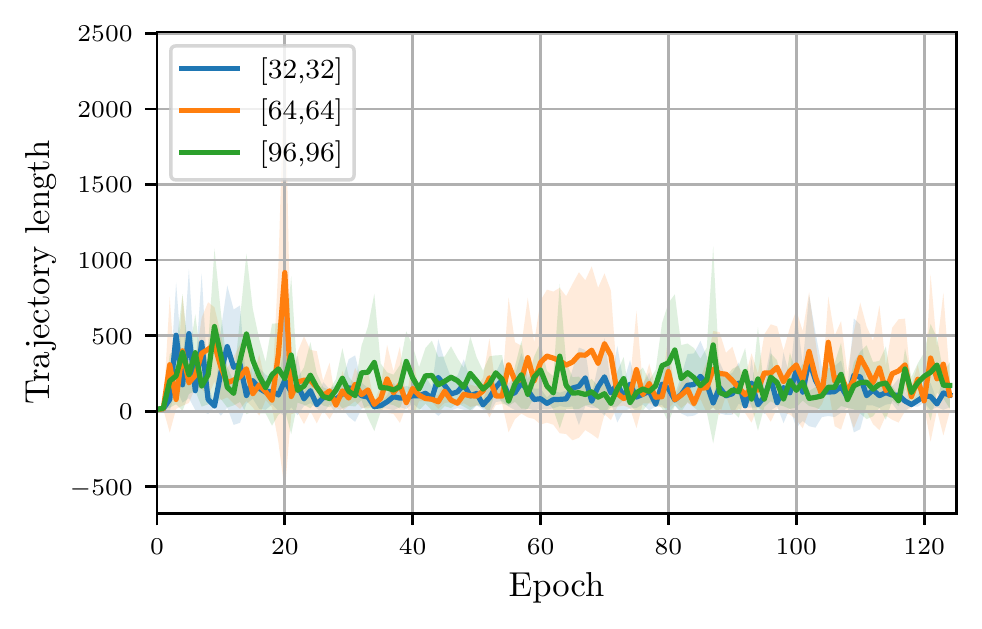}
            \caption{{\small Swimmer - SAC}}    
            \label{fig:sac-length-curr-s}
        \end{subfigure}
        \caption{Evolution of the length of trajectories sampled from the current snapshot of the policy ($\tau$) during training different tasks with SAC. 
        }
        \label{fig:sac-lenght-curr}
    \end{figure}

\begin{figure}[H]
        \centering
        \begin{subfigure}[b]{0.45\textwidth}  
            \centering 
            \includegraphics[width=\textwidth]{figs/sac/repeat-visits-current-cheetah.pdf}
            \caption{{\small HalfCheetah - SAC}}    
            \label{fig:sac-repeat-visits-current-hc}
        \end{subfigure}
        \hfill
        \begin{subfigure}[b]{0.45\textwidth}   
            \centering 
            \includegraphics[width=\textwidth]{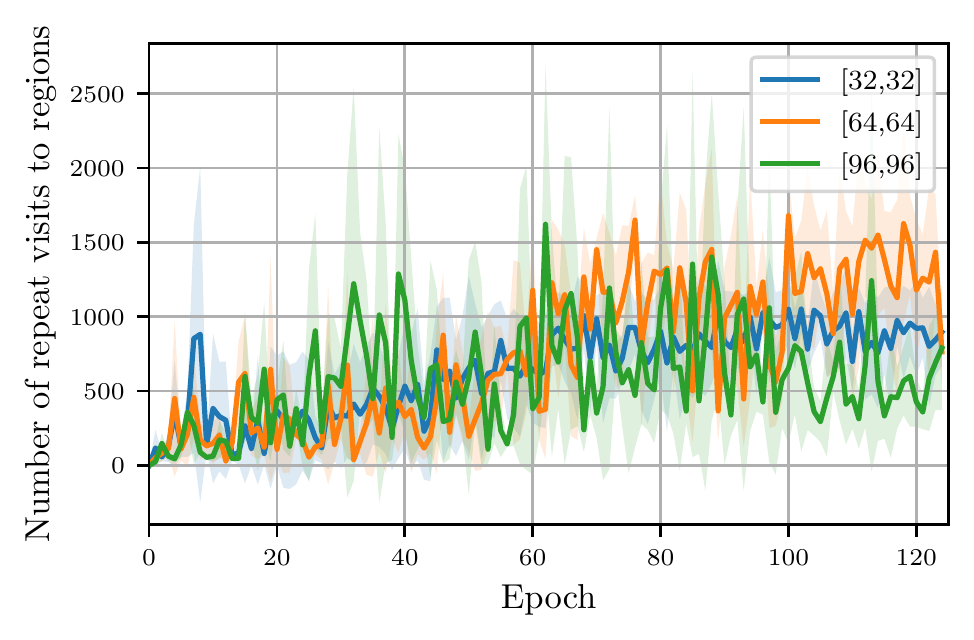}
            \caption{{\small Walker - SAC}}    
            \label{fig:sac-repeat-visits-current-w}
        \end{subfigure}
        \vskip\baselineskip
        \begin{subfigure}[b]{0.45\textwidth}
            \centering
            \includegraphics[width=\textwidth]{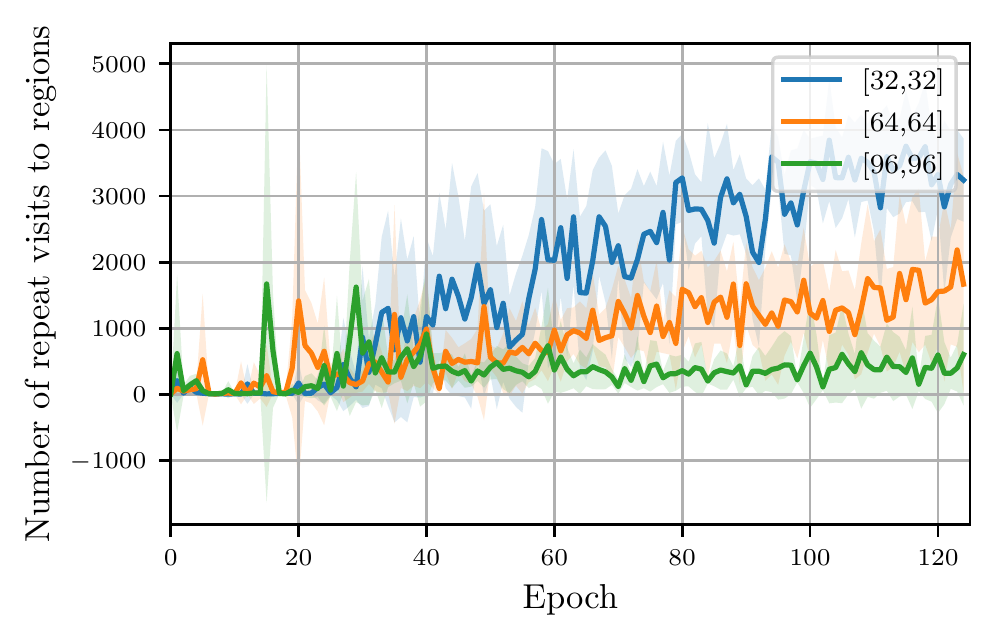}
            \caption{{\small Ant - SAC}}    
            \label{fig:sac-repeat-visits-current-a}
        \end{subfigure}
        \hfill        
        \begin{subfigure}[b]{0.45\textwidth}   
            \centering 
            \includegraphics[width=\textwidth]{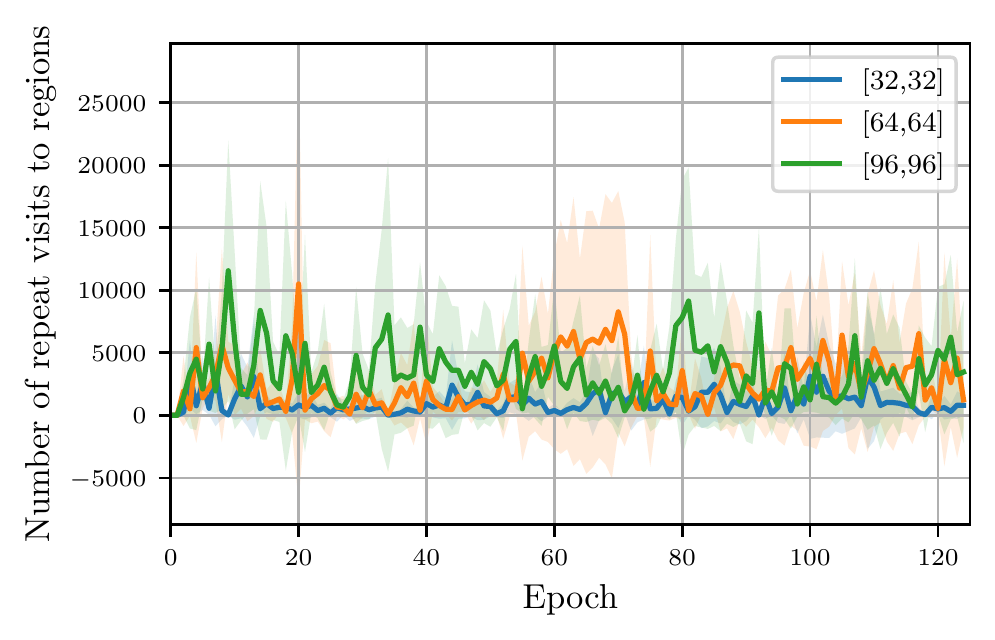}
            \caption{{\small Swimmer - SAC}}    
            \label{fig:sac-repeat-visits-current-s}
        \end{subfigure}
        \caption{Evolution of the number of repeat visits to regions over trajectories sampled from the current snapshot of the policy ($\tau$) during training different tasks with SAC. 
        }
        \label{fig:sac-repeat-visits-current}
    \end{figure}

\begin{figure}[H]
        \centering
        \begin{subfigure}[b]{0.45\textwidth}  
            \centering 
            \includegraphics[width=\textwidth]{figs/sac/density-random-lines-origin-cheetah.pdf}
            \caption{{\small HalfCheetah - SAC}}    
            \label{fig:sac-density-random-lines-origin-hc}
        \end{subfigure}
        \hfill
        \begin{subfigure}[b]{0.45\textwidth}   
            \centering 
            \includegraphics[width=\textwidth]{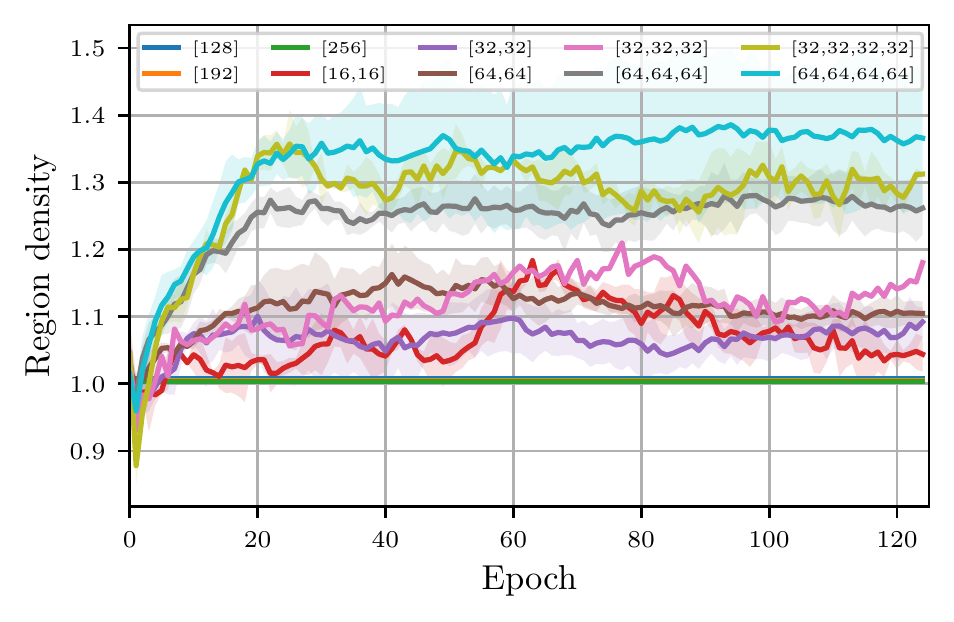}
            \caption{{\small Walker - SAC}}    
            \label{fig:sac-density-random-lines-origin-w}
        \end{subfigure}
        \vskip\baselineskip
        \begin{subfigure}[b]{0.45\textwidth}
            \centering
            \includegraphics[width=\textwidth]{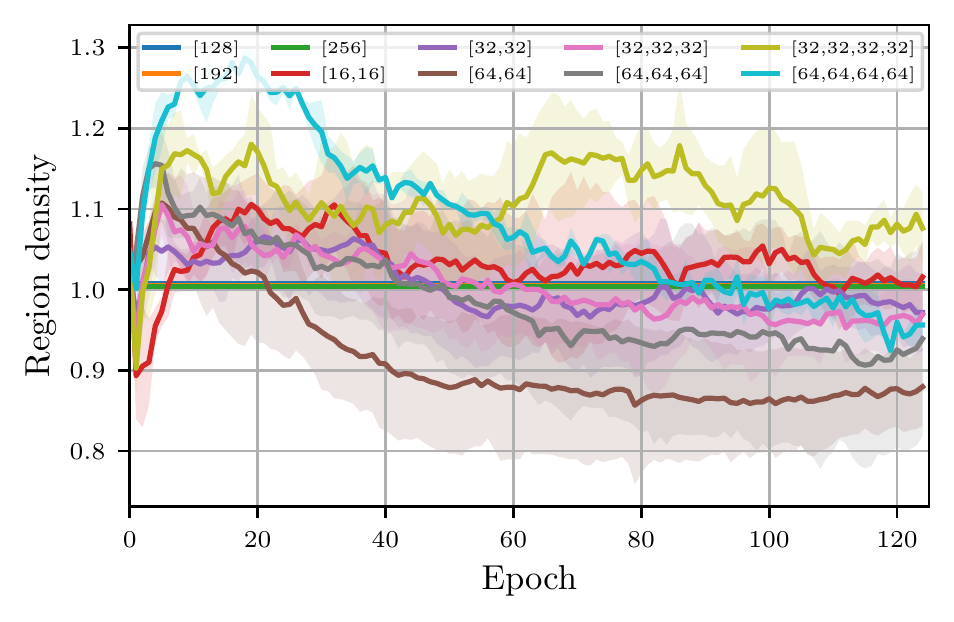}
            \caption{{\small Ant - SAC}}    
            \label{fig:sac-density-random-lines-origin-a}
        \end{subfigure}
        \hfill        
        \begin{subfigure}[b]{0.45\textwidth}   
            \centering 
            \includegraphics[width=\textwidth]{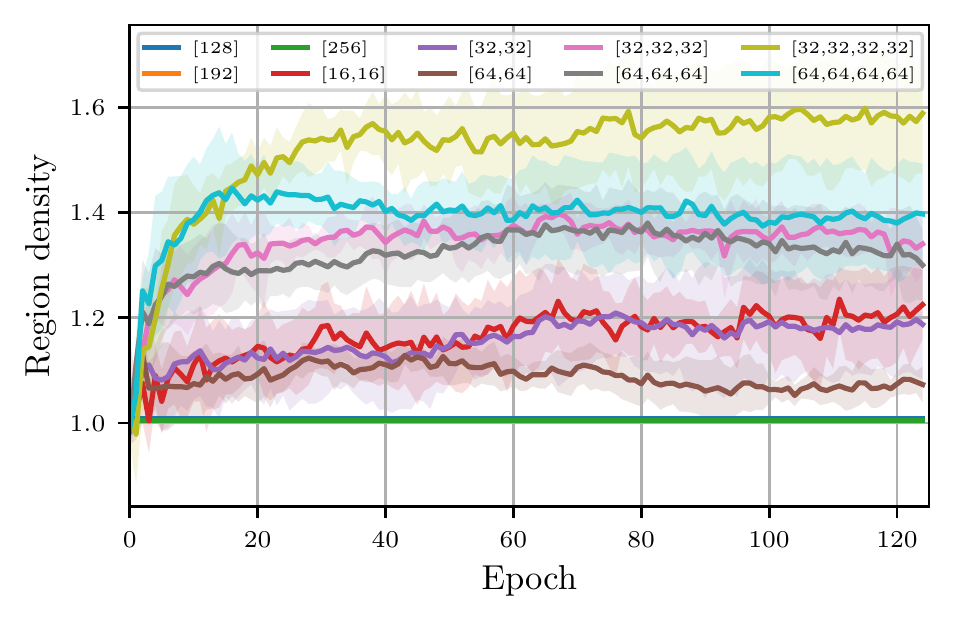}
            \caption{{\small Swimmer - SAC}}    
            \label{fig:sac-density-random-lines-origin-s}
        \end{subfigure}
        \caption{Evolution of the mean normalized region density over 100 random lines passing through the origin during training different tasks with SAC. For each policy network configuration, we sample 100 random lines and compute the density of transitions as we sweep along these lines. We then report the mean density observed over these 100 lines. 
        }
        \label{fig:sac-density-random-lines-origin}
    \end{figure}

\begin{figure}[H]
        \centering
        \begin{subfigure}[b]{0.45\textwidth}  
            \centering 
            \includegraphics[width=\textwidth]{figs/sac/density-random-traj-cheetah.pdf}
            \caption{{\small HalfCheetah - SAC}}    
            \label{fig:sac-density-random-traj-hc}
        \end{subfigure}
        \hfill
        \begin{subfigure}[b]{0.45\textwidth}   
            \centering 
            \includegraphics[width=\textwidth]{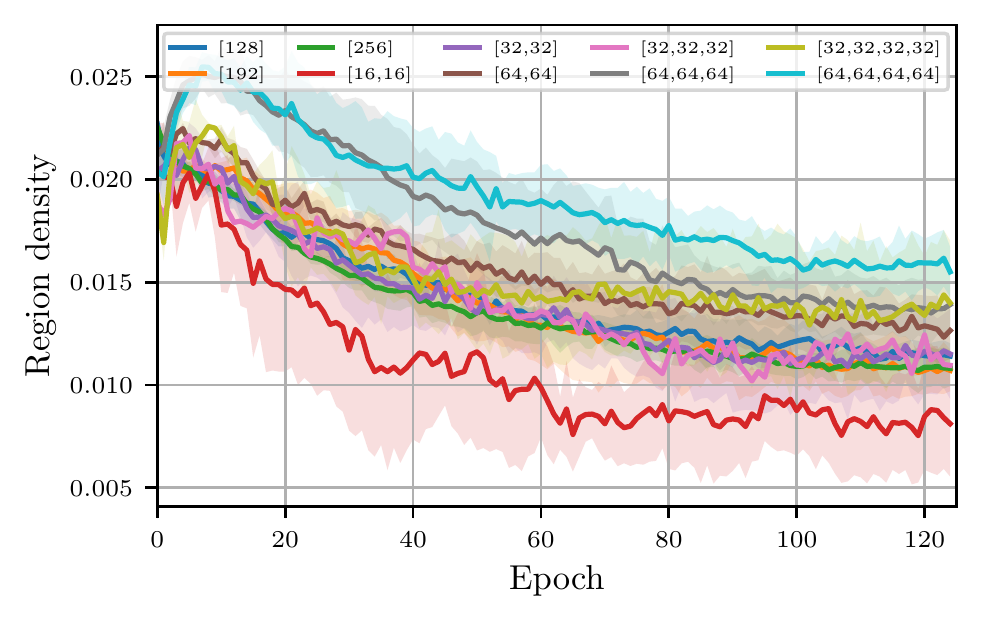}
            \caption{{\small Walker - SAC}}    
            \label{fig:sac-density-random-traj-w}
        \end{subfigure}
        \vskip\baselineskip
        \begin{subfigure}[b]{0.45\textwidth}
            \centering
            \includegraphics[width=\textwidth]{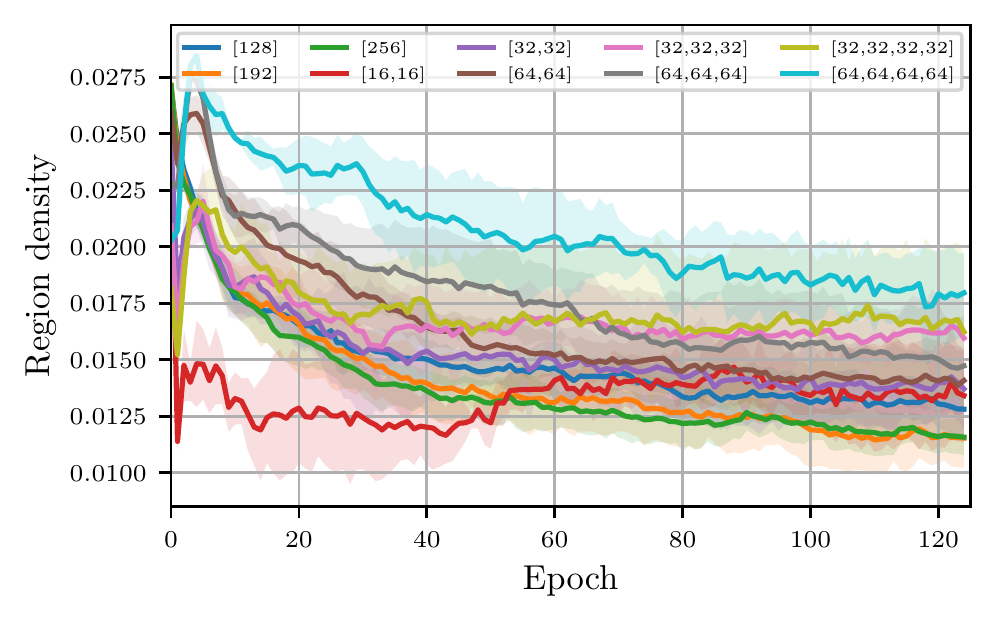}
            \caption{{\small Ant - SAC}}    
            \label{fig:sac-density-random-traj-a}
        \end{subfigure}
        \hfill        
        \begin{subfigure}[b]{0.45\textwidth}   
            \centering 
            \includegraphics[width=\textwidth]{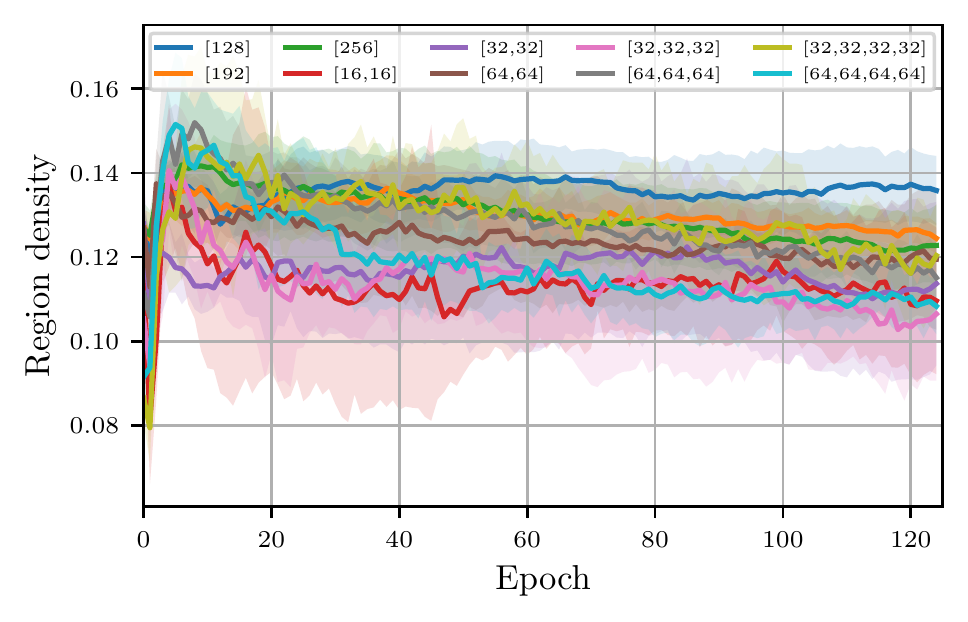}
            \caption{{\small Swimmer - SAC}}    
            \label{fig:sac-density-random-traj-s}
        \end{subfigure}
        \caption{Evolution of the mean normalized region density over 10 random-action trajectories ($\tau^R$) during training different tasks with SAC. For each policy network configuration, we sample 10 random-action trajectories and compute the density of transitions as we sweep along these trajectories, and report the mean value of these trajectories. 
        }
        \label{fig:sac-density-random-traj}
    \end{figure}

Figures~\ref{fig:sac-trans-fixed}-\ref{fig:sac-density-random-traj} of this section, show the same set of plots for all four environments, now grouped by metric type to enable easier comparison between environments. As can be observed the results for normalized region densities on the fixed final trajectory (Figure~\ref{fig:sac-trans-fixed}) are broadly consistent across environments (tasks), with Swimmer being an exception.

    \begin{table}[H]
        \caption{Policy network architectures in SAC experiments}
        \label{tbl:sac-policy-dimensions}
        \centering
            \begin{tabular}{l ccc}
            \toprule
                Neurons  & \multicolumn{3}{c}{Layer Widths} \\
                \midrule
                N=32 & - & 16,16 & - \\
                N=64 & - & 32,32 & - \\
                N=96 & - & - & 32,32,32 \\
                N=128 & 128 & 64,64  & 32,32,32,32 \\
                N=192 & 192 & 96,96  & 64,64,64 \\
                N=256 & 256 & 128,128  & 64,64,64 \\
            \bottomrule
            \end{tabular}
    \end{table}
    
    \begin{table}[H]
        \caption{Performance Metrics of SAC experiments}
        \label{tbl:sac-performance}
        \centering
            \begin{tabular}{lcccc}
            \toprule
            500K Step Return & Average & Max & Min & Std \\
            \midrule
            HalfCheetah & $5752.8$ & $8540.4$ & $1573.8$ & $1307.7$\\
            Walker & $3746.2$ & $4865.2$ & $1207.2$ & $812$\\
            Ant & $1998.6$ & $4982.4$ & $-55$ & $1110.9$ \\
            Swimmer & $48.2$ & $75.1$ & $-16.4$ & $10.1$  \\
            \midrule
            1M Step Return & Average & Max & Min & Std \\
            \midrule
            HalfCheetah & $6572.9$ & $9436.1$ & $1703.1$ & $1540.6$\\
            Walker & $4238.4$ & $5131.5$ & $1863$ & $568.8$\\
            Ant & $2501.6$ & $5248.1$ & $-115.5$ & $1116.7$\\
            Swimmer & $51.2$ & $97.4$ & $39.3$ & $8.4$  \\
            \bottomrule
            \end{tabular}
    \end{table}
    
    \begin{table}[H]
        \caption{Hyperparameters of SAC experiments}
        \label{tbl:sac-hyperparams}
        \centering
            \begin{tabular}{l ccc}
            \toprule
                Hyperparameter  & Value \\
                \midrule
                Training steps  & $1 \times 10^6$ \\
                Learning rate & $3 \times 10^{-4}$ \\
                Number of epochs  & $125$ \\
                Batch size  & $256$ \\
                Discount ($\gamma$) & $0.99$ \\
                Soft update coefficient ($\tau$) & $0.005$\\
                Entropy coeff. (initial)& $0.1$\\
                Hardware & CPU \\
            \bottomrule
            \end{tabular}
    \end{table}

\clearpage
\section{Decision Regions viewed via Embedding Planes}\label{appendix:Decision Regions viewed via Embedding Planes}
\begin{figure}[H]
\setlength\tabcolsep{1pt} 
\centering
\begin{tabular}{@{} r M{0.22\linewidth} M{0.22\linewidth} M{0.22\linewidth} M{0.22\linewidth} @{}}
& $\texttt{Epoch}=0$ & $\texttt{Epoch}=10$ & $\texttt{Epoch}=20$ & $\texttt{Epoch}=100$\\
\begin{subfigure}{0.07\linewidth} \caption*{$\tau^*$} \end{subfigure} 
  & \includegraphics[width=\hsize, trim={0 0 0 0.7cm },clip] {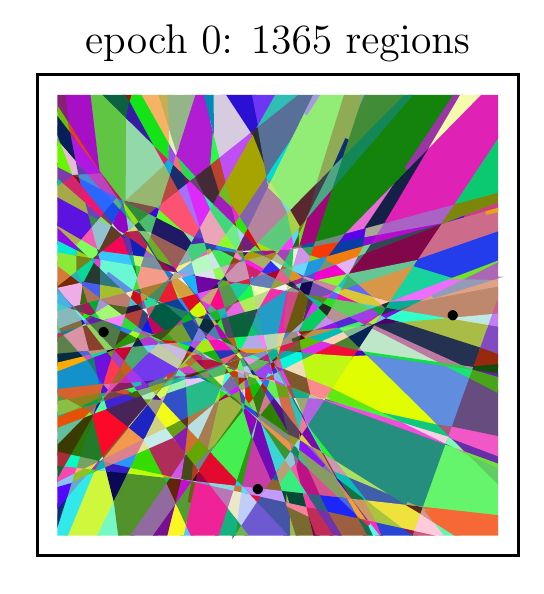} 
  & \includegraphics[width=\hsize, trim={0 0 0 0.7cm },clip] {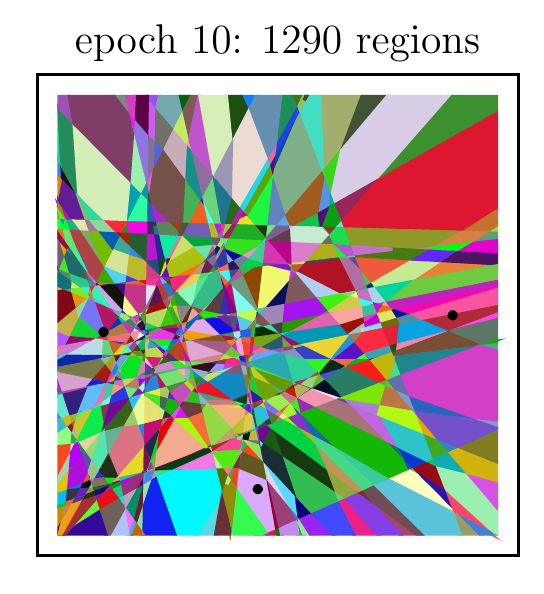} 
  & \includegraphics[width=\hsize, trim={0 0 0 0.7cm },clip] {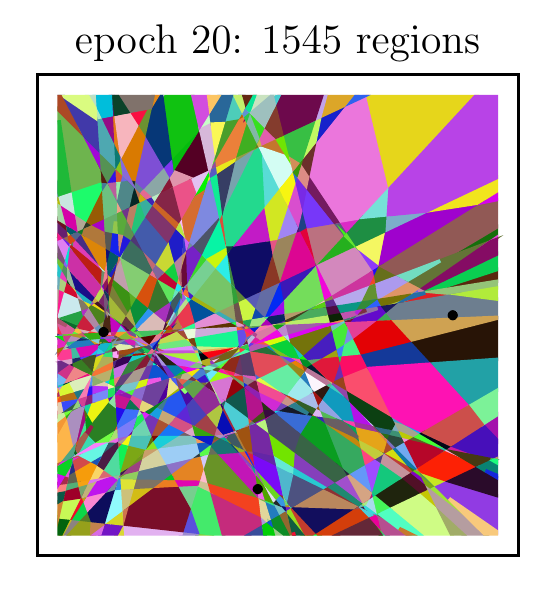} 
  & \includegraphics[width=\hsize, trim={0 0 0 0.7cm },clip] {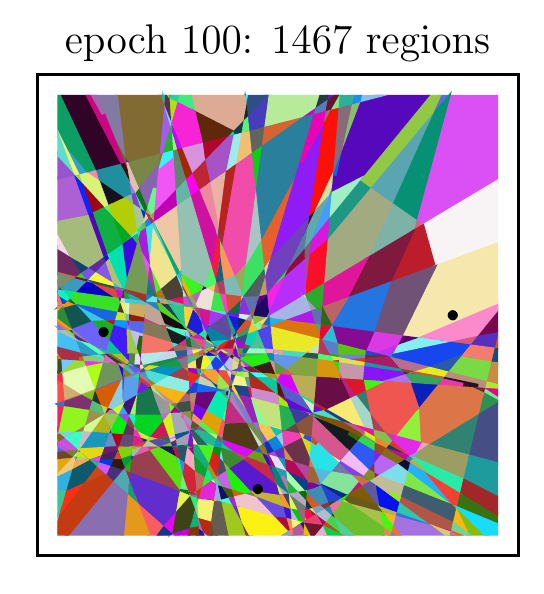}\\ \addlinespace
\begin{subfigure}{0.07\linewidth} \caption*{$\tau^R$} \end{subfigure} 
  & \includegraphics[width=\hsize, trim={0 0 0 0.7cm },clip] {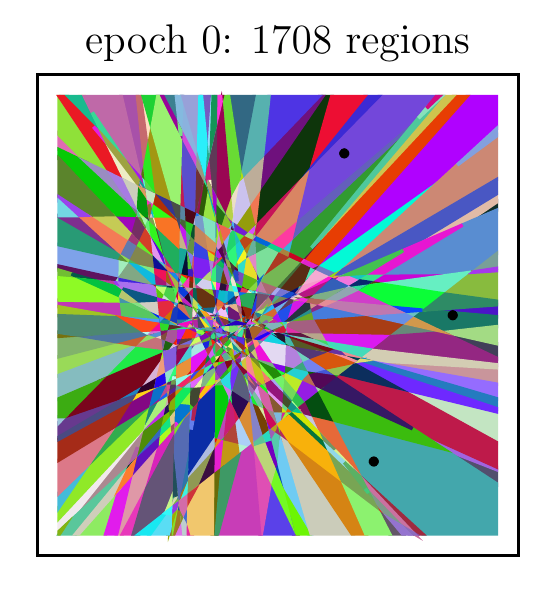}    
  & \includegraphics[width=\hsize, trim={0 0 0 0.7cm },clip] {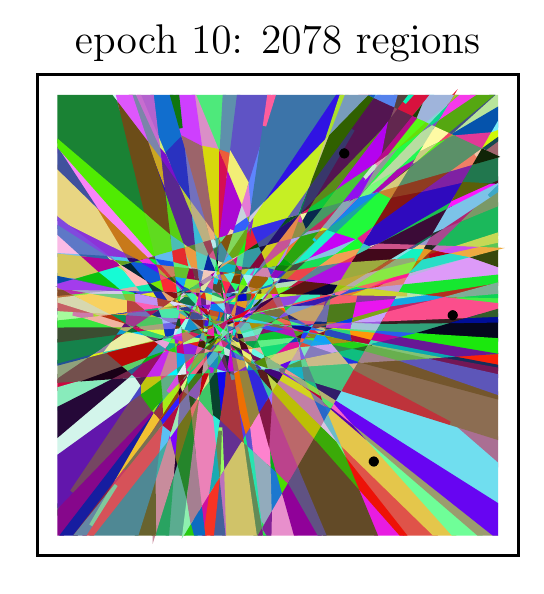} 
  & \includegraphics[width=\hsize, trim={0 0 0 0.7cm },clip] {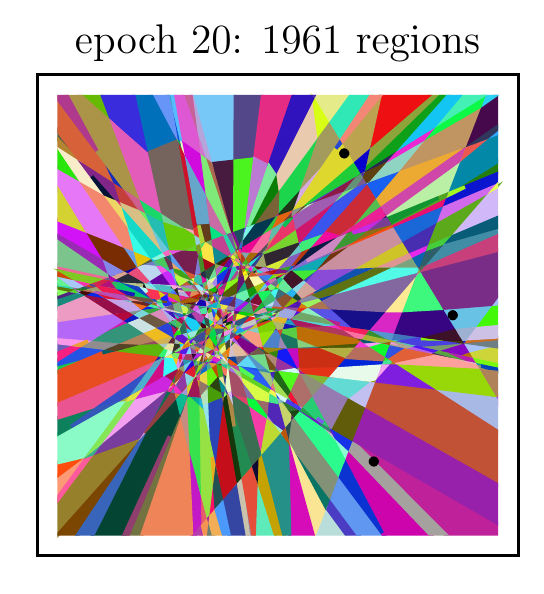} 
  & \includegraphics[width=\hsize, trim={0 0 0 0.7cm },clip] {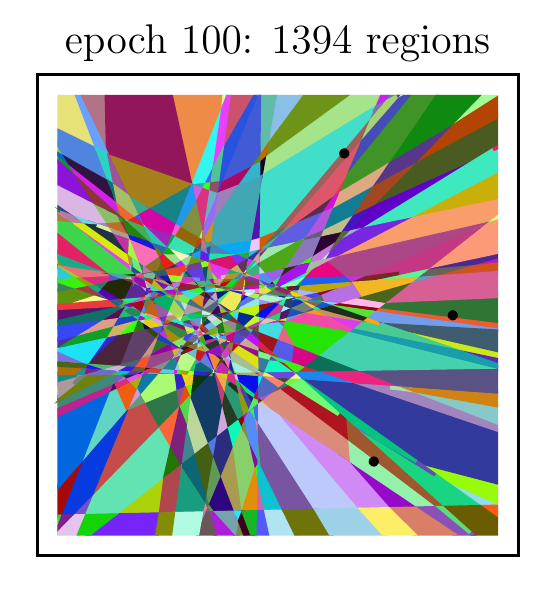}\\
\end{tabular}
\caption{Linear regions that intersect a 2D plane through the input space for a network of depth 2 and width 32 trained on HalfCheetah-v2. Black dots indicate three points from the input space on which the plane is defined. In the first row, the three points are randomly sampled from the final trajectory trajectory $\tau^*$ whereas in the second row, the three points are randomly sampled from a random-action trajectory $\tau^R$.}
\label{fig:hd_visualization}
\end{figure}

To gain insights on what the linear regions may look like in our high dimensional setting, similar to \citet{novak2018sensitivity} and \citet{hanin2019complexity}, we visualize the linear regions over a 2-dimensional slice through the input space defined by three points sampled from a trajectory emerged from the policy. Figure~\ref{fig:hd_visualization} shows the linear regions of a $(32,32)$ policy network trained on HafCheetah in the 17-dimensional input space, over a 2-dimensional plane crossing three projection points sampled from the final trajectory ($\tau^*$) in the first row, and sampled from the random-action trajectory ($\tau^R$) in the second row. Comparing the visualizations of projected on points from $\tau^*$ and $\tau^R$ (first and second row), we do not observe a significant difference between the granularity of regions over points of these two trajectories. Moreover, surprisingly, our visualizations are more consistent with the findings of \citet{novak2018sensitivity} with the projection points lying in regions of lower density.

\section{Decision Regions for a Non-Cyclic Task: LunarLander}\label{appendix:Decision Regions for a Non-Cyclic Task: LunarLander}
\begin{figure}[H]
        \centering
        \begin{subfigure}[b]{0.45\textwidth}  
            \centering 
            \includegraphics[width=\textwidth]{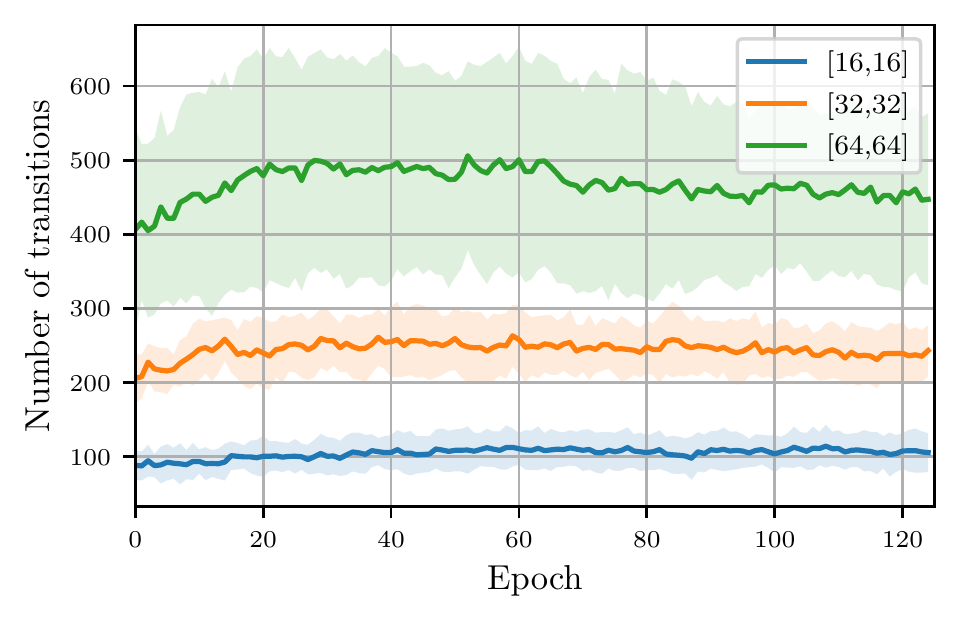}
            \caption{{\small Transition count, fixed trajectory $\tau^*$}}    
            \label{fig:llc-trans-fixed}
        \end{subfigure}
        \hfill
        \begin{subfigure}[b]{0.45\textwidth}   
            \centering 
            \includegraphics[width=\textwidth]{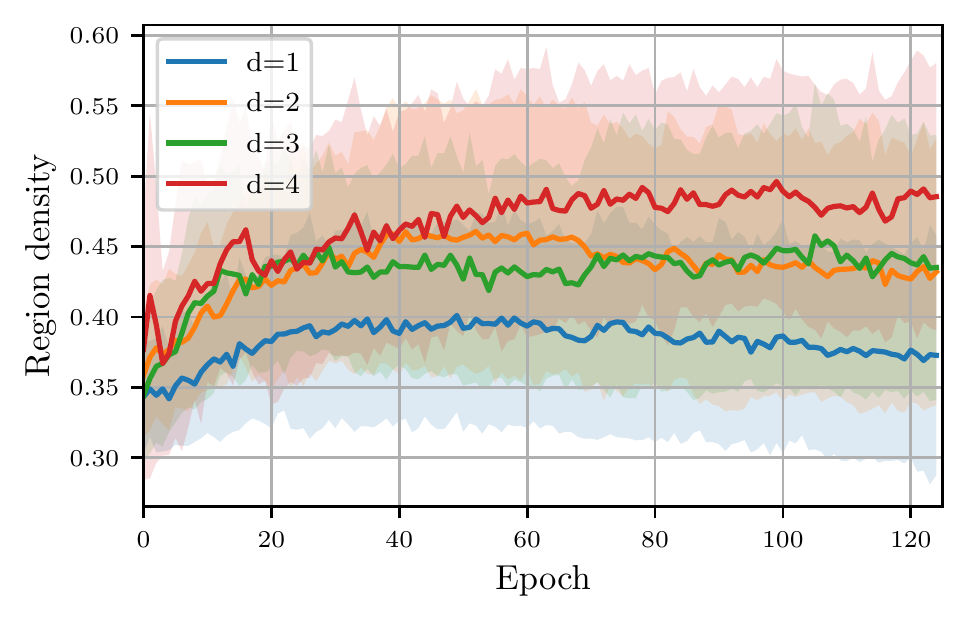}
            \caption{{\small Region density, fixed trajectory $\tau^*$}}    
            \label{fig:llc-density-fixed}
        \end{subfigure}
        \vskip\baselineskip
        \begin{subfigure}[b]{0.45\textwidth}
            \centering
            \includegraphics[width=\textwidth]{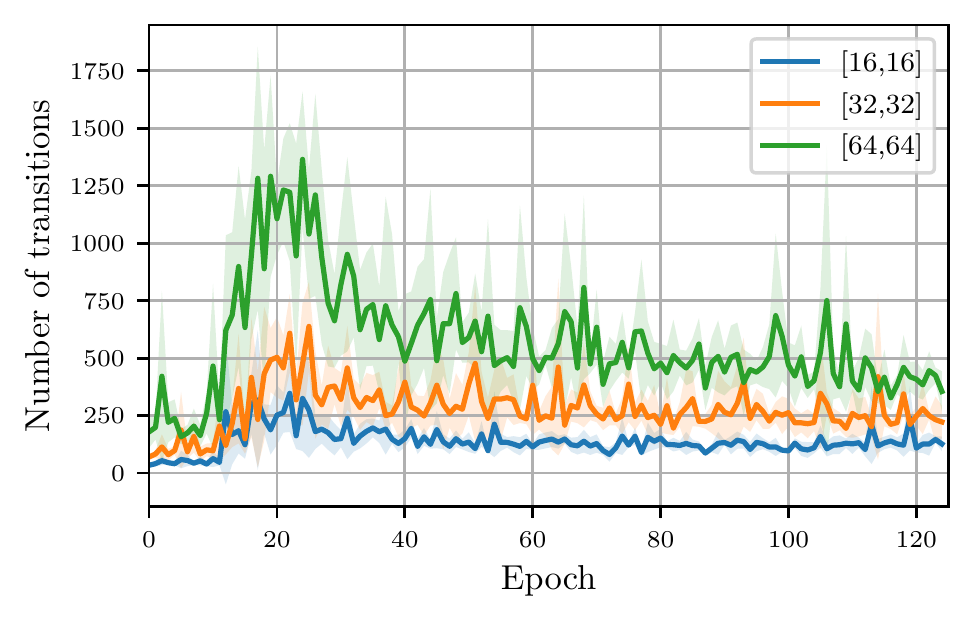}
            \caption{{\small Transition count, current trajectory $\tau$}}    
            \label{fig:llc-trans-current}
        \end{subfigure}
        \hfill        
        \begin{subfigure}[b]{0.45\textwidth}   
            \centering 
            \includegraphics[width=\textwidth]{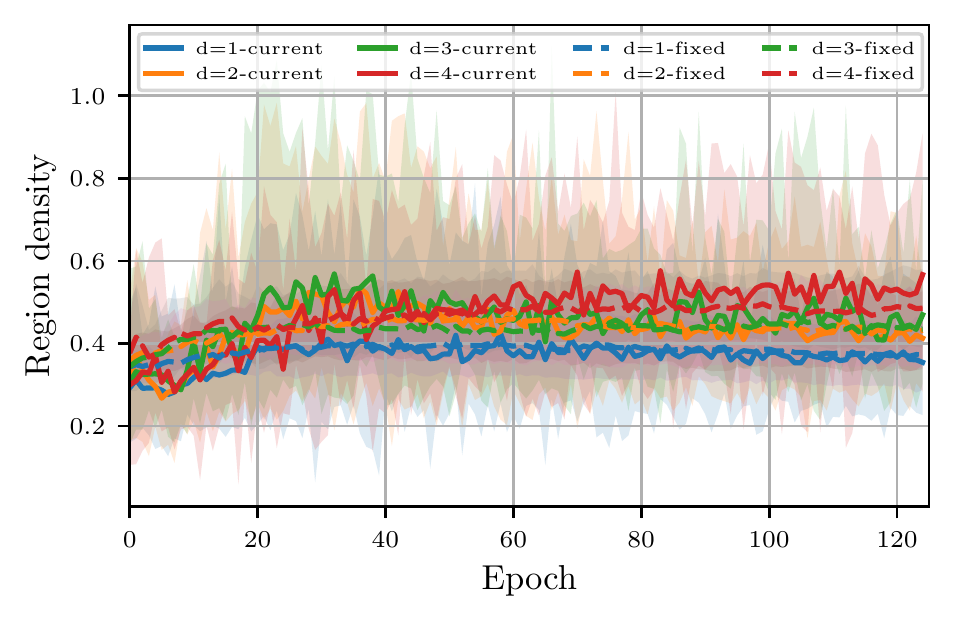}
            \caption{{\small Region density, $\tau^*$ and $\tau$}}    
            \label{fig:llc-density-both}
        \end{subfigure}
        \vskip\baselineskip
        \begin{subfigure}[b]{0.45\textwidth}   
            \centering 
            \includegraphics[width=\textwidth]{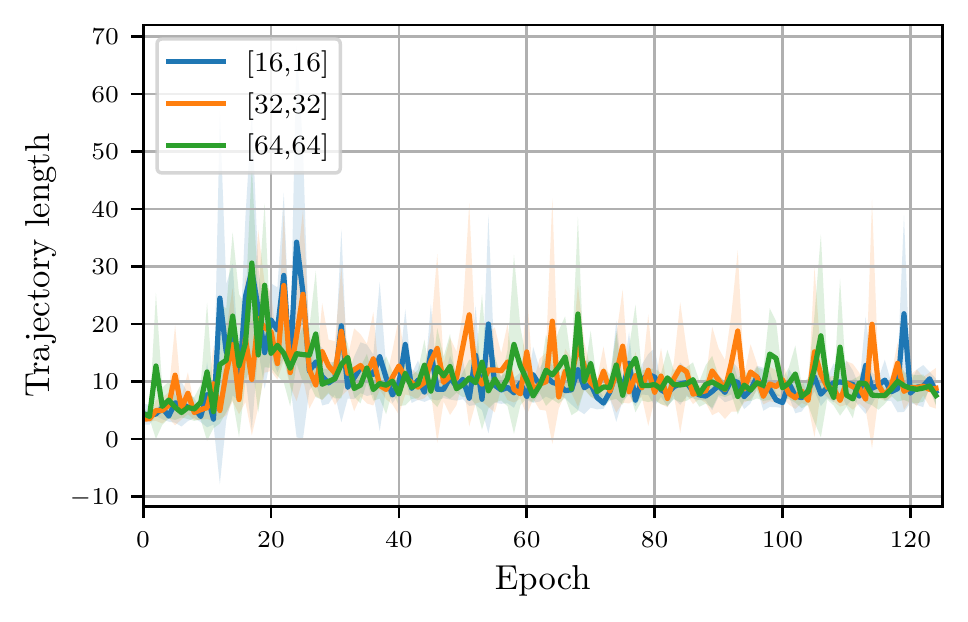}
            \caption{{\small Trajectory length, current trajectory $\tau$}}    
            \label{fig:llc-length-curr}
        \end{subfigure}
        \hfill        
        \begin{subfigure}[b]{0.45\textwidth}   
            \centering 
            \includegraphics[width=\textwidth]{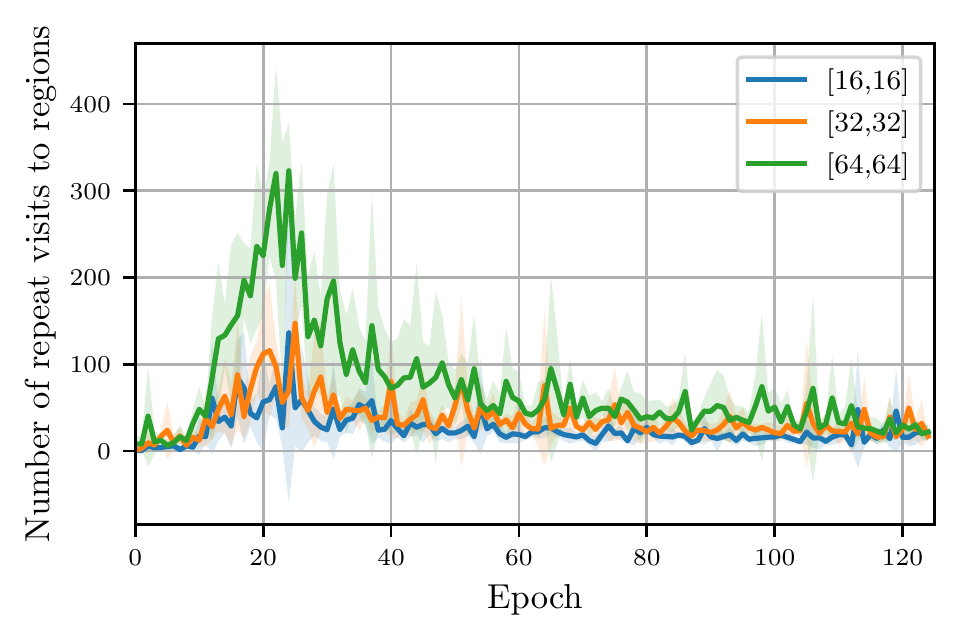}
            \caption{{\small Repeat visits to regions, current trajectory $\tau$}}    
            \label{fig:llc-repeat-visits-current}
        \end{subfigure}
        \vskip\baselineskip
        \begin{subfigure}[b]{0.45\textwidth}  
            \centering 
            \includegraphics[width=\textwidth]{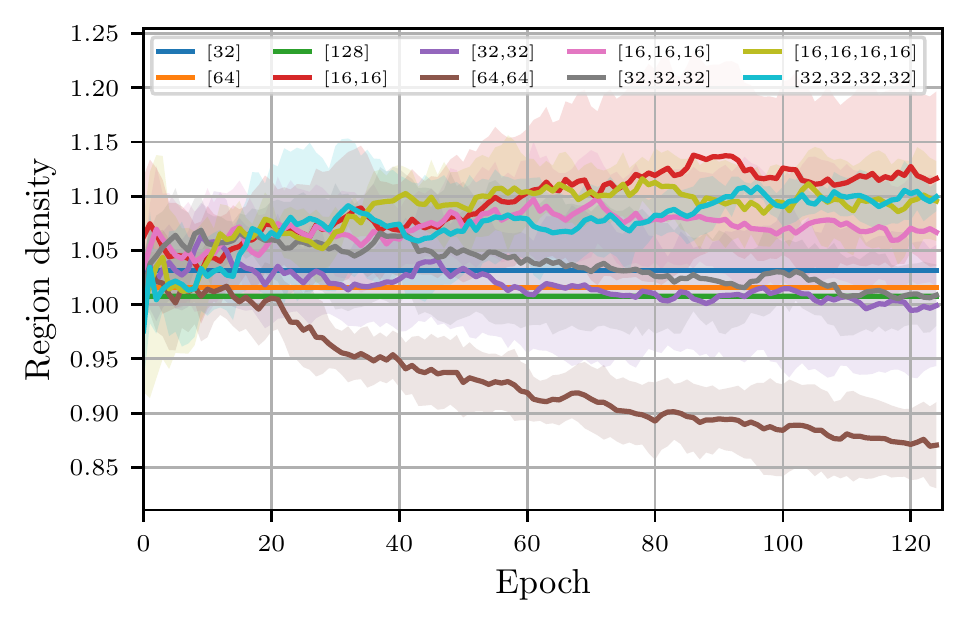}
            \caption{{\small Density for random lines through the origin}}    
            \label{fig:llc-density-random-line-origin}
        \end{subfigure}
        \hfill
        \begin{subfigure}[b]{0.45\textwidth}   
            \centering 
            \includegraphics[width=\textwidth]{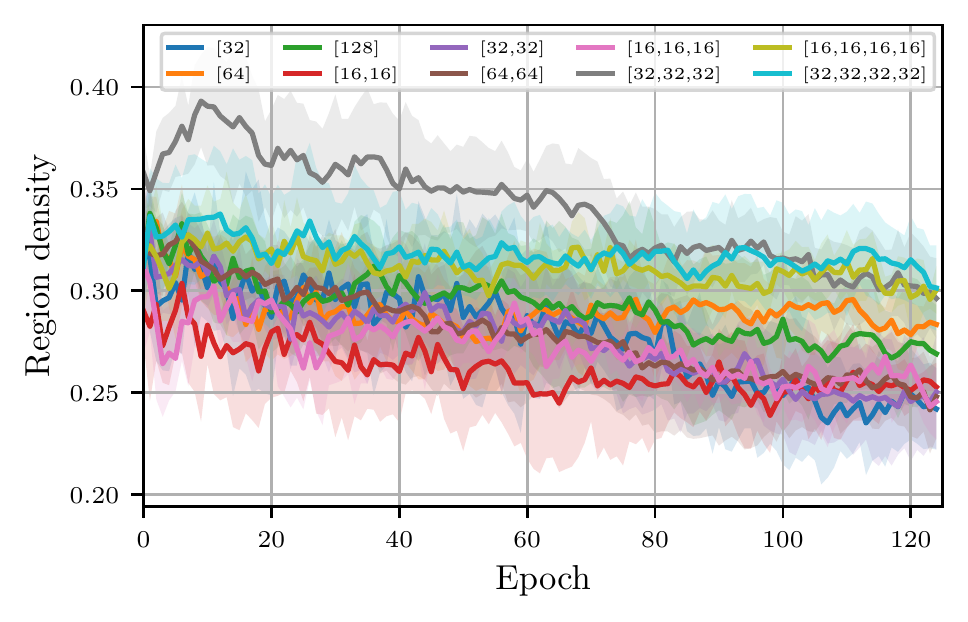}
            \caption{{\small Density for random-action trajectories}}    
            \label{fig:llc-density-random-traj}
        \end{subfigure}
        \caption{{\small LunarLander Results. Training is with PPO and the ranges indicate the standard error across 5 random seeds. In the legend $[n_1, ..., n_d]$ describes a network architecture of depth $d$ and $n_i$ neurons in each layer. Summary results are grouped by network depth, $d$.}}
        \label{fig:lunarlander}
    \end{figure}

In this section, we aim to shed light on the possible effects of cyclic nature of locomotion tasks on our previous PPO results. Therefore, we repeat our experiments by training PPO on LunarLanderContinuous-v2 from OpenAI gym. Figure~\ref{fig:lunarlander} shows the plots from this experiment. The linear-region evolution behavior is in fact still similar with the only discrepancy being the evolution pattern of observed density over current trajectories. For LunarLander, observed density over current trajectories does not decrease during training while having the same range of values as the density over fixed and random-action trajectories. This is due to the difference between the nature of non-cyclic and cyclic tasks. For locomotion tasks, the length of the cyclic trajectories increased during training, while for LunarLander, the trajectory length initially increases, then plateaus once the the agent converges.

\section{Value Network Analysis}\label{appendix:Value Network Analysis}
\begin{figure}[H]
        \centering
        \begin{subfigure}[b]{0.45\textwidth}  
            \centering 
            \includegraphics[width=\textwidth]{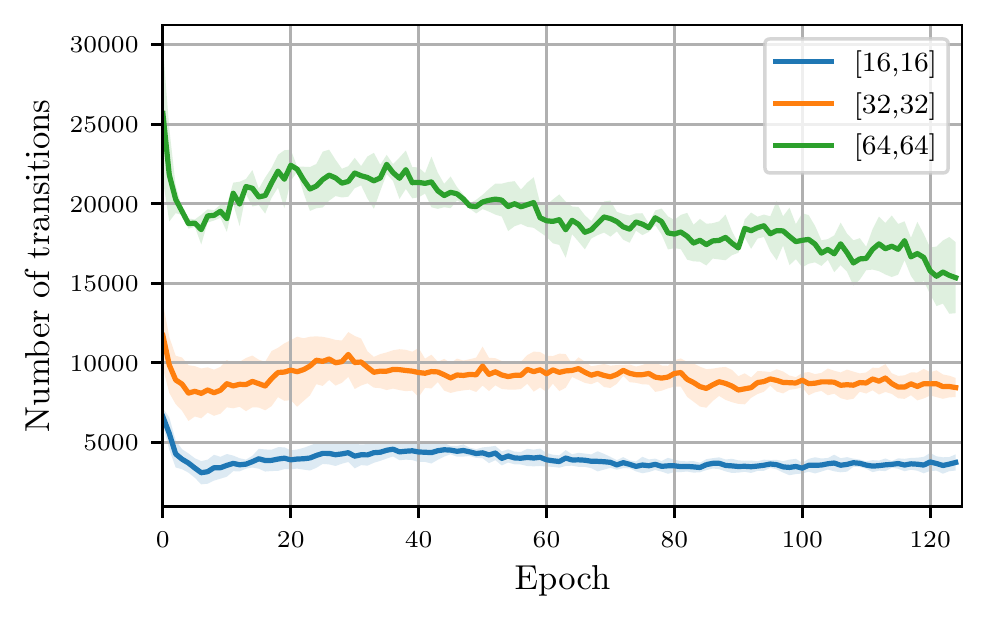}
            \caption{{\small Transition count, fixed trajectory $\tau^*$}}    
            \label{fig:vf-trans-fixed}
        \end{subfigure}
        \hfill
        \begin{subfigure}[b]{0.45\textwidth}   
            \centering 
            \includegraphics[width=\textwidth]{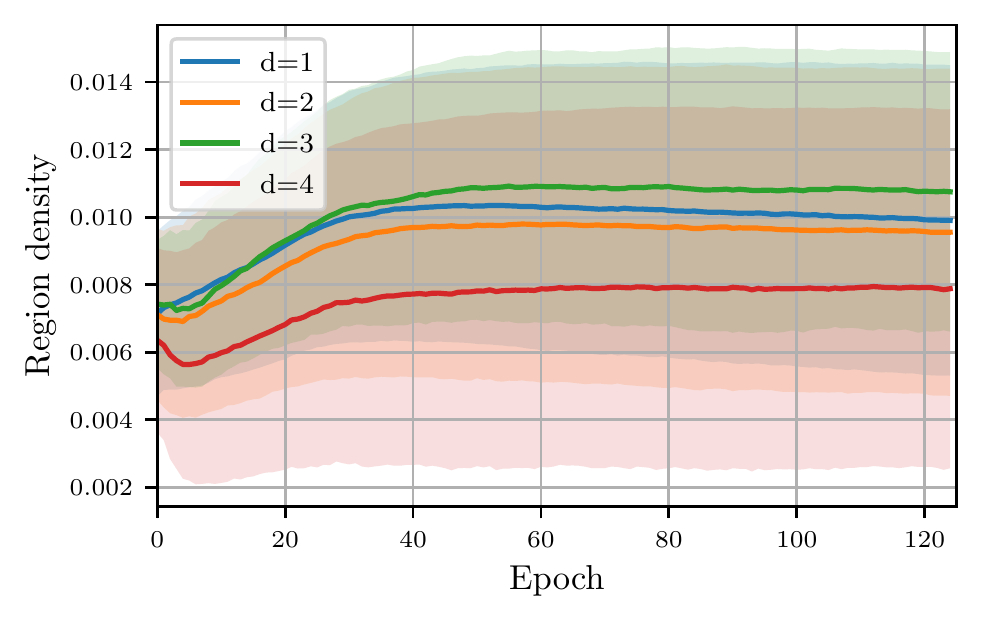}
            \caption{{\small Region density, fixed trajectory $\tau^*$}}    
            \label{fig:vf-density-fixed}
        \end{subfigure}
        \vskip\baselineskip
        \begin{subfigure}[b]{0.45\textwidth}
            \centering
            \includegraphics[width=\textwidth]{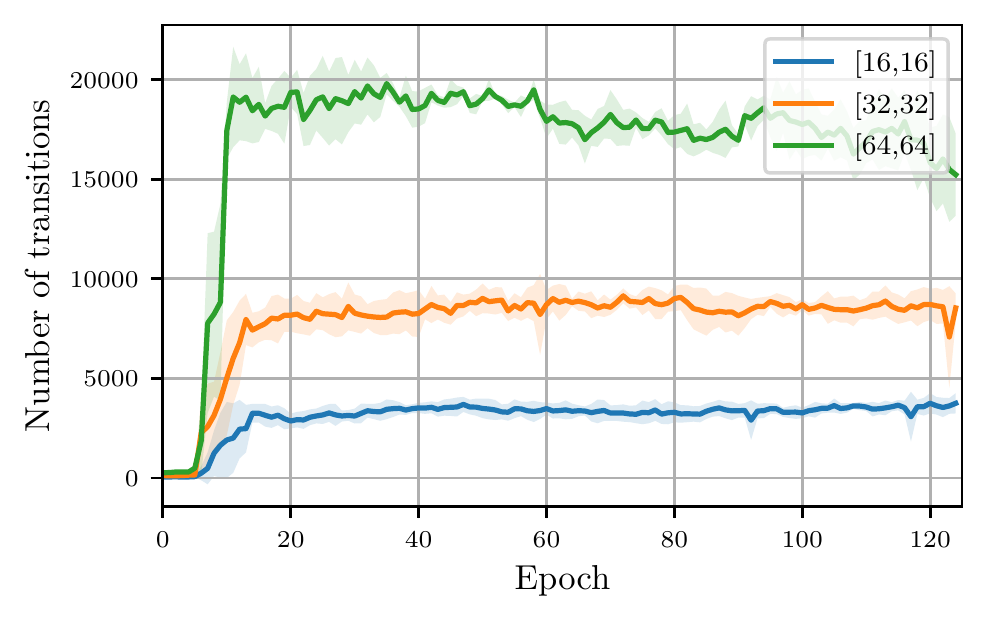}
            \caption{{\small Transition count, current trajectory $\tau$}}    
            \label{fig:vf-trans-current}
        \end{subfigure}
        \hfill        
        \begin{subfigure}[b]{0.45\textwidth}   
            \centering 
            \includegraphics[width=\textwidth]{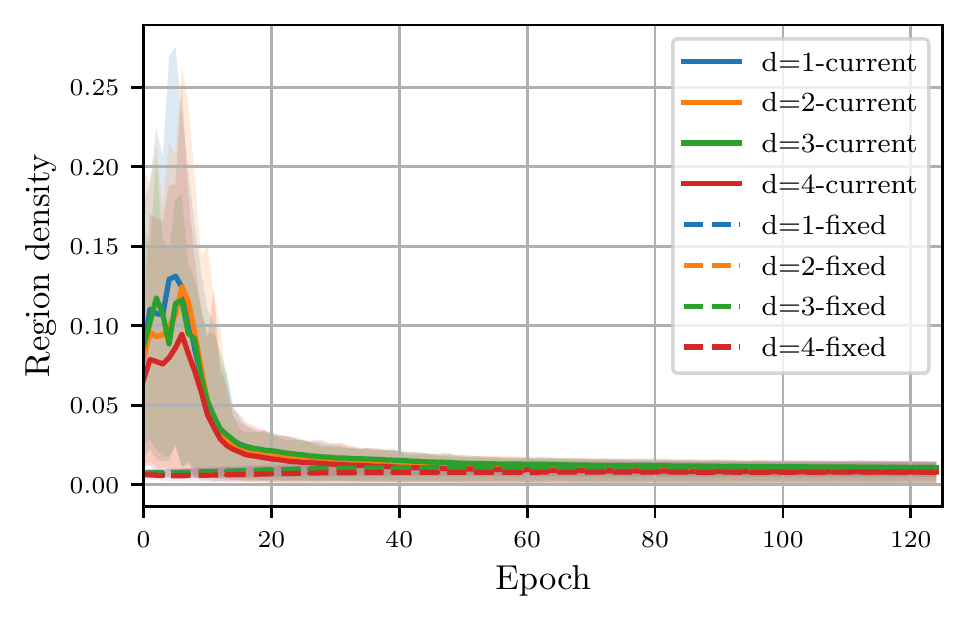}
            \caption{{\small Region density, $\tau^*$ and $\tau$}}    
            \label{fig:vf-density-both}
        \end{subfigure}
        \vskip\baselineskip
        \begin{subfigure}[b]{0.45\textwidth}   
            \centering 
            \includegraphics[width=\textwidth]{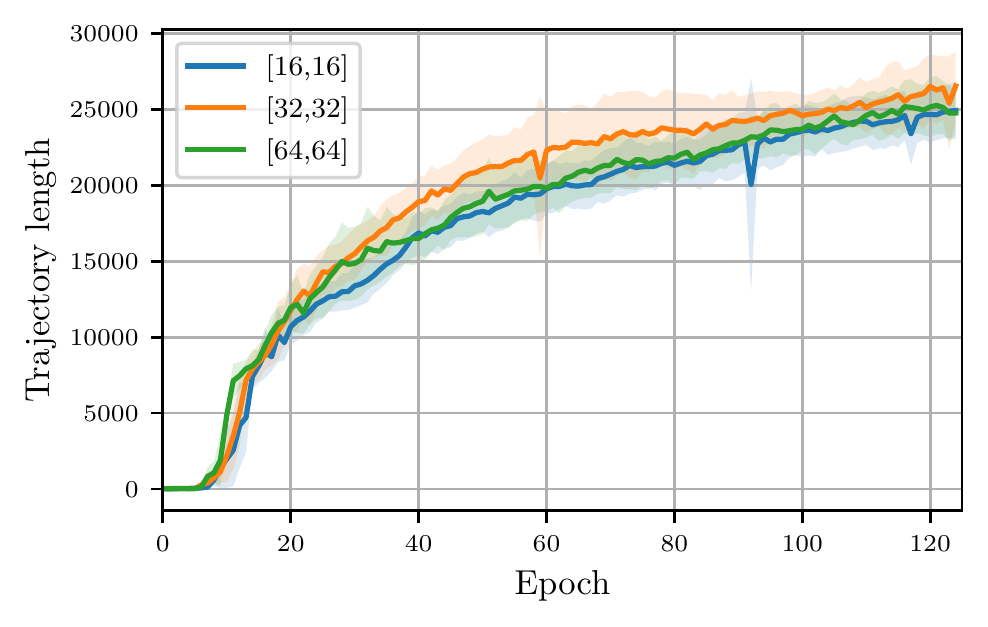}
            \caption{{\small Trajectory length, current trajectory $\tau$}}    
            \label{fig:vf-length-curr}
        \end{subfigure}
        \hfill        
        \begin{subfigure}[b]{0.45\textwidth}   
            \centering 
            \includegraphics[width=\textwidth]{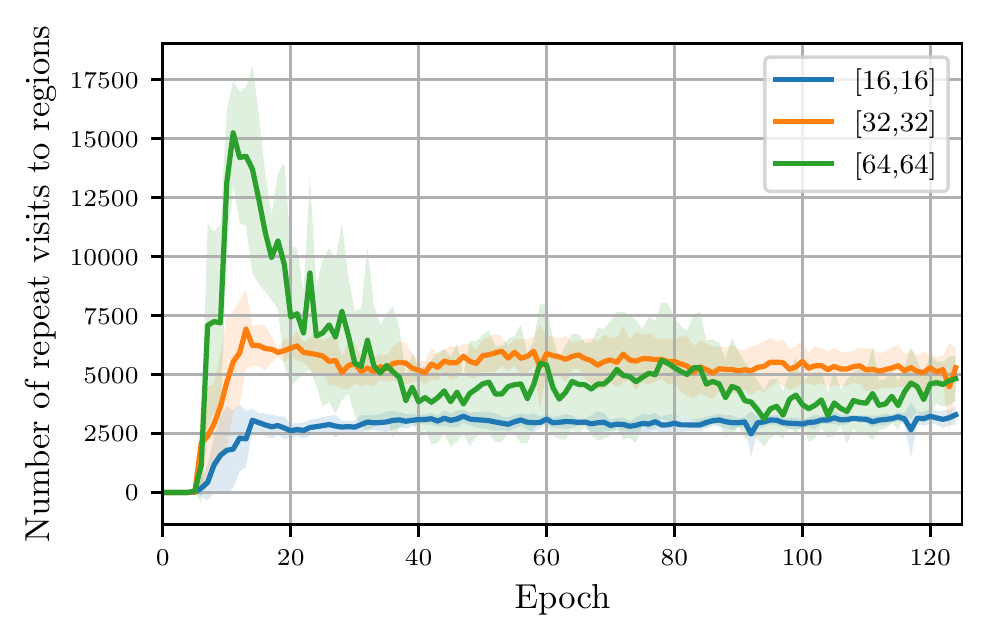}
            \caption{{\small Repeat visits to regions, current trajectory $\tau$}}    
            \label{fig:vf-repeat-visits-current}
        \end{subfigure}
        \vskip\baselineskip
        \begin{subfigure}[b]{0.45\textwidth}  
            \centering 
            \includegraphics[width=\textwidth]{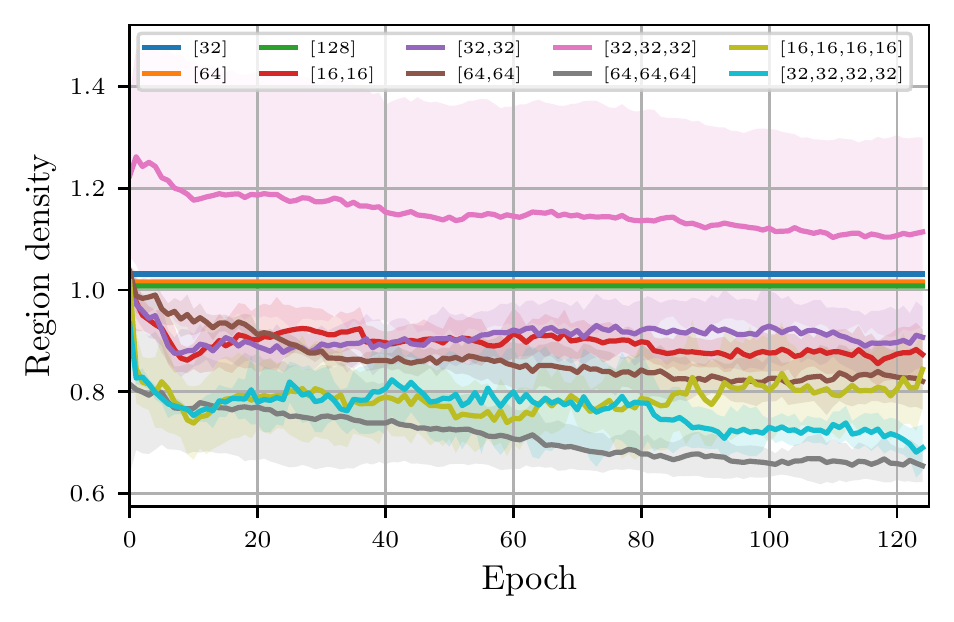}
            \caption{{\small Density for random lines through the origin}}    
            \label{fig:vf-density-random-line-origin}
        \end{subfigure}
        \hfill
        \begin{subfigure}[b]{0.45\textwidth}   
            \centering 
            \includegraphics[width=\textwidth]{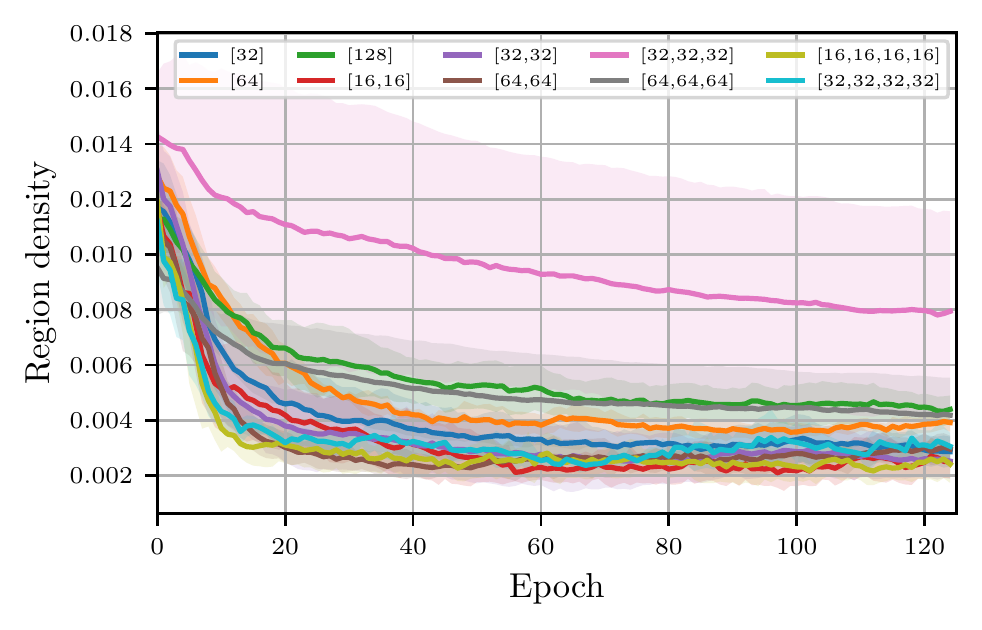}
            \caption{{\small Density for random-action trajectories}}    
            \label{fig:vf-density-random-traj}
        \end{subfigure}
        \caption{{\small Evolution of the number of transitions, linear region densities, and length of trajectories for the value function, during training HalfCheetah with PPO. The ranges indicate the standard error across 5 random seeds. $[n_1, ..., n_d]$ in the legend corresponds to a network architecture with depth $d$ and $n_i$ neurons in each layer. Summary results are grouped by network depth, $d$.}}
        \label{fig:valuefunction}
    \end{figure}

We also explore decision regions defined by the value-function network. We repeat our main experiments on the value network of PPO agents. More particularly, we train a policy network with a fixed network structure of $(64,64)$ with 18 different value network configurations on HalfCheetah using the PPO algorithm. The value network configurations used here are the same as the ones used for the policy network and are listed in Table~\ref{tbl:dimensions}, having $N \in \{32,48,64,96,128,192\}$ neurons, widths $w \in \{8,16,32,64\}$, and depths $d \in \{1,2,3,4\}$. These network configurations are chosen such that the network is fully-capable of learning the task and achieves near state-of-the-art cumulative reward on the particular task it is trained on. To compute the metrics, we now use the weights and biases of the value-function network, instead of the policy network. Figure~\ref{fig:valuefunction} shows the resulting plots. In comparing Figure~\ref{fig:valuefunction} and Figure~\ref{fig:halfCheetah}, we see that the evolution of the region densities are remarkably similar in structure, for both fixed and current trajectories.
\clearpage

\section{Behavior Cloning}\label{appendix:Behavior Cloning}
We perform an additional experiment to study whether linear regions emerging from policies trained with deep RL on non-IID data are different from those emerging from policies trained with BC on IID data. For a direct comparison, we repeat our experiments by first collecting expert data from each of the 18 network architectures previously trained with PPO on HalfCheetah. For each expert PPO policy, we initialize a neural network policy with the same architecture and train it using BC on the collected expert data. We then evaluate this new set of trained policies using the same evaluation method.

For policy architecture $\mathcal{A}$ previously trained with PPO on HalfCheetah, we randomly pick one of the 5 experts (from the random seeds), and collect 500 episodes of expert data by letting the policy interact with the environment. Actions are sampled from the stochastic policy. We then train a new policy network with the same architect $\mathcal{A}$ using BC on the collected dataset. We repeat each experiment with 5 random seeds. To ensure BC-trained policies are fully trained, we choose hyperparameters such that the policy achieves similar average episodic return to its expert PPO-trained policy. We present the details of the choices of hyperparameters in Table~\ref{tbl:bc-hyperparams}. We use the default initialization of linear layers used by PyTorch where weights are He normal with gain=$\sqrt{5}$, and biases are IID normal with variance $1/\texttt{fan-in}$.

    \begin{table}[H]
        \caption{Hyperparameters of BC experiments}
        \label{tbl:bc-hyperparams}
        \centering
            \begin{tabular}{l ccc}
            \toprule
                Hyperparameter  & Value \\
                \midrule
                Training epochs  & $125$ \\
                Learning rate & $10^{-3}$ \\
                Batch size  & $64$ \\
                Optimizer & SGD \\
                Hardware & CPU \\
            \bottomrule
            \end{tabular}
    \end{table}

\begin{figure}[H]
        \centering
        \begin{subfigure}[b]{0.45\textwidth}  
            \centering 
            \includegraphics[width=\textwidth]{figs/ppo/transitions-fixed-cheetah.pdf}
            \caption{{\small HalfCheetah - PPO}}    
            \label{fig:bc-compare-trans-fixed-ppo}
        \end{subfigure}
        \hfill
        \begin{subfigure}[b]{0.45\textwidth}   
            \centering 
            \includegraphics[width=\textwidth]{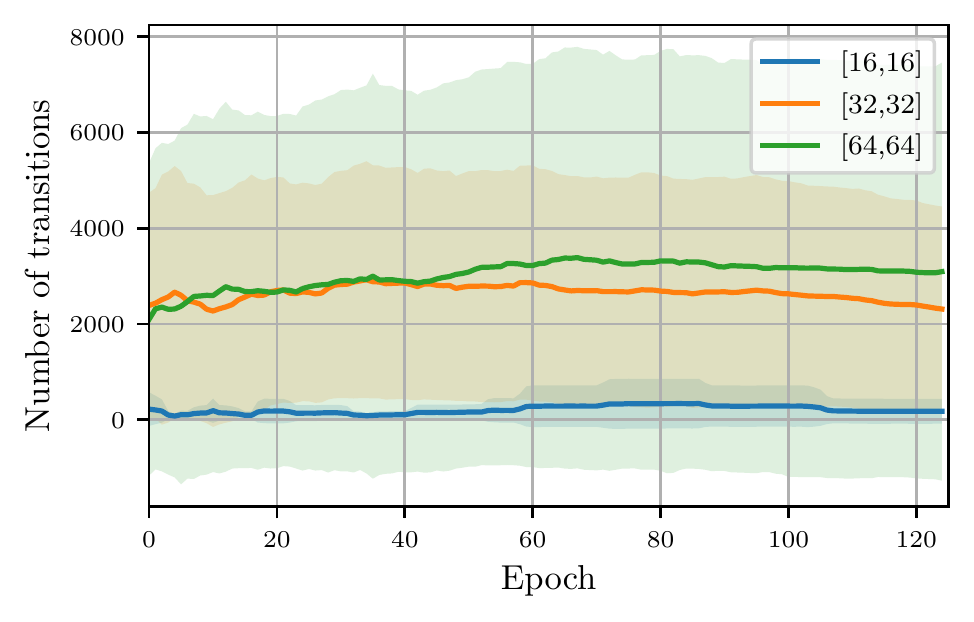}
            \caption{{\small HalfCheetah - BC}}    
            \label{fig:bc-compare-trans-fixed-bc}
        \end{subfigure}
        \caption{Evolution of the number of transitions over a fixed trajectory sampled from the final fully trained policy ($\tau^*$) during training HalfCheetah with BC and PPO algorithms. Plots show the mean and standard error across 5 random seeds. In the legend, $[n_1, ..., n_d]$ corresponds to a network architecture with depth $d$ and $n_i$ neurons in each layer. }
        \label{fig:bc-compare-trans-fixed}
    \end{figure}

\begin{figure}[H]
        \centering
        \begin{subfigure}[b]{0.45\textwidth}  
            \centering 
            \includegraphics[width=\textwidth]{figs/ppo/density-fixed-depth-cheetah.pdf}
            \caption{{\small HalfCheetah - PPO}}    
            \label{fig:bc-compare-density-fixed-ppo}
        \end{subfigure}
        \hfill
        \begin{subfigure}[b]{0.45\textwidth}   
            \centering 
            \includegraphics[width=\textwidth]{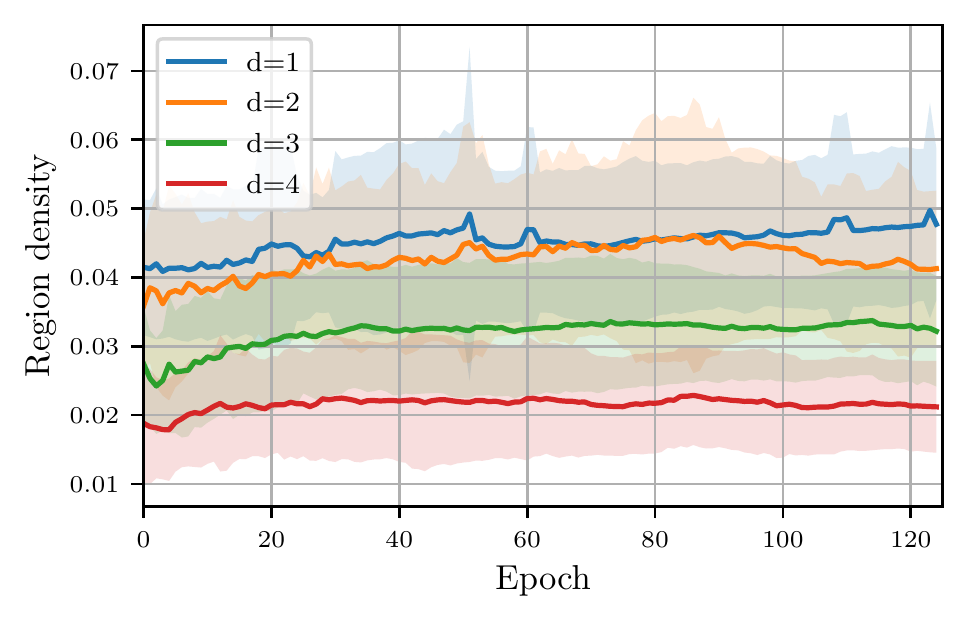}
            \caption{{\small HalfCheetah - BC}}    
            \label{fig:bc-compare-density-fixed-bc}
        \end{subfigure}
    \caption{Evolution of the normalized region density over a fixed trajectory sampled from the final fully trained policy ($\tau^*$) during training HalfCheetah with BC and PPO algorithms.}
        \label{fig:bc-compare-density-fixed}
    \end{figure}

\begin{figure}[H]
        \centering
        \begin{subfigure}[b]{0.45\textwidth}  
            \centering 
            \includegraphics[width=\textwidth]{figs/ppo/transitions-current-cheetah.pdf}
            \caption{{\small HalfCheetah - PPO}}    
            \label{fig:bc-compare-trans-current-ppo}
        \end{subfigure}
        \hfill
        \begin{subfigure}[b]{0.45\textwidth}   
            \centering 
            \includegraphics[width=\textwidth]{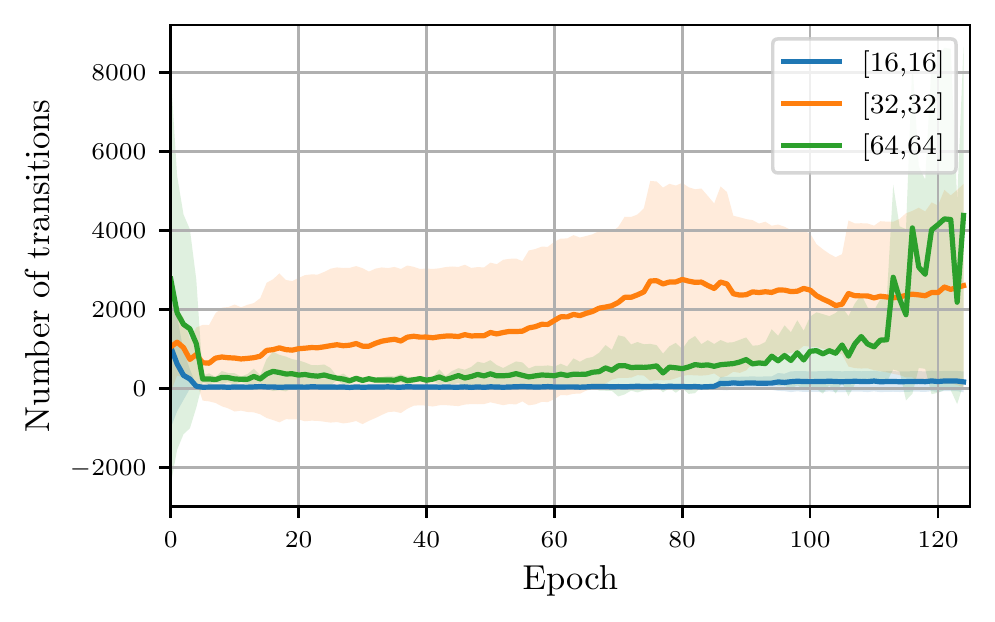}
            \caption{{\small HalfCheetah - BC}}    
            \label{fig:bc-compare-trans-current-bc}
        \end{subfigure}
        \caption{Evolution of the number of transitions over trajectories sampled from the current snapshot of the policy ($\tau$) during training HalfCheetah with BC and PPO algorithms.}
        \label{fig:bc-compare-trans-current}
    \end{figure}

\begin{figure}[H]
        \centering
        \begin{subfigure}[b]{0.45\textwidth}  
            \centering 
            \includegraphics[width=\textwidth]{figs/ppo/density-curr-fixed-cheetah.pdf}
            \caption{{\small HalfCheetah - PPO}}    
            \label{fig:bc-compare-density-both-ppo}
        \end{subfigure}
        \hfill
        \begin{subfigure}[b]{0.45\textwidth}   
            \centering 
            \includegraphics[width=\textwidth]{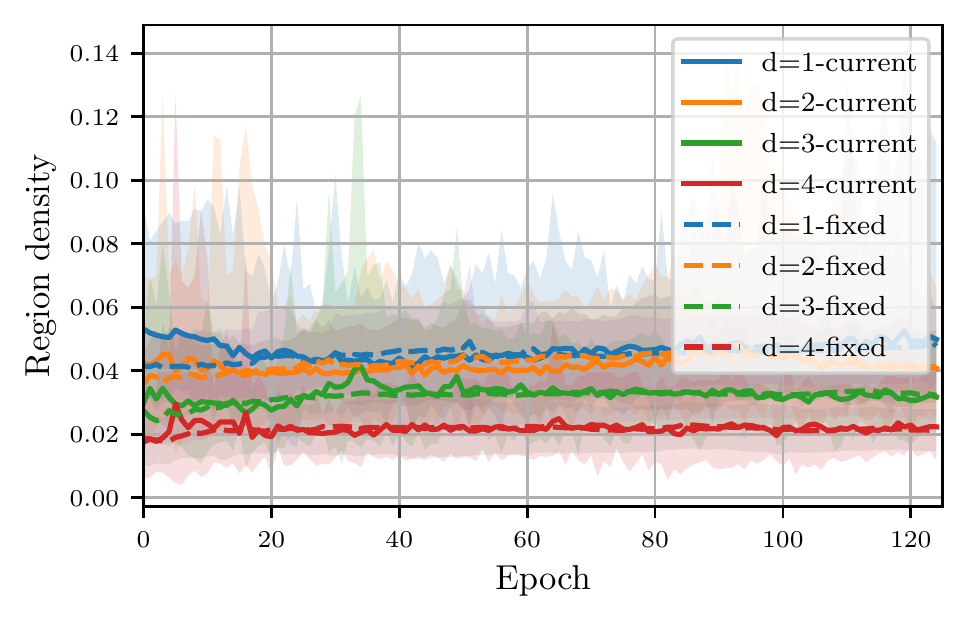}
            \caption{{\small HalfCheetah - BC}}    
            \label{fig:bc-compare-density-both-bc}
        \end{subfigure}
        \caption{Evolution of the normalized region density over both a fixed trajectory sampled from the final fully trained policy ($\tau^*$) and current trajectories sampled from the current snapshot of the policy ($\tau$) during training HalfCheetah with BC and PPO algorithms.}
        \label{fig:bc-compare-density-both}
    \end{figure}

\begin{figure}[H]
        \centering
        \begin{subfigure}[b]{0.45\textwidth}  
            \centering 
            \includegraphics[width=\textwidth]{figs/ppo/length-curr-cheetah.pdf}
            \caption{{\small HalfCheetah - PPO}}    
            \label{fig:bc-compare-length-curr-both-ppo}
        \end{subfigure}
        \hfill
        \begin{subfigure}[b]{0.45\textwidth}   
            \centering 
            \includegraphics[width=\textwidth]{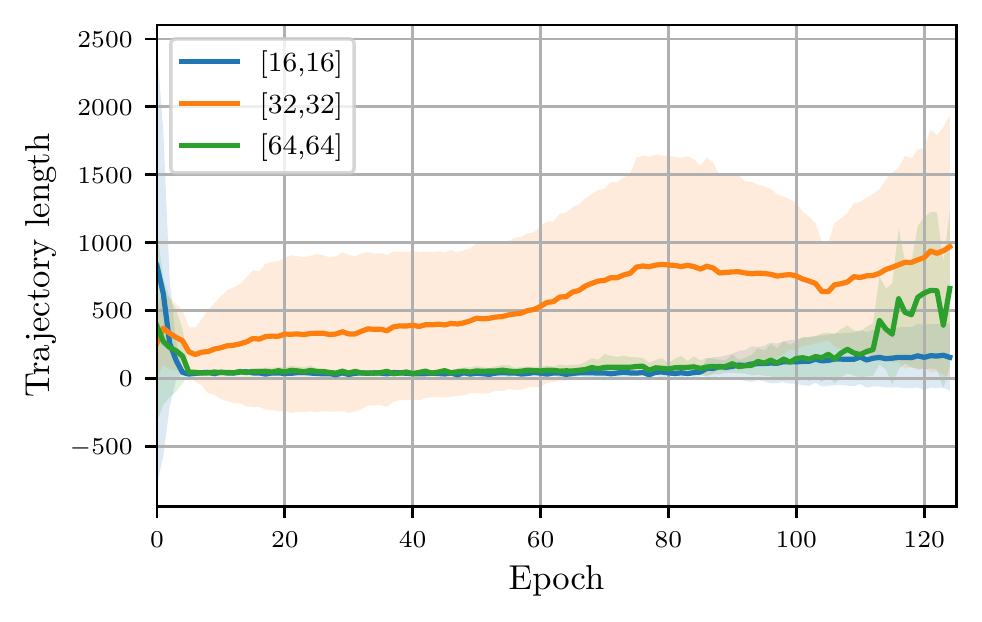}
            \caption{{\small HalfCheetah - BC}}    
            \label{fig:bc-compare-length-curr-both-bc}
        \end{subfigure}
        \caption{Evolution of the length of trajectories sampled from the current snapshot of the policy ($\tau$) during training HalfCheetah with BC and PPO algorithms. }
        \label{fig:bc-compare-length-curr-both}
    \end{figure}

\begin{figure}[H]
        \centering
        \begin{subfigure}[b]{0.45\textwidth}  
            \centering 
            \includegraphics[width=\textwidth]{figs/ppo/repeat-visits-current-cheetah.pdf}
            \caption{{\small HalfCheetah - PPO}}    
            \label{fig:bc-compare-repeat-visits-current-ppo}
        \end{subfigure}
        \hfill
        \begin{subfigure}[b]{0.45\textwidth}   
            \centering 
            \includegraphics[width=\textwidth]{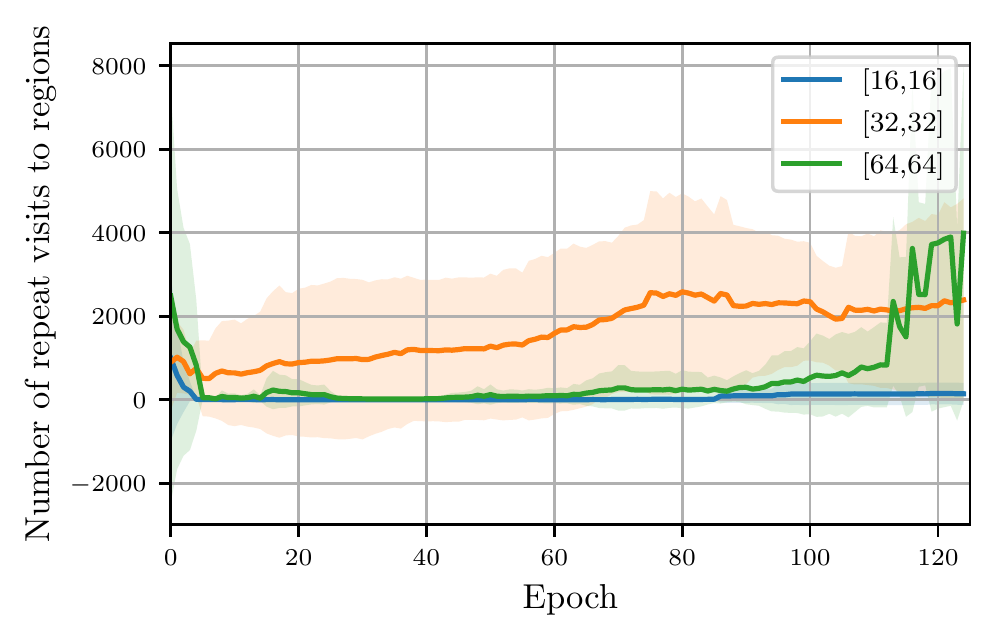}
            \caption{{\small HalfCheetah - BC}}    
            \label{fig:bc-compare-repeat-visits-current-bc}
        \end{subfigure}
        \caption{Evolution of the number of repeat visits to regions over trajectories sampled from the current snapshot of the policy ($\tau$) during training HalfCheetah with BC and PPO algorithms.}
        \label{fig:bc-compare-repeat-visits-current}
    \end{figure}

\begin{figure}[H]
        \centering
        \begin{subfigure}[b]{0.45\textwidth}  
            \centering 
            \includegraphics[width=\textwidth]{figs/ppo/density-random-lines-origin-cheetah.pdf}
            \caption{{\small HalfCheetah - PPO}}    
            \label{fig:bc-compare-density-random-lines-origin-ppo}
        \end{subfigure}
        \hfill
        \begin{subfigure}[b]{0.45\textwidth}   
            \centering 
            \includegraphics[width=\textwidth]{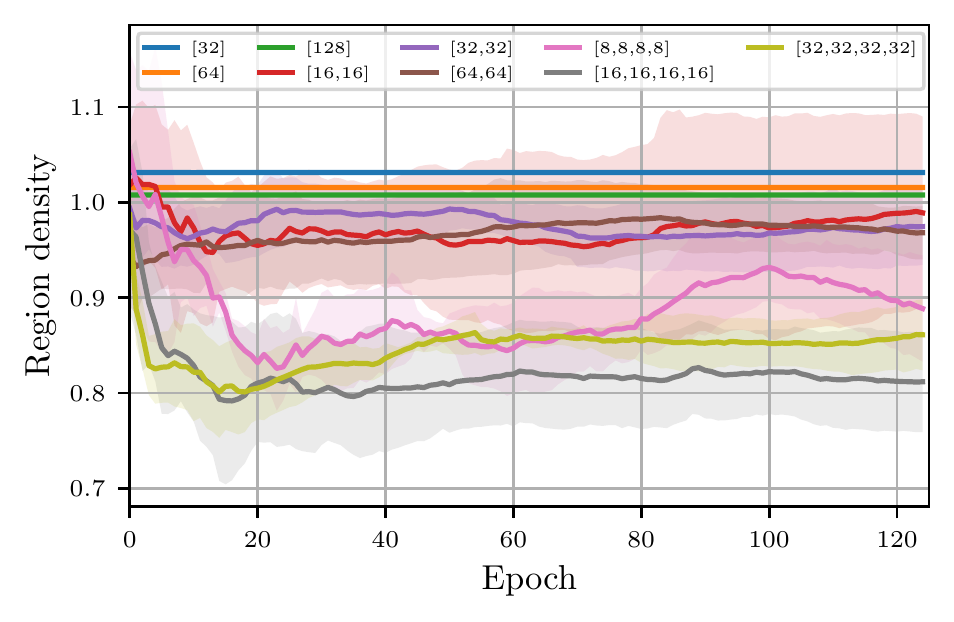}
            \caption{{\small HalfCheetah - BC}}    
            \label{fig:bc-compare-density-random-lines-origin-bc}
        \end{subfigure}
        \caption{Evolution of the mean normalized region density over 100 random lines passing through the origin during training HalfCheetah with BC and PPO algorithms. For each policy network configuration, we sample 100 random lines and compute the density of transitions as we sweep along these lines. We then report the mean density observed over these 100 lines.}
        \label{fig:bc-compare-density-random-lines-origin}
    \end{figure}

\begin{figure}[H]
        \centering
        \begin{subfigure}[b]{0.45\textwidth}  
            \centering 
            \includegraphics[width=\textwidth]{figs/ppo/density-random-traj-cheetah.pdf}
            \caption{{\small HalfCheetah - PPO}}    
            \label{fig:bc-compare-density-random-traj-ppo}
        \end{subfigure}
        \hfill
        \begin{subfigure}[b]{0.45\textwidth}   
            \centering 
            \includegraphics[width=\textwidth]{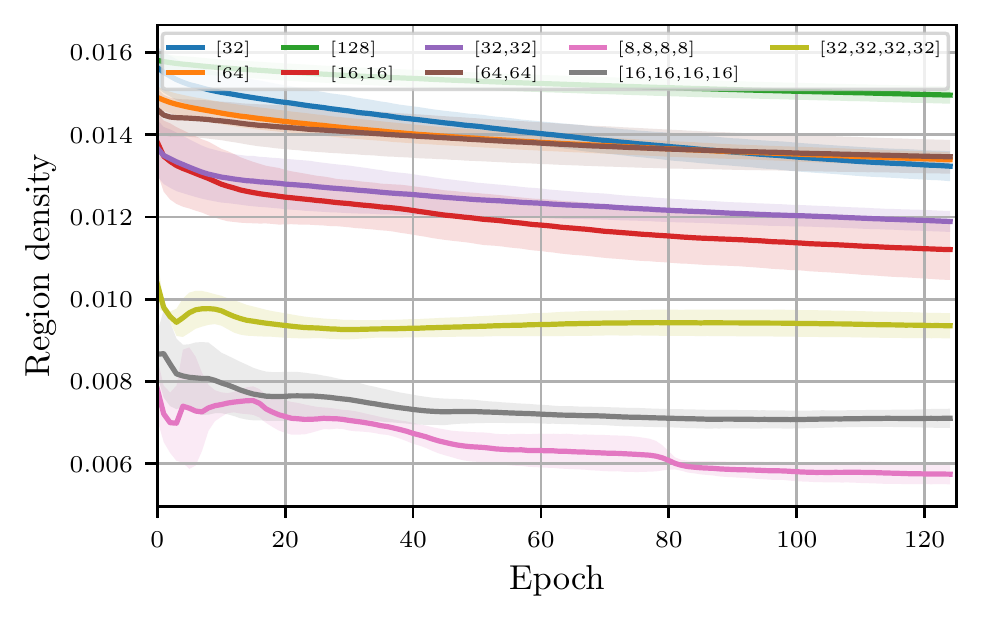}
            \caption{{\small HalfCheetah - BC}}    
            \label{fig:bc-compare-density-random-traj-bc}
        \end{subfigure}
        \caption{Evolution of the mean normalized region density over 10 random-action trajectories ($\tau^R$) during training HalfCheetah with BC and PPO algorithms. For each policy network configuration, we sample 10 random-action trajectories and compute the density of transitions as we sweep along these trajectories, and report the mean value of these trajectories.}
        \label{fig:bc-compare-density-random-traj}
    \end{figure}

Figures~\ref{fig:bc-compare-trans-fixed}-\ref{fig:bc-compare-density-random-traj} of this section, show the side-by-side comparison between the results of BC trained on HalfCheetah and results of PPO trained on HalfCheetah previously shown in Figure~\ref{fig:halfCheetah}. 
We can see that the observed region densities during training are much smaller for BC than they are for PPO and that the general trend of increased density is not visible for BC. We hypothesize that these differences are due to a combination of (i) different evolution of linear regions for networks trained with RL and supervised learning due to their inherent difference in levels of information about the state space (ii) the different network initializations used for the BC and PPO implementations.

\newpage
\section{Additional Results}\label{appendix:Additional Results}
\begin{figure}[H]
        \centering
        \begin{subfigure}[b]{0.45\textwidth}  
            \centering 
            \includegraphics[width=\textwidth]{figs/ppo/transitions-fixed-cheetah.pdf}
            \caption{{\small HalfCheetah}}    
            \label{fig:supp-trans-fixed-hc}
        \end{subfigure}
        \hfill
        \begin{subfigure}[b]{0.45\textwidth}   
            \centering 
            \includegraphics[width=\textwidth]{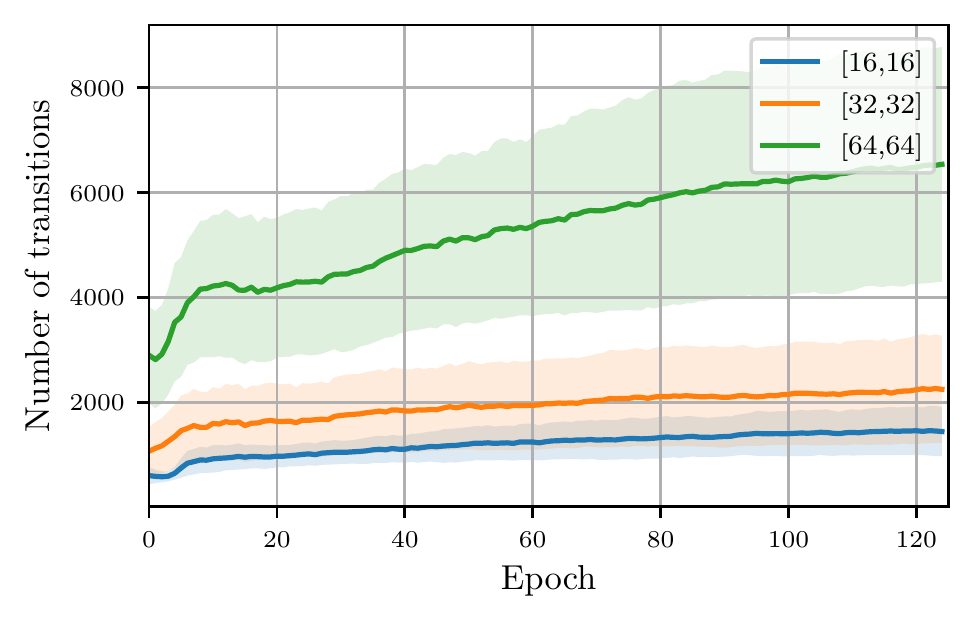}
            \caption{{\small Walker}}    
            \label{fig:supp-trans-fixed-w}
        \end{subfigure}
        \vskip\baselineskip
        \begin{subfigure}[b]{0.45\textwidth}
            \centering
            \includegraphics[width=\textwidth]{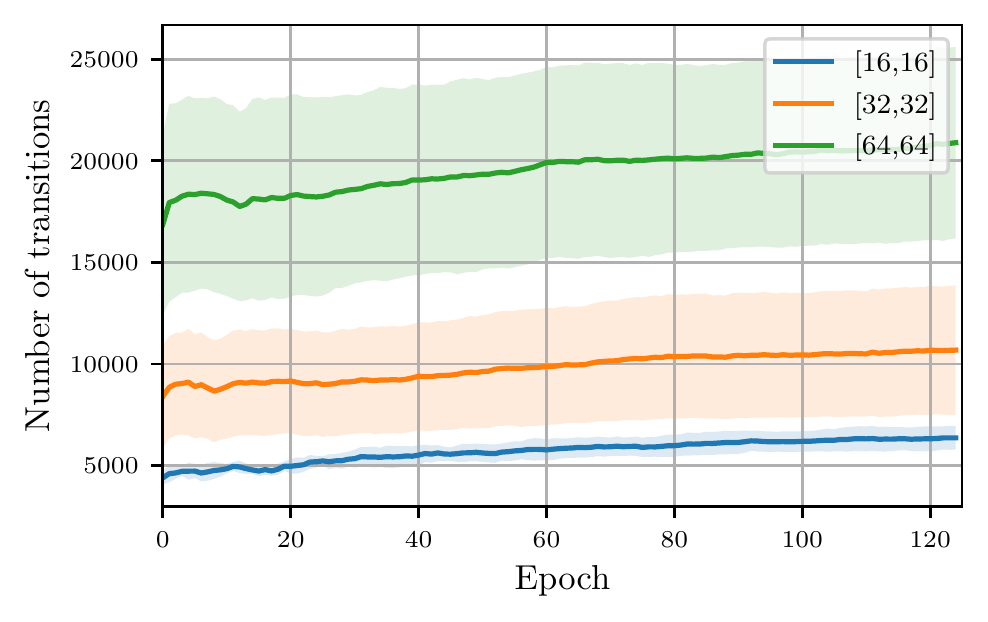}
            \caption{{\small Ant}}    
            \label{fig:supp-trans-fixed-a}
        \end{subfigure}
        \hfill        
        \begin{subfigure}[b]{0.45\textwidth}   
            \centering 
            \includegraphics[width=\textwidth]{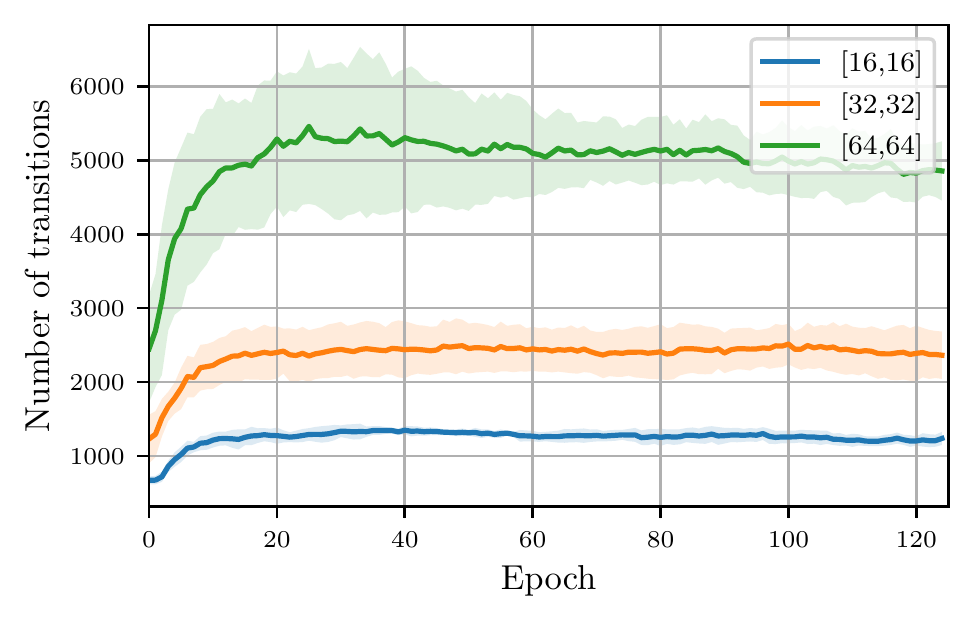}
            \caption{{\small Swimmer}}    
            \label{fig:supp-trans-fixed-s}
        \end{subfigure}
        \caption{Evolution of the number of transitions over a fixed trajectory sampled from the final fully trained policy ($\tau^*$) during training for different tasks. Plots show the mean and standard error across 5 random seeds. In the legend, $[n_1, ..., n_d]$ corresponds to a network architecture with depth $d$ and $n_i$ neurons in each layer. These plots show a moderate and gradual increase in the number of transitions over a fixed trajectory observed during training, with larger policy networks having more transitions.}
        \label{fig:supp-trans-fixed}
    \end{figure}

\begin{figure}[H]
        \centering
        \begin{subfigure}[b]{0.45\textwidth}  
            \centering 
            \includegraphics[width=\textwidth]{figs/ppo/density-fixed-depth-cheetah.pdf}
            \caption{{\small HalfCheetah}}    
            \label{fig:supp-density-fixed-hc}
        \end{subfigure}
        \hfill
        \begin{subfigure}[b]{0.45\textwidth}   
            \centering 
            \includegraphics[width=\textwidth]{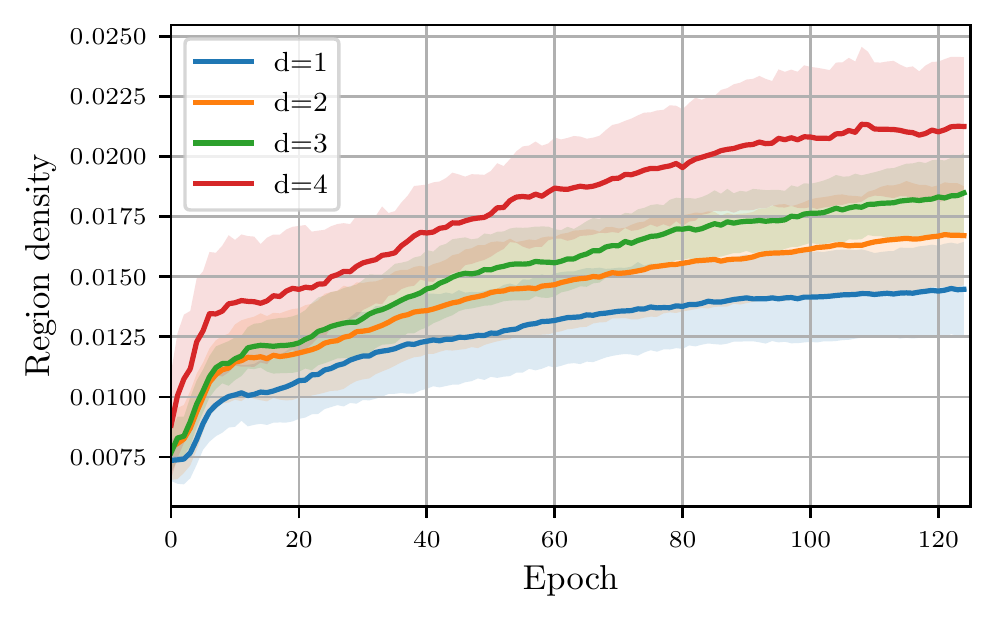}
            \caption{{\small Walker}}    
            \label{fig:supp-density-fixed-w}
        \end{subfigure}
        \vskip\baselineskip
        \begin{subfigure}[b]{0.45\textwidth}
            \centering
            \includegraphics[width=\textwidth]{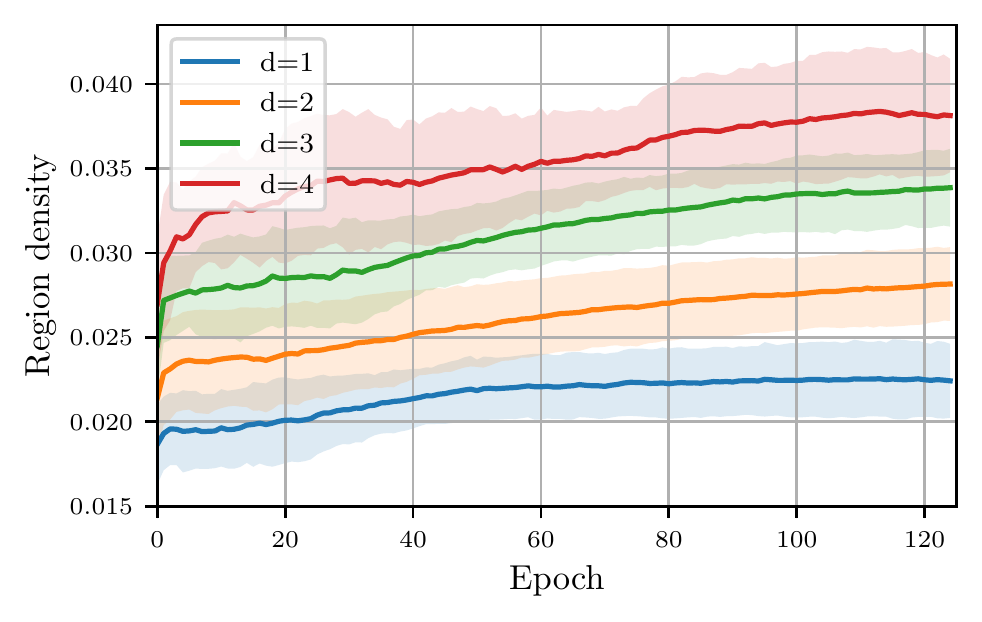}
            \caption{{\small Ant}}    
            \label{fig:supp-density-fixed-a}
        \end{subfigure}
        \hfill        
        \begin{subfigure}[b]{0.45\textwidth}   
            \centering 
            \includegraphics[width=\textwidth]{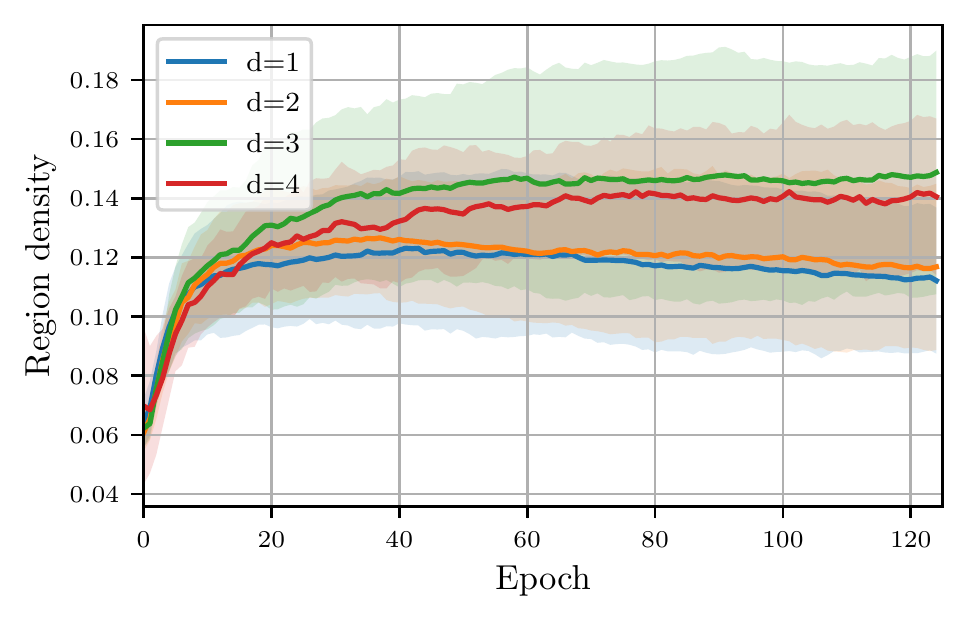}
            \caption{{\small Swimmer}}    
            \label{fig:supp-density-fixed-s}
        \end{subfigure}
        \caption{Evolution of the normalized region density over a fixed trajectory sampled from the final fully trained policy ($\tau^*$) during training for different tasks. Plots show the mean and the standard error over all networks with equal depth. Note that the vertical axes do not begin at zero. These plots show a moderate and gradual increase in the density of transitions over a fixed trajectory during training. We can also observe that deeper policy networks results in moderately denser regions in the learned policies. }
        \label{fig:supp-density-fixed}
    \end{figure}

\begin{figure}[H]
        \centering
        \begin{subfigure}[b]{0.45\textwidth}  
            \centering 
            \includegraphics[width=\textwidth]{figs/ppo/transitions-current-cheetah.pdf}
            \caption{{\small HalfCheetah}}    
            \label{fig:supp-trans-current-hc}
        \end{subfigure}
        \hfill
        \begin{subfigure}[b]{0.45\textwidth}   
            \centering 
            \includegraphics[width=\textwidth]{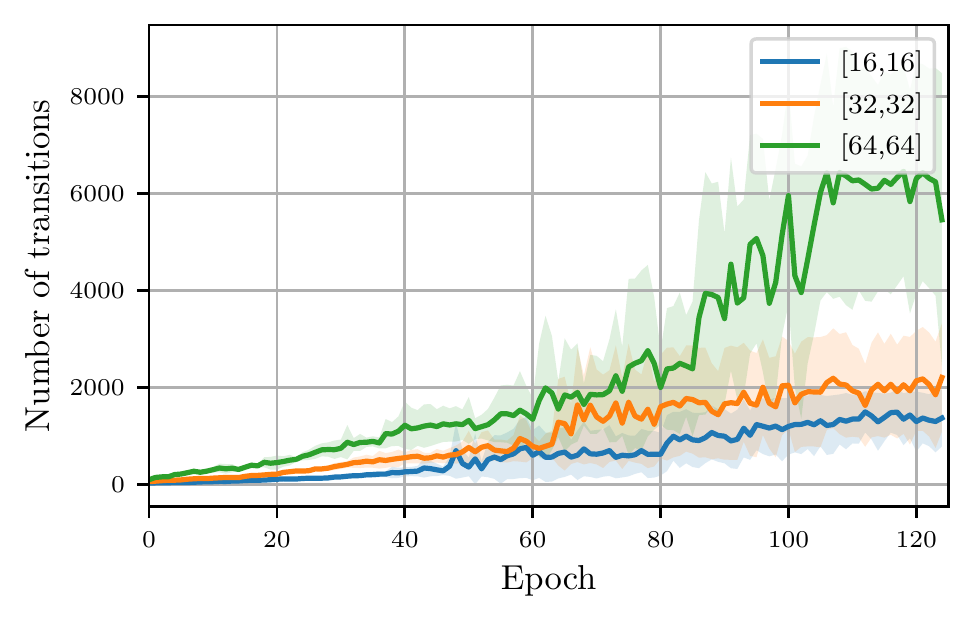}
            \caption{{\small Walker}}    
            \label{fig:supp-trans-current-w}
        \end{subfigure}
        \vskip\baselineskip
        \begin{subfigure}[b]{0.45\textwidth}
            \centering
            \includegraphics[width=\textwidth]{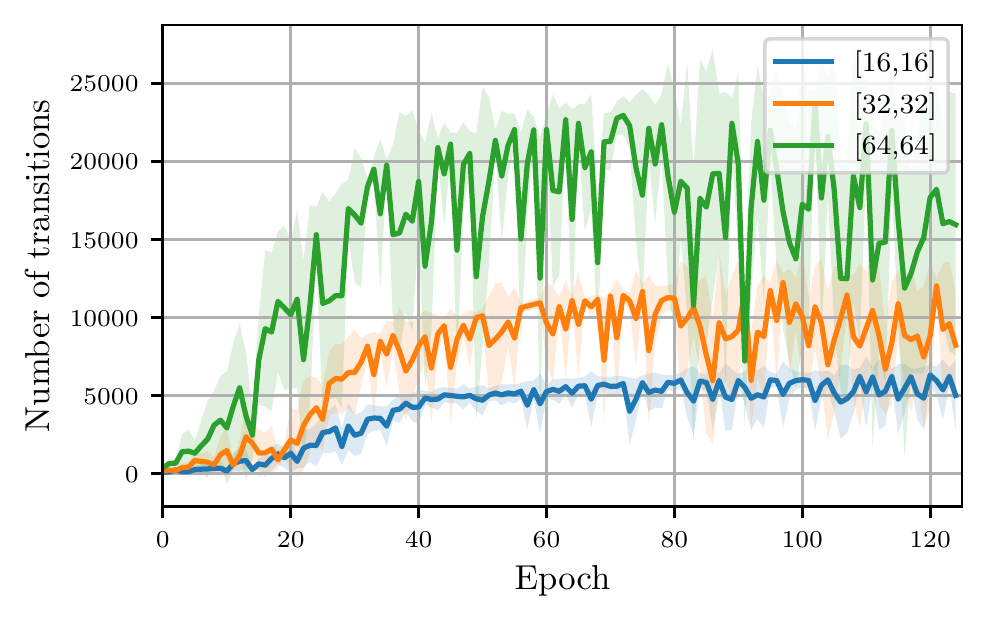}
            \caption{{\small Ant}}    
            \label{fig:supp-trans-current-a}
        \end{subfigure}
        \hfill        
        \begin{subfigure}[b]{0.45\textwidth}   
            \centering 
            \includegraphics[width=\textwidth]{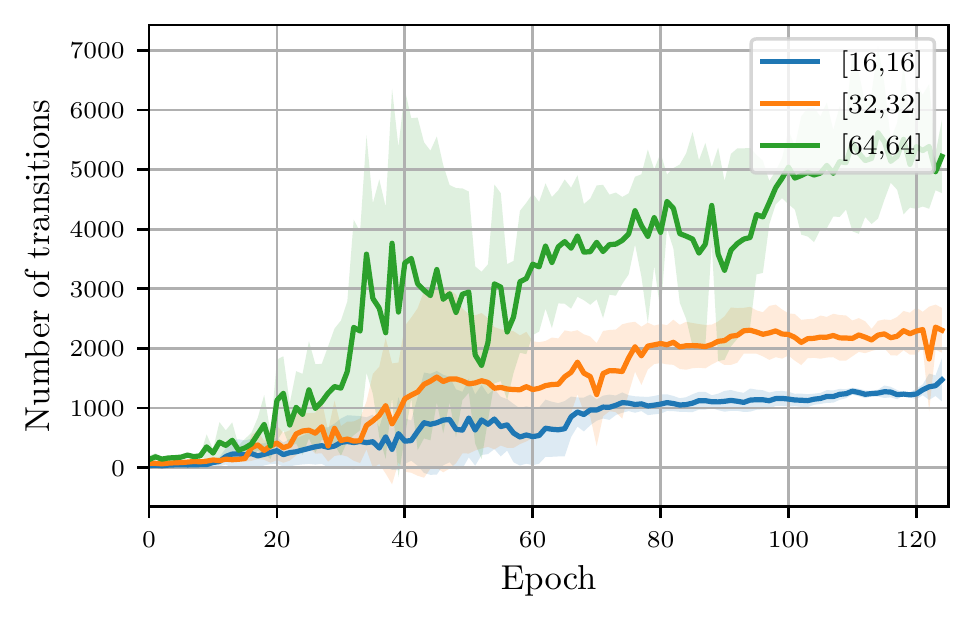}
            \caption{{\small Swimmer}}    
            \label{fig:supp-trans-current-s}
        \end{subfigure}
        \caption{Evolution of the number of transitions over trajectories sampled from the current snapshot of the policy ($\tau$) during training for different tasks. These plots show an increase in the number of transitions observed on trajectories sampled from current snapshots of the policy during training.}
        \label{fig:supp-trans-current}
    \end{figure}

\begin{figure}[H]
        \centering
        \begin{subfigure}[b]{0.45\textwidth}  
            \centering 
            \includegraphics[width=\textwidth]{figs/ppo/density-curr-fixed-cheetah.pdf}
            \caption{{\small HalfCheetah}}    
            \label{fig:supp-density-both-hc}
        \end{subfigure}
        \hfill
        \begin{subfigure}[b]{0.45\textwidth}   
            \centering 
            \includegraphics[width=\textwidth]{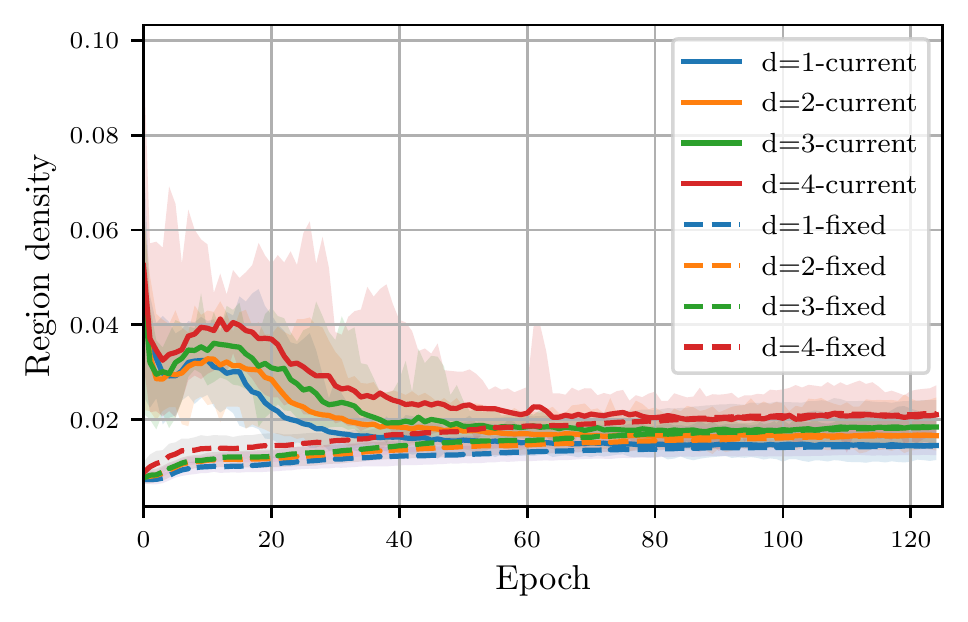}
            \caption{{\small Walker}}    
            \label{fig:supp-density-both-w}
        \end{subfigure}
        \vskip\baselineskip
        \begin{subfigure}[b]{0.45\textwidth}
            \centering
            \includegraphics[width=\textwidth]{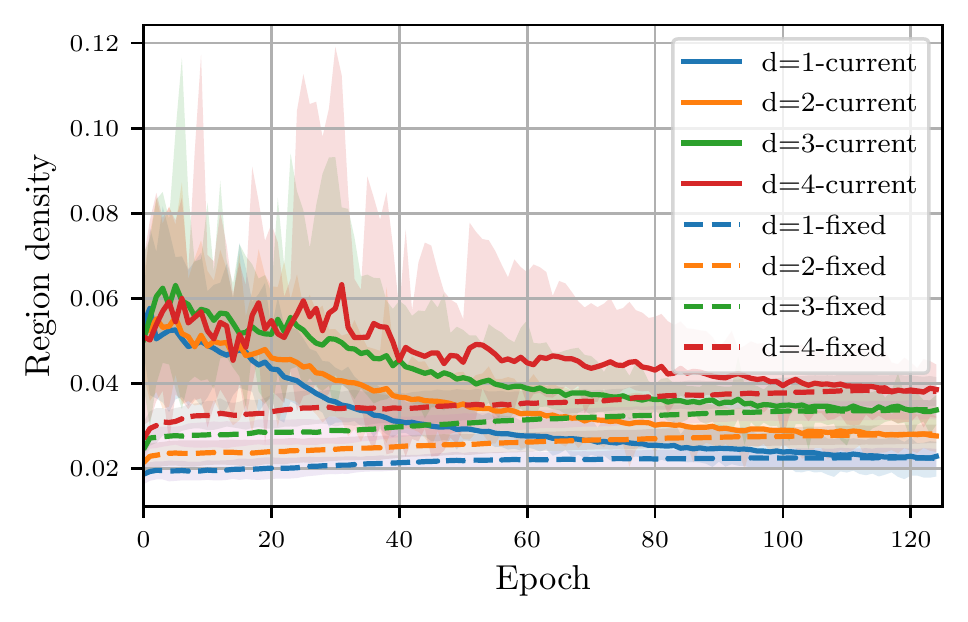}
            \caption{{\small Ant}}    
            \label{fig:supp-density-both-a}
        \end{subfigure}
        \hfill        
        \begin{subfigure}[b]{0.45\textwidth}   
            \centering 
            \includegraphics[width=\textwidth]{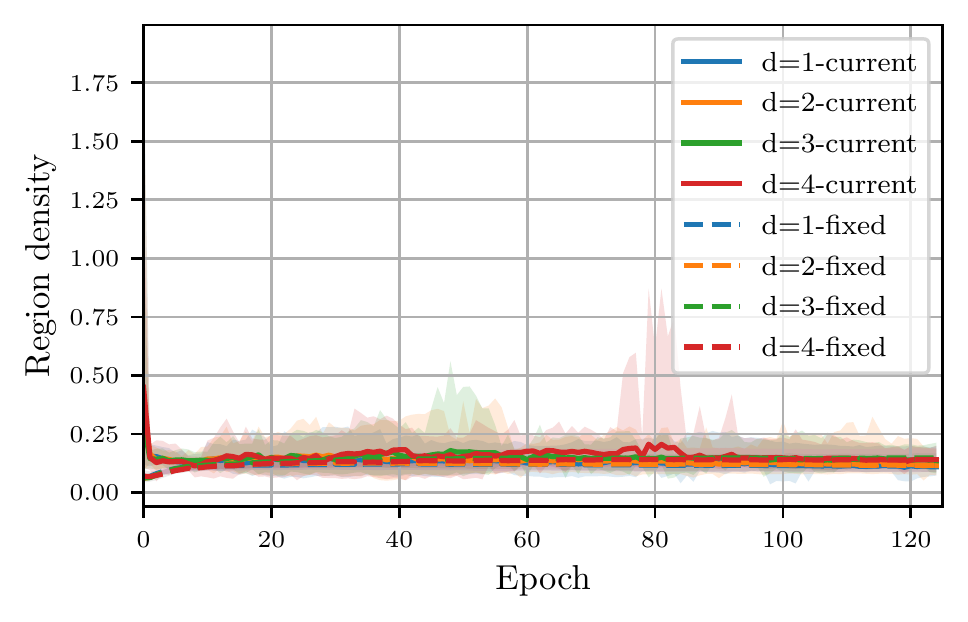}
            \caption{{\small Swimmer}}    
            \label{fig:supp-density-both-s}
        \end{subfigure}
        \caption{Evolution of the normalized region density over both a fixed trajectory sampled from the final fully trained policy ($\tau^*$) and current trajectories sampled from the current snapshot of the policy ($\tau$) during training for different tasks. These plots show that the density over fixed and current trajectories converges to the same values over training, while it increases for former and decreases for the latter. We speculate that the density is higher for current trajectories earlier during training due to early exploration and the form of network initialization.}
        \label{fig:supp-density-both}
    \end{figure}

\begin{figure}[H]
        \centering
        \begin{subfigure}[b]{0.45\textwidth}  
            \centering 
            \includegraphics[width=\textwidth]{figs/ppo/length-curr-cheetah.pdf}
            \caption{{\small HalfCheetah}}    
            \label{fig:supp-length-curr-hc}
        \end{subfigure}
        \hfill
        \begin{subfigure}[b]{0.45\textwidth}   
            \centering 
            \includegraphics[width=\textwidth]{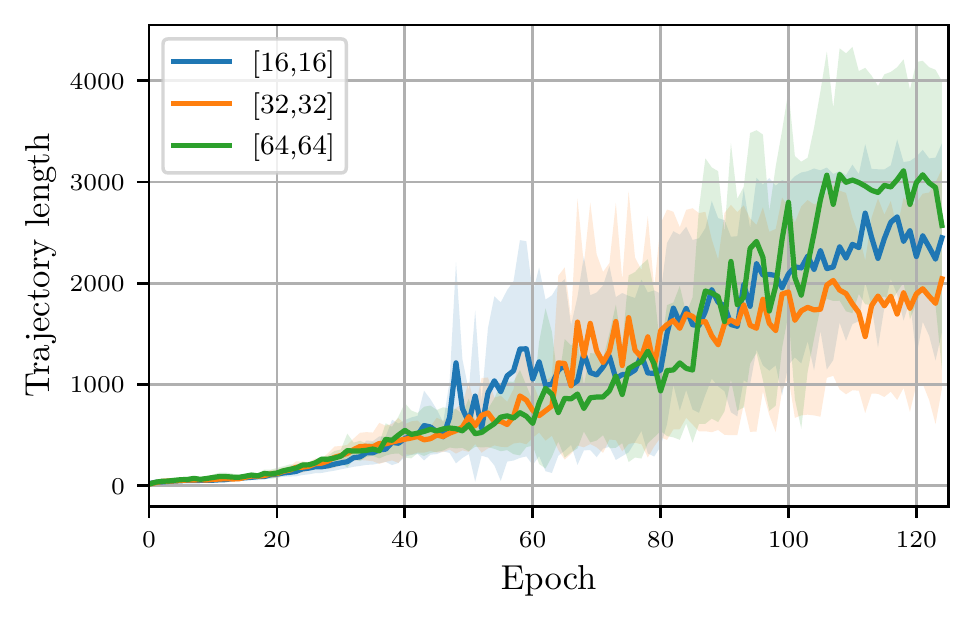}
            \caption{{\small Walker}}    
            \label{fig:supp-length-curr-w}
        \end{subfigure}
        \vskip\baselineskip
        \begin{subfigure}[b]{0.45\textwidth}
            \centering
            \includegraphics[width=\textwidth]{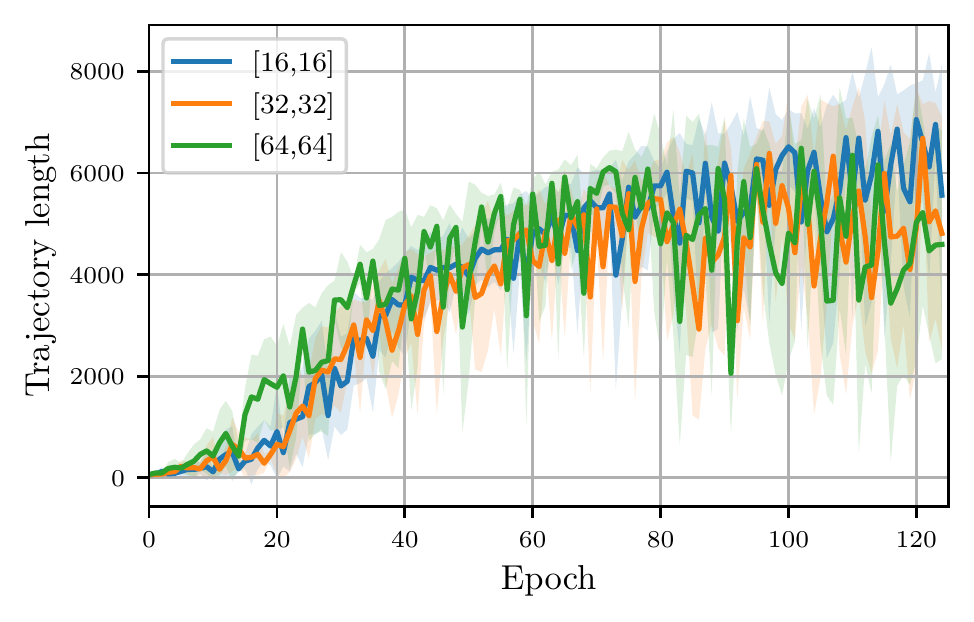}
            \caption{{\small Ant}}    
            \label{fig:supp-length-curr-a}
        \end{subfigure}
        \hfill        
        \begin{subfigure}[b]{0.45\textwidth}   
            \centering 
            \includegraphics[width=\textwidth]{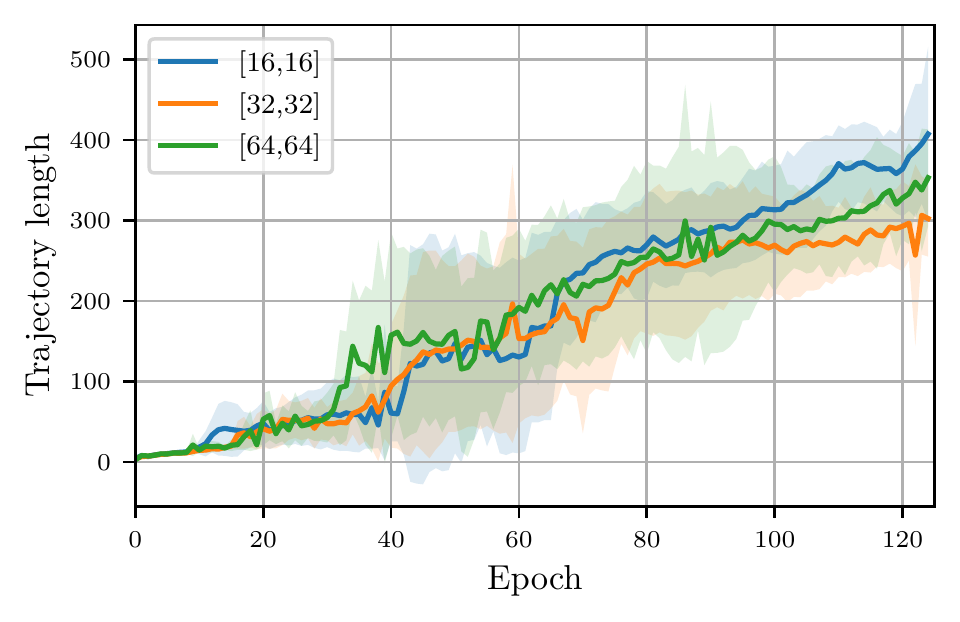}
            \caption{{\small Swimmer}}    
            \label{fig:supp-length-curr-s}
        \end{subfigure}
        \caption{Evolution of the length of trajectories sampled from the current snapshot of the policy ($\tau$) during training for different tasks. These plots show that the length of the trajectories increase with training.}
        \label{fig:supp-lenght-curr}
    \end{figure}

\begin{figure}[H]
        \centering
        \begin{subfigure}[b]{0.45\textwidth}  
            \centering 
            \includegraphics[width=\textwidth]{figs/ppo/repeat-visits-current-cheetah.pdf}
            \caption{{\small HalfCheetah}}    
            \label{fig:supp-repeat-visits-current-hc}
        \end{subfigure}
        \hfill
        \begin{subfigure}[b]{0.45\textwidth}   
            \centering 
            \includegraphics[width=\textwidth]{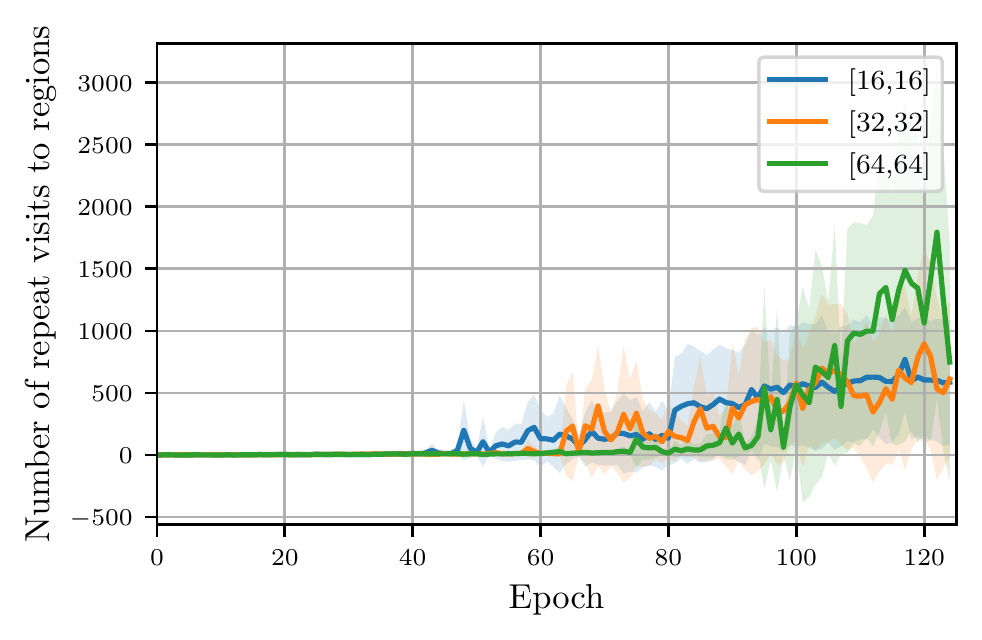}
            \caption{{\small Walker}}    
            \label{fig:supp-repeat-visits-current-w}
        \end{subfigure}
        \vskip\baselineskip
        \begin{subfigure}[b]{0.45\textwidth}
            \centering
            \includegraphics[width=\textwidth]{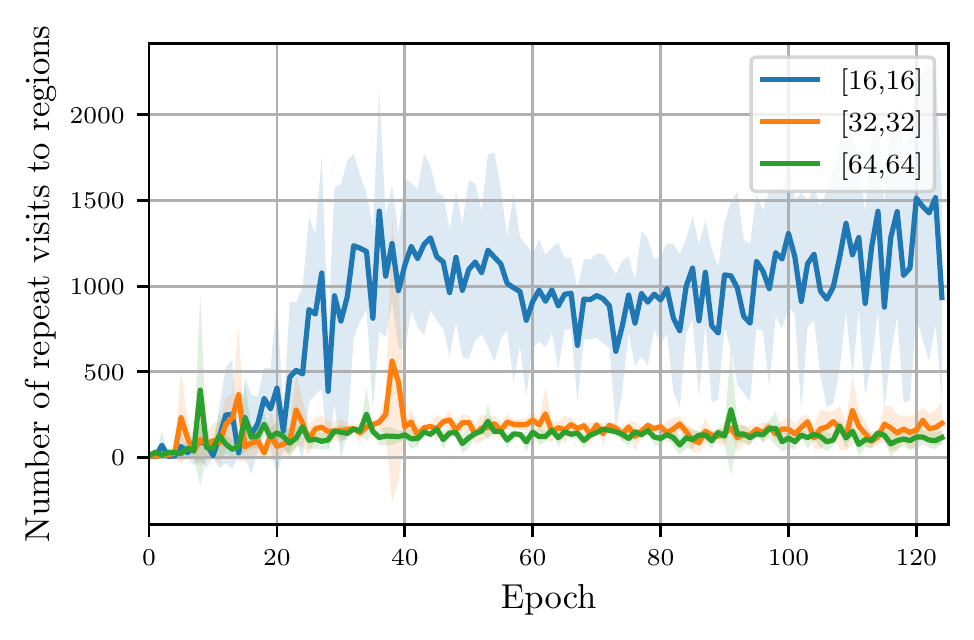}
            \caption{{\small Ant}}    
            \label{fig:supp-repeat-visits-current-a}
        \end{subfigure}
        \hfill        
        \begin{subfigure}[b]{0.45\textwidth}   
            \centering 
            \includegraphics[width=\textwidth]{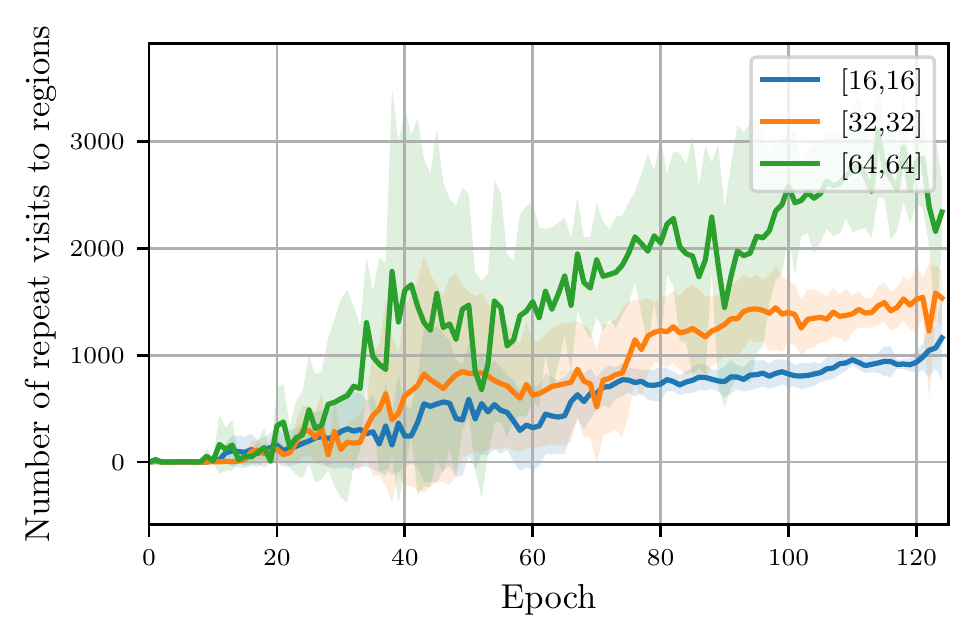}
            \caption{{\small Swimmer}}    
            \label{fig:supp-repeat-visits-current-s}
        \end{subfigure}
        \caption{Evolution of the number of repeat visits to regions over trajectories sampled from the current snapshot of the policy ($\tau$) during training for different tasks. We can see that the number of repeat visits generally increase with training  because of the cyclic trajectories resulting from locomotion-based tasks. We speculate that for HalfCheetah, repeat visits are high earlier during training because of limited exploration.}
        \label{fig:supp-repeat-visits-current}
    \end{figure}

\begin{figure}[H]
        \centering
        \begin{subfigure}[b]{0.45\textwidth}  
            \centering 
            \includegraphics[width=\textwidth]{figs/ppo/density-random-lines-origin-cheetah.pdf}
            \caption{{\small HalfCheetah}}    
            \label{fig:supp-density-random-lines-origin-hc}
        \end{subfigure}
        \hfill
        \begin{subfigure}[b]{0.45\textwidth}   
            \centering 
            \includegraphics[width=\textwidth]{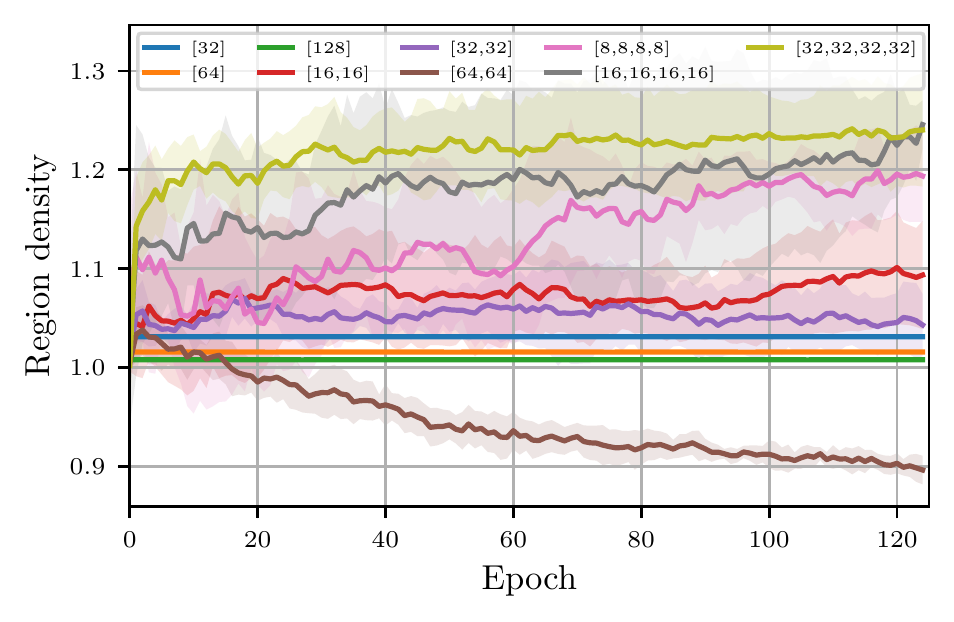}
            \caption{{\small Walker}}    
            \label{fig:supp-density-random-lines-origin-w}
        \end{subfigure}
        \vskip\baselineskip
        \begin{subfigure}[b]{0.45\textwidth}
            \centering
            \includegraphics[width=\textwidth]{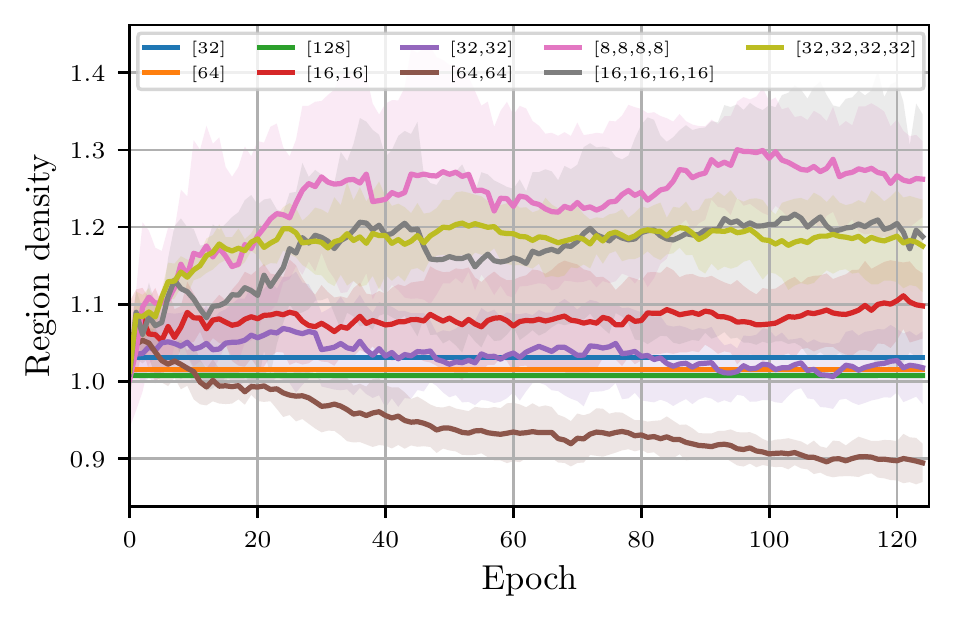}
            \caption{{\small Ant}}    
            \label{fig:supp-density-random-lines-origin-a}
        \end{subfigure}
        \hfill        
        \begin{subfigure}[b]{0.45\textwidth}   
            \centering 
            \includegraphics[width=\textwidth]{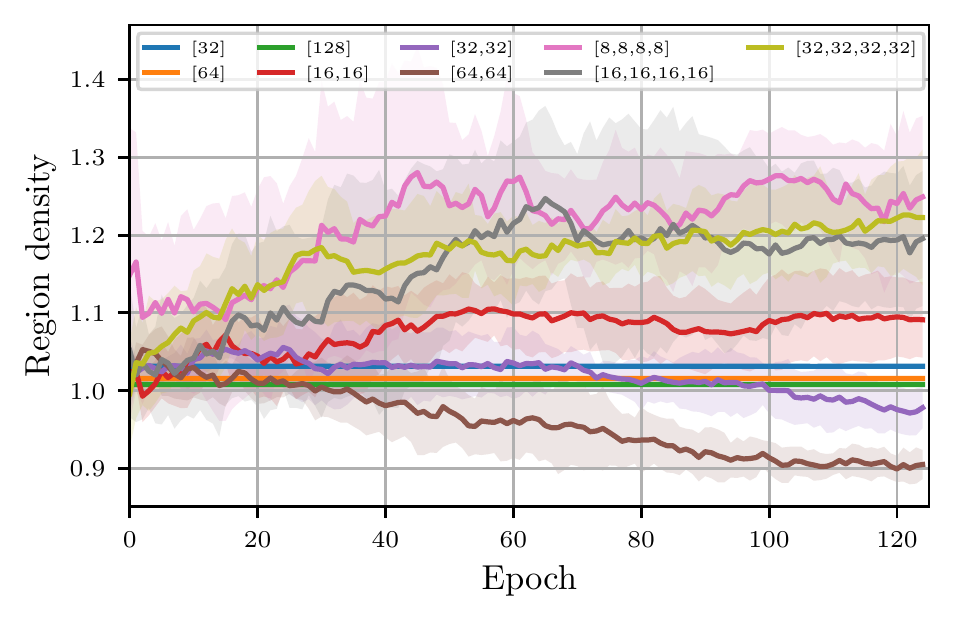}
            \caption{{\small Swimmer}}    
            \label{fig:supp-density-random-lines-origin-s}
        \end{subfigure}
        \caption{Evolution of the mean normalized region density over 100 random lines passing through the origin during training for different tasks. For each policy network configuration, we sample 100 random lines and compute the density of transitions as we sweep along these lines. We then report the mean density observed over these 100 lines. These plots show that the mean normalized density starts close to $1$ at initialization and remains roughly constant during training which is consistent with the findings of \citet{hanin2019complexity}. Note that because the vertical axis does not begin at zero, the  variations around 1.0 is scaled.}
        \label{fig:supp-density-random-lines-origin}
    \end{figure}

\begin{figure}[H]
        \centering
        \begin{subfigure}[b]{0.45\textwidth}  
            \centering 
            \includegraphics[width=\textwidth]{figs/ppo/density-random-traj-cheetah.pdf}
            \caption{{\small HalfCheetah}}    
            \label{fig:supp-density-random-traj-hc}
        \end{subfigure}
        \hfill
        \begin{subfigure}[b]{0.45\textwidth}   
            \centering 
            \includegraphics[width=\textwidth]{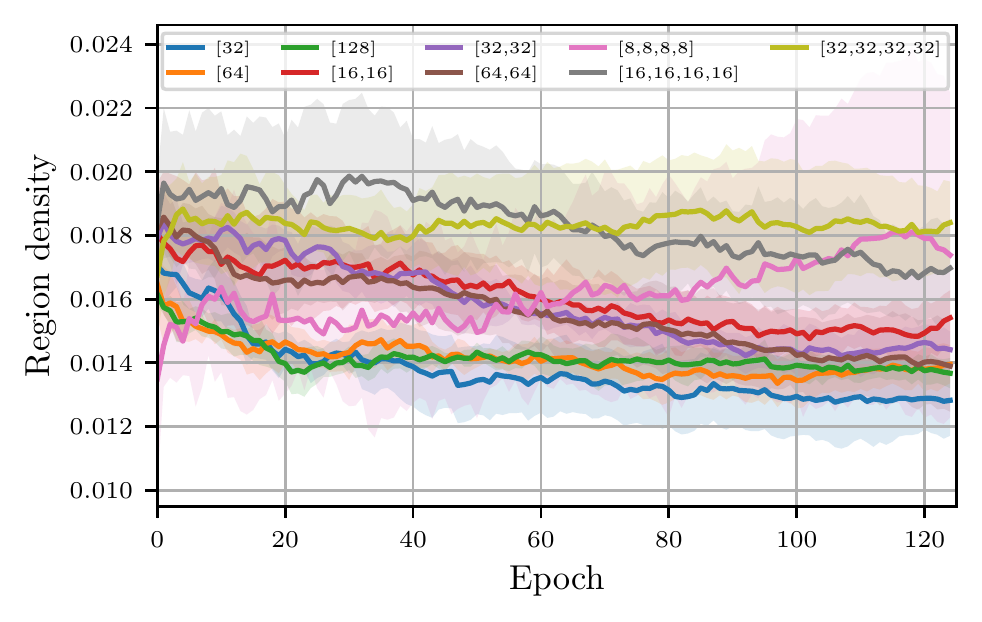}
            \caption{{\small Walker}}    
            \label{fig:supp-density-random-traj-w}
        \end{subfigure}
        \vskip\baselineskip
        \begin{subfigure}[b]{0.45\textwidth}
            \centering
            \includegraphics[width=\textwidth]{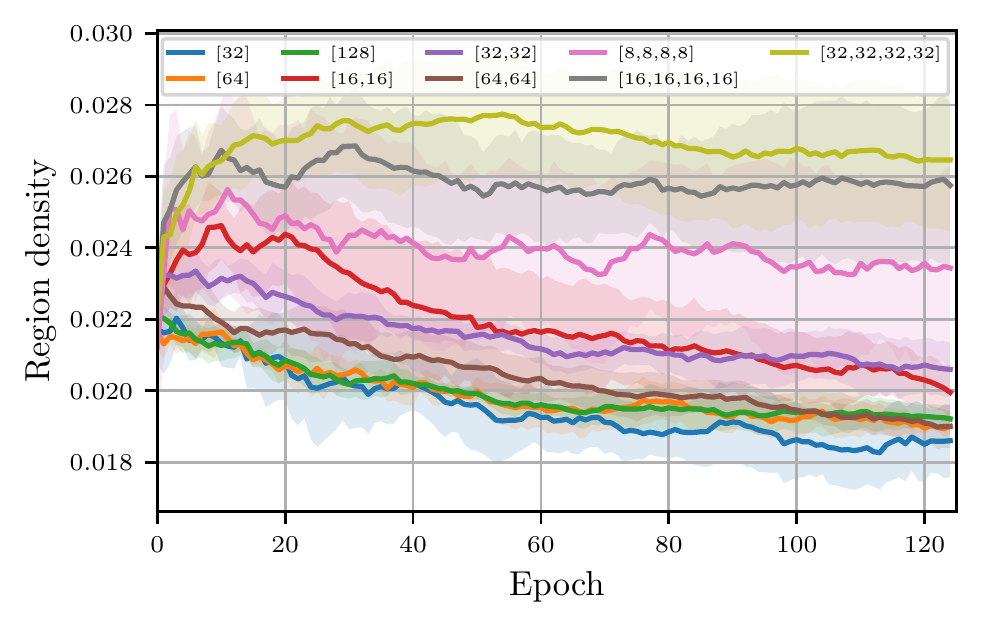}
            \caption{{\small Ant}}    
            \label{fig:supp-density-random-traj-a}
        \end{subfigure}
        \hfill        
        \begin{subfigure}[b]{0.45\textwidth}   
            \centering 
            \includegraphics[width=\textwidth]{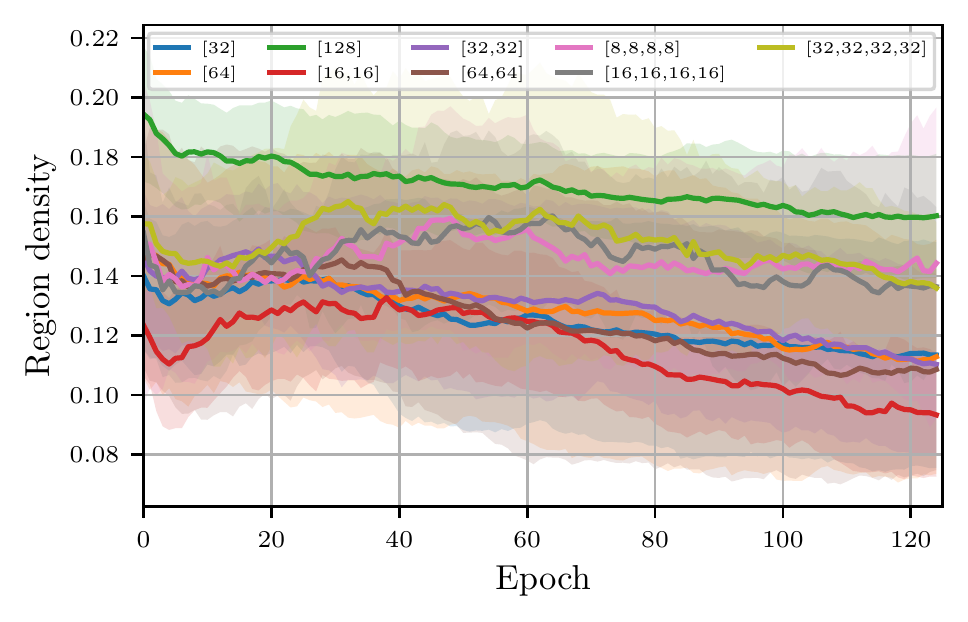}
            \caption{{\small Swimmer}}    
            \label{fig:supp-density-random-traj-s}
        \end{subfigure}
        \caption{Evolution of the mean normalized region density over 10 random-action trajectories ($\tau^R$) during training for different tasks. For each policy network configuration, we sample 10 random-action trajectories and compute the density of transitions as we sweep along these trajectories, and report the mean value of these trajectories. These plots show that the observed normalized density for random-action trajectories decreases slightly with training. When compared with the results of Figure~\ref{fig:supp-density-both}, we observe that the observed densities are marginally less than that of fixed and current trajectories.}
        \label{fig:supp-density-random-traj}
    \end{figure}

\begin{figure}[H]
        \centering
        \begin{subfigure}[b]{0.45\textwidth}  
            \centering 
            \includegraphics[width=\textwidth]{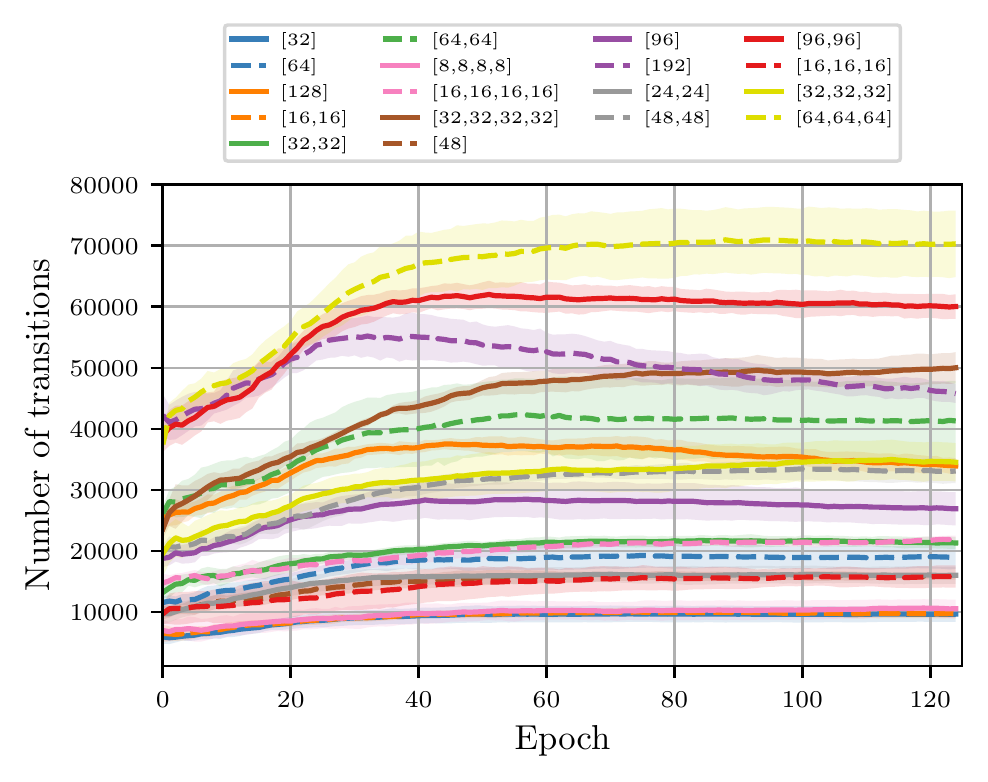}
            \caption{{\small Transition count, fixed trajectory $\tau^*$}}    
            \label{fig:supp-accomp-fixed-hc}
        \end{subfigure}
        \hfill
        \begin{subfigure}[b]{0.45\textwidth}   
            \centering 
            \includegraphics[width=\textwidth]{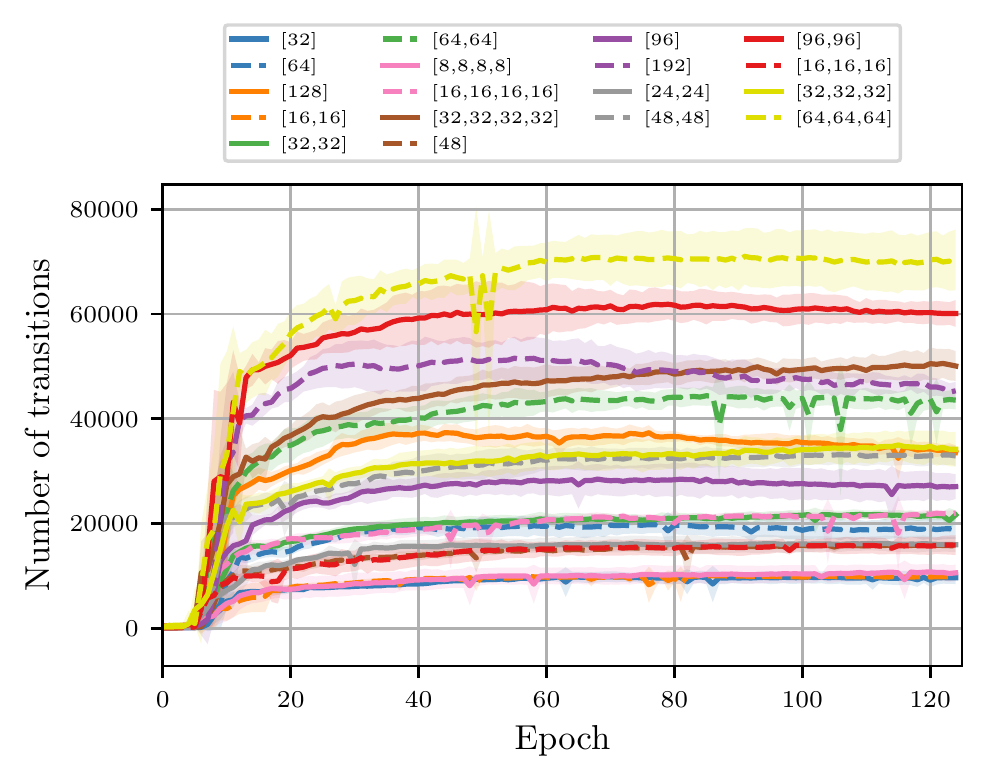}
            \caption{{\small Transition count, current trajectory $\tau$}}    
            \label{fig:supp-accomp-current-hc}
        \end{subfigure}
        \vskip\baselineskip
        \caption{Evolution of the number of transitions over fixed and current trajectories during training for all policy network architectures trained on HalfCheetah. This figure is included for the sake of completeness.}
        \label{fig:supp-accomp-hc}
    \end{figure}

We previously provided the results for the HalfCheetah environment in Figure~\ref{fig:halfCheetah}.
Figures~\ref{fig:supp-trans-fixed}-\ref{fig:supp-density-random-traj} of this section, show the same set of plots for the three remaining environments trained with PPO. Plots are now grouped by metric type instead of environment, to enable easier comparison between environments.
Figure~\ref{fig:supp-accomp-hc} shows the full set of results of Figures~\ref{fig:trans-fixed} and \ref{fig:trans-current} for all policy network architectures trained on HalfCheetah.

\clearpage
    
\end{document}